\RequirePackage{silence}
\RequirePackage{fix-cm}
\documentclass[journal]{IEEEtran}

\usepackage[hypertexnames=false,hidelinks]{hyperref}
\usepackage{float,caption,dsfont,bm,amsfonts,color,pgffor}
\usepackage{amsmath,amsfonts,amssymb,commath,mathtools,physics,algorithm,algpseudocode}%
\usepackage{enumitem,microtype,subfigure,array,,comment,booktabs,bookmark,graphicx,threeparttable}

\newtheorem{theorem}{Theorem}
\newtheorem{problem}{Problem}
\newtheorem{remark}{Remark}[section]
\newtheorem{definition}{Definition}

\newcommand{\x}{\mathbf x}
\newcommand{\p}{\mathbf p}
\newcommand{\one}{{\mathds 1}}
\newcommand{\argmax}{\arg\max}
\newcommand{\linref}[1]{line~$\ref{#1}$}
\newcommand{\linsref}[2]{lines~$\ref{#1}$-$\ref{#2}$}

\renewcommand{\secref}[1]{Section~\ref{#1}}
\renewcommand{\figref}[1]{Fig.~$\ref{#1}$}
\renewcommand{\algref}[1]{Algorithm~$\ref{#1}$}

\newcommand{\tabref}[1]{Table~$\ref{#1}$}
\newcommand{\prbref}[1]{Problem~\ref{#1}}
\newcommand{\defref}[1]{Definition~\ref{#1}}

\newcolumntype{F}[1]{%
    >{\raggedright\arraybackslash\hspace{0pt}}p{#1}}%

\let\oldforall\forall
\let\forall\undefined
\DeclareMathOperator{\forall}{\oldforall}

\newcommand\makemathcal[1]{\foreach\z in{#1}{\expandafter\gdef\csname\z\expandafter\endcsname\expandafter{\expandafter\mathcal\expandafter{\z}}}}

\newcommand\makemathbfvec[1]{\foreach\z in{#1}{\expandafter\gdef\csname\z\expandafter\endcsname\expandafter{\expandafter\mathbf\expandafter{\z}}}}

\newcommand\makemathbfmat[1]{\foreach\a\b in{#1}{\expandafter\gdef\csname\a\b\expandafter\endcsname\expandafter{\expandafter\mathbf\expandafter{\a}}}}

\makeatletter
\count@=1
\loop
\expanded{\noexpand\bmdefine\csname\@Alph\count@\@Alph\count@\@Alph\count@\endcsname
   {\noexpand\mathsf{\@Alph\count@}}}
\ifnum\count@<26
\advance\count@\@ne
\repeat

\makemathcal{A,...,Z}
\makemathbfvec{a,...,z}
\makemathbfmat{AA,...,ZZ}

\renewcommand{\EE}{\mathbb E}
\renewcommand{\RR}{\mathbb R}

\newtheorem{assumption}{Assumption}
\graphicspath{{figures/}}

\def\imwidth{2.4in}
\usepackage[noadjust]{cite}

\title{Intermittent Deployment for Large-Scale Multi-Robot Forage Perception: Data Synthesis, Prediction, and Planning}

\author{Jun Liu$^{1}$, Murtaza Rangwala$^{1}$, Kulbir Singh Ahluwalia$^{2}$, Shayan Ghajar$^{3}$, Harnaik Singh Dhami$^{4}$, Pratap Tokekar$^{4}$, Benjamin F. Tracy$^{5}$, and Ryan K. Williams$^{1}$
\thanks{This work was supported by the National Institute of Food and Agriculture under Grant 2018-67007-28380.}
\thanks{$^{1}$The authors are with the Department of Electrical and Computer Engineering, Virginia Tech, Blacksburg, VA 24061 USA (e-mail: junliu@vt.edu; murtazar@vt.edu; rywilli1@vt.edu).}
\thanks{$^{2}$The author is with the Department of Agriculture and Biological Engineering, University of Illinois, Urbana-Champaign, IL 61801 USA (e-mail: ksa5@illinois.edu).}
\thanks{$^{3}$The author is with the Department of Crop and Soil Science, Oregon State University, Corvallis, OR 97331 USA (e-mail: ghajars@oregonstate.edu).}
\thanks{$^{4}$The author is with the Department of Computer Science, University of Maryland, College Park, MD 20782 USA (e-mail: dhami@umd.edu; tokekar@umd.edu).}
\thanks{$^{5}$The author is with the School of Plant and Environmental Sciences, Virginia Tech, Blacksburg, VA 24061, USA (e-mail: bftracy@vt.edu).}}

\begin{document}
\bstctlcite{IEEEexample:BSTcontrol}
\maketitle
\thispagestyle{empty}

\begin{abstract}
	Monitoring the health and vigor of grasslands is vital for informing management decisions to optimize rotational grazing in agriculture applications. To take advantage of forage resources and improve land productivity, we require knowledge of pastureland growth patterns that is simply unavailable at state of the art.  In this paper, we propose to deploy a team of robots to monitor the evolution of an unknown pastureland environment to fulfill the above goal. To monitor such an environment, which usually evolves slowly, we need to design a strategy for rapid assessment of the environment over large areas at a low cost. Thus, we propose an integrated pipeline comprising of data synthesis, deep neural network training and prediction along with a multi-robot deployment algorithm that monitors pasturelands \emph{intermittently}. Specifically, using expert-informed agricultural data coupled with novel data synthesis in ROS Gazebo, we first propose a new neural network architecture to learn the spatiotemporal dynamics of the environment. Such predictions help us to understand pastureland growth patterns on \emph{large scales} and make appropriate monitoring decisions for the future. Based on our predictions, we then design an intermittent multi-robot deployment policy for low-cost monitoring. Finally, we compare the proposed pipeline with other methods, from data synthesis to prediction and planning, to corroborate our pipeline's performance.
\end{abstract}

\def\abstractname{Note to Practitioners}
\begin{abstract}
	Pasturelands are an integral part of agricultural production in the United States. To take full advantage of the forage resource and avoid environmental degradation, pastureland must be managed optimally. This paper focuses on the question of how to deploy robot teams to sense and model physical processes over varying timescales. The goal of this work is to develop a new integrated pipeline for long-term deployment of heterogeneous robot teams grounded in the problem of autonomous monitoring in precision grazing to improve land productivity. By using the proposed pipeline in grassland ecosystem management, we will have a better understanding of the physical environment while respecting energy budgets.
\end{abstract}


\section{Introduction}

Grasslands provide many ecosystem services such as livestock production, wildlife habitat, water infiltration, and carbon sequestration \cite{Bengtsson_Bullock_Egoh_Everson_Everson_O_Connor_O_Farrell_Smith_Lindborg_2019}. Consequently, monitoring the health and vigor of grasslands is vital for informing management decisions to protect or optimize these ecosystem services \cite{Kallenbach_2015}. Sward or canopy height data provide valuable insights into the productivity, habitat value, or maturity of grasslands. When sward height data are monitored over time, changes in sward height can indicate whether a grassland is deteriorating, maintaining its vigor, or becoming more productive. In agricultural systems, monitoring height data can inform decisions about the appropriate timing and intensity of grazing in order to meet economic and ecological goals.

Traditional methods of measuring aboveground height or aboveground biomass rely on labor-intensive methods \cite{Tay_Erfmeier_Kalwij_2018}. For height, this necessitates measuring canopy height by hand using a meter stick or Robel pole \cite{Harmoney_Moore_George_Brummer_Russell_1997}. Aboveground biomass measurements often consist of destructive harvest of forage. Samples are usually cut by hand from a quadrat or frame, bagged, then dried in a forced-air forage oven until reaching a constant weight, known as the dry matter. Depending on the size of the pasture or scope of the monitoring project, height and biomass sampling may involve dozens or hundreds of such samples to ensure that the collected data represents the entirety of the pasture or landscape being monitored. 
Advancements in proximal sensing technologies can provide accurate measurements of height and biomass predictions faster and over larger areas than these labor-intensive traditional methods. However, regular field measurements from pastures through remote sensing methods such as Unmanned Aerial Vehicles (UAVs) are constrained by multiple factors such as limited spatial coverage and low frequency of UAV deployments. Moreover, adverse weather conditions and other resource constraints also contribute to limited deployment and consequently insufficient field measurements of the pasture required to plan grazing activities. In spite of these limitations, precision agriculture involving the use of UAVs to monitor the growth of crops is a promising approach to covering large areas in reasonable times. Analyzing point clouds obtained from the LiDAR attached to UAVs can help to determine the spatial distribution of plants, growth in different regions of the farm and consequently help the farmer focus their resources in regions where they are required the most. Developing prediction models for pastures that work within the limitation of UAV deployment schedules is a promising approach and helps alleviate resource-limited field measurements.  Additionally, motivation for this methodology can be found in related applications such as lost-person search and rescue, where powerful predictive models and UAV teams have shown great promise in large-scale efforts \cite{Heintzman2021-mw, Williams2020-sv}.  Overall, we argue that an intermittent robot deployment scheme coupled with a tightly integrated prediction and planning model would generate maximum efficiency for livestock grazing and pasture recovery processes.

\begin{figure}[!tbp]
	\centering
	\includegraphics[width = \imwidth]{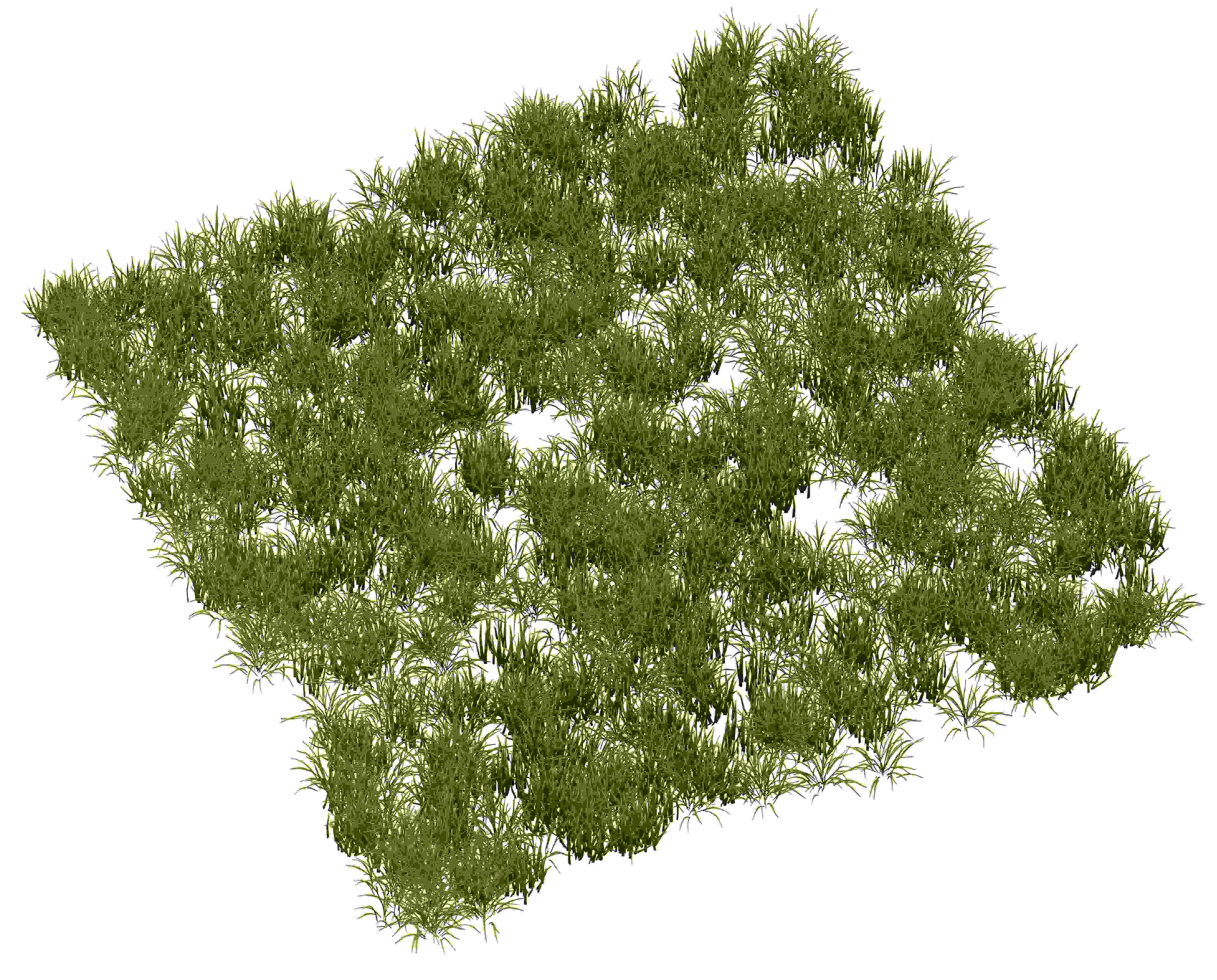}
	\caption{A patch ($2$m $\times 2$m) of the simulated pastureland with a density of $250$ grass models per square meter.}
	\label{fig: patch example}
\end{figure}

With the above motivation, this paper explores the use of autonomous robots to facilitate environmental monitoring for improving land productivity. At a high level, our proposed pipeline works as follows (\figref{fig: pipeline}).  We first use historical data to synthesize a 3D dynamic field to simulate the spatiotemporal environmental process of the site that needs to be monitored (\secref{sec: data synthesis}). For illustration, a small patch ($2$m $\times 2$m) of the generated pasture is shown in \figref{fig: patch example}, where the density is $250$ grass models per square meter. We then use this data to train neural networks to learn the dynamics of this field (\secref{sec: learning and prediction}). Then, an intermittent deployment policy is designed for multi-robot teams (UAVs in our case) using the future pasture height predictions while respecting system budgets (\secref{sec: decision making}). After that, we simulate a pastureland environment in ROS Gazebo \cite{gazebo} to test the performance of the synthesized data (\secref{sec: Gazebo simulation}). Finally, we evaluate the performance of each aspect of our solution and compare it to competing methods (\secref{sec: evaluation}).

In order to synthesize appropriate training data for high-quality predictions, historical data were generated using the expert-informed Agricultural Production Systems sIMulator (APSIM) Next Generation in \secref{sec: data synthesis}. APSIM is designed to model long-term agricultural production in a variety of systems \cite{Holzworth_Huth_deVoil_Zurcher_Herrmann_McLean_Chenu_van_Oosterom_Snow_Murphy_2014}. Using historical meteorological data from three sites in Iowa, APSIM simulated 30 years of tall fescue (Schedonorus arundinacea) pasture dry matter production for each site. Simulated pasture yield was then used to generate average pasture height data based on the equation reported by Schaefer and Lamb \cite{Schaefer_Lamb_2016} describing the relationship between pasture green dry matter and LiDAR-measured pasture height. Based on the historical average pasture height data, we then use a Gaussian mixture model (GMM) to simulate the dynamics of this field over the desired monitoring horizon in \secref{sec: data synthesis}.

Even with a sufficient training data set based on expert-informed historical data, developing an effective prediction model for estimating pasture growth is challenging due to changes incurred by the climate and the spatiotemporal characteristics of the growth. Previous studies for forecasting spatiotemporal dependencies have been based on conventional approaches. These methods require a complex and meticulous simulation of the physical environment for a particular application. Instead, machine learning-based models provide increased flexibility in tackling the difficult spatiotemporal sequence prediction problem that large-scale forage monitoring poses. To address this problem from a deep learning perspective, we use our proposed long short-term memory (LSTM) and convolutional neural network (CNN) based architecture \cite{rangwala2021deeppastl} to model the pasture growth forecasting problem in \secref{sec: learning and prediction}.

Finally, with high-quality pasture growth predictions, a multi-robot deployment policy can be designed to guide future field measurements, as detailed in \secref{sec: decision making}. Our proposed deployment policy aims to maximize the sum of collected information (uncertainty) from the environment while considering waiting penalties and energy constraints of the system. In general, we formulate this deployment problem as a submodular maximization problem \cite{schrijver2003combinatorial} with matroid constraints \cite{oxley2006matroid}. The energy constraints will be formulated as matroid constraints, while the collected information and the waiting penalties will be part of our objective function. Additionally, as the deployment policies gather field measurements, the prediction model can be updated iteratively to generate more accurate estimates of future pasture growth. Through an optimized prediction and deployment model, we show that UAV deployments for pasture monitoring can be scheduled based on the prediction model instead of regular intervals, effectively reducing the required number of deployments. This \textit{need-based intelligent} deployment policy substantially reduces the cost of gathering field measurements and effectively allows resource-constrained enterprises to manage pasture grazing more effectively. 

Beyond the forage monitoring problem, the proposed pipeline can be mapped to other large-scale monitoring problems for sufficiently slow spatiotemporal processes where intermittent deployment is reasonable. That is, the data synthesis -- neural network training/prediction -- intermittent multi-robot deployment pipeline is quite general, given access to expert-informed data as a starting point. For example, in ocean monitoring applications \cite{smith2011persistent,mccammon2021ocean}, we can also use the proposed pipeline to generate and refine deployment policies given the availability of high-quality models of aquatic processes.

\emph{Contributions:}
In summary, the contributions of this work are as follows:
\begin{itemize}
	\item We propose an integrated pipeline\footnote{\href{https://github.com/precision-grazing/project}{https://github.com/precision-grazing/project}} to simulate and solve an important problem in the precision agricultural space, forage monitoring, including data synthesis, prediction, planning, and perception (\secref{sec: preliminaries and problem formulation}).
	\item We demonstrate how to exploit historical agricultural data to synthesize a pastureland environment in a manner that accommodates both neural network training and rapid prototyping in a simulation environment (\secref{sec: data synthesis}).
	\item We demonstrate a scalable spatiotemporal learning architecture that can be used to integrate with an intermittent multi-robot planning strategy (\secref{sec: learning and prediction}).
	\item We introduce a novel intermittent deployment policy that integrates neural network-based predictions while considering budgets to deploy robots for autonomous environmental monitoring (\secref{sec: decision making}).
	\item We have built (and made available) a high-fidelity pastureland simulation environment in ROS Gazebo, allowing for rapid prototyping of pasture monitoring with LiDAR point clouds (\secref{sec: Gazebo simulation}).
\end{itemize}


\section{Related work}
\label{sec: related work}

Predictive modeling of agricultural systems can provide helpful information on their long-term resilience and productivity. Predictive models have been developed for many crops and regions, tested extensively, and refined for increased predictive accuracy for commodities such as corn, soybeans, wheat and rice, to name a few.  As model accuracy increases, it can provide insights into crop responses to abiotic stressors such as climate change \cite{Asseng_Ewert_Rosenzweig_Jones_Hatfield_Ruane_Boote_Thorburn_Rotter_Cammarano_et_al_2013,Li_Newton_Lieffering_2013}, or identify optimal crop rotations between species over a period of years \cite{Balboa_Archontoulis_Salvagiotti_Garcia_Stewart_Francisco_Prasad_Ciampitti_2019}.
However, agricultural modeling of forage systems presents unique challenges compared to row cropping systems, as forage systems are often perennial rather than annual, the forage crop can be ``harvested'' multiple times per year by grazing livestock or cutting for hay, and pastures are typically multispecies ecosystems rather than monospecific crops with homogeneous phenotypes and physiology \cite{Jones_Jones_McDonald_1995}.
Several forage modeling programs have been developed despite the aforementioned challenges. These include—but are not limited to—the Simulation of Production and Utilization on Rangelands (SPUR) first developed on rangelands in the 1980s \cite{Corson_Skinner_Rotz_2006}; the GRAzing SImulation Model (GRASIM) developed in the late 1990s \cite{Mohtar_Zhai_Chen_2000}; the Dairy Forage System Model (DAFOSYM) developed in the late 1980s and updated into the Integrated Farm System Model (IFSM) \cite{Rotz_Gupta_1996}; and the Agricultural Production Systems sIMulator (APSIM) Next Generation, which can model both row-cropping systems and pastures \cite{Holzworth_Huth_Fainges_Brown_Zurcher_Cichota_Verrall_Herrmann_Zheng_Snow_2018}. We select APSIM as the optimal modeling program for our research, as the program has been used previously for pasture modeling in a variety of contexts \cite{Ahmed_Parsons_Morel_Uttam_Sandstrom_Lanna_Wallsten_2020,Bosi_Sentelhas_Huth_Pezzopane_Andreucci_Santos_2020}; provides daily timestep outputs which fulfilled our need for fine-grained temporal data; and the program's modular nature allows for rapid customization of input parameters \cite{Li_Snow_Holzworth_Johnson_2010,Li_Newton_Lieffering_2013}.

The problem of predicting an evolving pastureland over time may be solved with conventional methods such as Gaussian processes (GPs) \cite{williams2006gaussian}, which focus on stochastic Gaussian processes to model the regression for different pasture heights or observations. This method is also known as Kriging \cite{williams2006gaussian}. GPs are non-parametric methods by defining the internal relations between observations \cite{liu2020data}. Similarly, Gaussian Markov Random Fields (GMRFs) are often used to model spatial environmental fields \cite{williams2006gaussian}. These conventional methods are suitable for problems where there is significant prior knowledge of the environment that needs to be monitored. The historical data used in this paper can give us a general trend of the average height change of the field. However, as we must make predictions on large scales, this prior is insufficient for us to generate a reasonable conventional model for the entire pastureland environment (not to mention the issues with computational scaling). Moreover, as the growth patterns may differ across the field, conventional models are not as flexible for modeling the heterogeneity of growth patterns spatially and temporally. We, therefore, sought to apply a neural network-based method to tackle this prediction problem when we have a large dataset for modeling the heterogeneity of the environment.

Specifically, we implement a framework that integrates a neural network-based encoder-decoder architecture to learn the historical data's underlying patterns over time for future predictions. The problem of predicting future pasture heights is analogous to video frame prediction \cite{oliu2018folded}, with the key challenge in our case lying in predicting the growth of pasture surfaces. In the deep learning-based prediction domain, sequence-to-sequence problems were originally introduced through recurrent neural networks (RNN) and long short-term memory (LSTM) models and provide a baseline for solving temporal forecasting problems that we encounter in this work \cite{hochreiter1997long, graves2013generating, rangwala2020}. To incorporate spatial features, prior works have generally used multiple architectures to consider spatial and temporal features separately by combining the autoencoder or a generative adversarial network (GAN) model with an RNN model \cite{oliu2018folded, michalski2014modeling, srivastava2015unsupervised}. Convolutional LSTMs (ConvLSTMs) \cite{xingjian2015convolutional} have been successfully used for spatiotemporal predictive learning and were originally proposed for precipitation nowcasting over radar images. Recent works have used the architecture for learning frame representations \cite{lotter2016deep}. Motion-content network (MCNet) \cite{villegas2017decomposing} uses an additional LSTM apart from the image encoder to model the motion dynamics. The outputs of both the encoders are combined and fed to the decoder to predict frames. Recently, \cite{wang2017predrnn, wang2018predrnn++} proposed ST-LSTM to learn structural information for spatiotemporal sequences and a new model structure PredRNN and PredRNN++ to explicitly decouple memory cells and improve the cross-layer interaction of memory states across different LSTMs. These methods help improve stronger spatial correlation and short-term dynamics for powerful modeling and prediction capabilities. We utilize the recent developments in ConvLSTM applications and propose a novel prediction architecture that is particularly effective in predicting the rapidly changing dynamics of the pasture over long horizons. We achieve this through the use of recurrent encoder-decoder networks based on ConvLSTMs over different resolutions of the pasture to effectively capture different features and dependencies of the pasture dynamics. Moreover, by using appropriate processing described in \secref{sec: learning and prediction} over raw pasture data, our pasture terrain forecasting technique can scale to any terrain size.

Standard learning-based models do not capture model uncertainty. Unlike regression problems where a model can output a predictive probability, as we deal with a sequence-to-sequence prediction problem, we need to extrapolate prediction uncertainty from training data. Bayesian probability theory has been used in deep learning as a tool to deal with uncertainty \cite{neal2012bayesian, mackay1992practical}, where the weights of the neural network are defined as distributions. Bayesian neural networks (BNN) are more robust to over-fitting. However, they add significant computation complexity. Sampling-based and stochastic variational inference methods have been used to approximate Bayesian neural networks \cite{paisley2012variational, kingma2013auto, rezende2014stochastic, hoffman2013stochastic}. Similar to BNN, these methods incur high computational costs without additional benefits to improving accuracy. An alternative approach \cite{gal2016dropout} with minimal computational and model complexity in deep learning models was proposed through the use of Dropouts \cite{baldi2013understanding}. The authors show that any neural network with arbitrary depth and non-linearities, modeled with dropouts behind every layer, is equivalent to the approximation of a probabilistic deep Gaussian process \cite{damianou2013deep}. In our previous work \cite{rangwala2021deeppastl}, we integrate this concept of uncertainty estimation from dropout over our architecture to provide the necessary uncertainty maps for multi-robot monitoring from a typical machines learning perspective. Whereas in this work, our focus is about integrating a combinatorial optimization based planner with proposed deep learning based predictions \cite{rangwala2021deeppastl} and performing a deep evaluation of the entire pipeline with realistic simulations.
From the multi-robot deployment perspective, to estimate the evolving processes of pastureland environments more efficiently, we cannot deploy robots (UAVs) to collect observations frequently because it is not energy efficient. Those types of observations can be point clouds, heightmaps, sonar data, etc. Meanwhile, the deployment strategy should be generated based on environmental information instead of being artificially defined. This intermittent idea can also be found in other robotics applications. In \cite{hollinger2012multirobot}, the robots in a team are designed to communicate intermittently while working together to explore an environment. In \cite{gini2017multi}, the authors use time windows to model the availability of robots at different times in task allocation applications. Therefore, robots are not required to work continuously. The idea of intermittence is contrary to that of persistence, where robots are required to work continuously to fulfill different tasks \cite{lan2013planning,tokekar2015visibility}. In \cite{li2021attention}, the authors proposed a deep learning-based method for combining environmental prediction and path planning, where the spatial path planning is the main concern in the paper.
In \cite{hollinger2014sampling}, the authors investigated a sampling-based path planning method for variance reducing. Similarly, in \cite{nguyen2015information}, the author used an adaptive sampling strategy for reducing entropy generated from GP modeling. In \cite{xu2013efficient}, a GMRF-based method was proposed for spatial predictions.

Since the monitored environment evolves spatially and temporally, we need to utilize the predicted environmental information to make a spatiotemporal deployment plan. In \cite{liu2018optimal}, we use a partially observable Markov decision process (POMDP) to model the dynamics of an environmental process. Then, a submodular objective function is applied to model the false alarm and delay cost. This method works only when the process models are known and can be modeled as POMDPs. In \cite{liu2019submodular,liu2020monitoring}, we use non-parametric Gaussian processes \cite{williams2006gaussian} to model the dynamics of a monitored environmental process and then use mutual information as a metric to guide our deployments. Meanwhile, matroids are used to model the budget constants. Generally speaking, matroids \cite{schrijver2003combinatorial,oxley2006matroid} can be used to model the independence in constraints, which can be found in many robotics applications, e.g., task allocation \cite{williams2017decentralized}, multi-robot deployment \cite{liu2020coupled}, topology selection and planning \cite{Heintzman2021-qr}, and probabilistic security/resilience in multi-robot systems \cite{wehbe2020optimizing,Wehbe2021-vx}, etc. We will also use this tool to model the independence of different constraints in the deployment strategy in this work. In general, the environmental modeling methods used in our previous works are conventional GP-based methods, where a prior of the entire environment is needed to utilize historical data. While in this paper, the environment modeling method is a data-driven approach, which builds the dynamics of the environment through a historical dataset where an exact process model is not required. Moreover, the environment modeling and the deployment policy generation are highly connected in this paper. Finally, as the deployment policies gather more measurements, the proposed data-driven prediction model can be updated accordingly for better future predictions.


\section{Preliminaries and Problem Formulation}
\label{sec: preliminaries and problem formulation}

\subsection{Preliminaries}
\label{ssec: preliminaries}

A core aspect of this paper is to optimize multi-robot deployment plans by utilizing historical agricultural data.  We propose to tackle this problem from a combinatorial optimization perspective. To this end, we begin by reviewing the concepts related to our objective function and constraint modeling.

A set function $f: 2^\V \mapsto \RR$ is a function that maps any set $\A \subseteq \V$ into $\RR$, where $\V$ is the finite discrete ground set of $f(\cdot)$.

\begin{definition}[\cite{schrijver2003combinatorial}]
	A set function $f:2^\V \mapsto \RR$ is
	\begin{itemize}
		\item normalized, if $f(\emptyset) = 0$;
		\item monotone non-decreasing, if $f(\A) \le f(\B)$ when $\A, \B \subseteq \V$, and $\A \subseteq \B$;
		\item submodular, if $f(\A \cup \{v\}) - f(\A) \ge f(\B \cup \{v\}) - f(\B)$ when $\A, \B \subseteq \V$, $\A \subseteq \B$, and $v \in \V \setminus \B$.
	\end{itemize}
\end{definition}

The property of \emph{submodularity} is often described as having a diminishing returns property (as the above definition suggests), making it a natural option for modeling objective functions in robotics (e.g., sensor placement \cite{krause2008near}, set coverage \cite{schrijver2003combinatorial}, task allocation \cite{choi2009consensus}).  It is also convenient to work with the \emph{marginal return} when an element $v$ is added to $\A$, which is defined as $f(\{v\} \mid \A) = f\left(\A \cup \{v\}\right) - f(\A)$.  The submodularity level of a set function can be measured using curvature, a property of the set function itself. Curvature can be defined as follows.

\begin{definition}[\cite{conforti1984submodular}]
	Let $f: 2^\V \mapsto \RR$ be a monotone non-decreasing submodular function, we define the curvature of $f(\cdot)$ as
	\begin{equation*}
		c_f = \max_{v \in \V} \frac{f(\V) - f\left(\V \setminus \{v\}\right)}{f(v)},
	\end{equation*}
	where $\V$ is the ground set.
	\label{def: curvature}
\end{definition}

It holds that $0 \le c_f \le 1$. If $c_f = 0$, then $f(\cdot)$ is a modular function and $f\left(\A \cup \{v\}\right) - f(\A) = f(v), \forall \A \subseteq \V, v \in \V \setminus \A$ when $\V$ is the ground set. If $c_f = 1$, then $f\left(\A \cup \{v\}\right) - f(\A) = 0$, where $\A \subseteq \V$ and $v \in \V \setminus \A$. This means adding $v$ has no contribution to the function value $f(\cdot)$ if set $\A$ is selected.

Next, we introduce the related concept of constraint modeling. Matroids generalize the idea of independence in set systems. Meanwhile, efficient sub-optimal solutions can be found when constraints are modeled by matroids \cite{fisher1978analysis}.

\begin{definition}[\cite{oxley2006matroid}]
	A matroid $\M = (\V, \I)$ is a set system that contains a finite ground set $\V$ and a collection $\I$ of subsets of $\V$ with the following properties:
	\begin{enumerate}
		\item $\emptyset \in \I$;
		\item If $\A \subseteq \B \in \I$, then $\A \in \I$;
		\item If $\A, \B \in \I$ and $|\B| < |\A|$, there exists a $v \in \A \setminus \B$ such that $\B \cup \{v\} \in \I$.
	\end{enumerate}
\end{definition}

The intersection of $L$ matroids can be written as $\M = (\V, \I)$, where $\M_i = (\V, \I_i)$ is the $i$th matroid and $\I = \bigcap_{i=1}^L \I_i$. The cardinality of this matroid intersection constraint is $|\M| = L$. A set belonging to a matroid means that this set should satisfy the matroid conditions. Examples of matroid constraint modeling in robotics can be found in task allocation \cite{williams2017decentralized}, orienteering \cite{jorgensen2017matroid}, action selection \cite{liu2021distributed}, etc.


\subsection{Problem Formulation}
\label{ssec: problem formulation}

Consider a spatiotemporal forage process evolving in a 2D pasture environment (which we model rigorously in the sequel). In simple terms, we are interested in determining forage height for any location in the pasture environment over a long time horizon by deploying multi-robot teams. To fulfill this goal, we divide our efforts into two tasks: forage process prediction and multi-robot planning.  In the following, we outline the fundamentals of these tasks and conclude with a concrete problem statement.


\paragraph{Prediction} Assume that we are given historical 2D heightmaps $\XX_t \in \RR^{M \times N}$ of a pasture field at different times, where $t$ is the associated time index, and $M, N$ are the width and the length of the pasture heightmap, which correspond to the discretization resolution. This resolution will be used for both the prediction problem and the planning problem. Denote by $\X = \{\XX_t \in \RR^{M \times N} \mid t \in \T_x\}$ the historical dataset containing all the historical measurements, where $\T_x$ contains all the time indexes associated with each $\XX_t \in \X$. The extracted height of the pasture in location $(x,y)$ with respect to $\XX_t$ is denoted by $\XX_t(x,y)$. Our first goal is to train a neural network to predict future pasture heights $\mathbf{\bar{Y}}_t$ for the prediction horizon $\T_y$ using the historical dataset $\X$. This prediction process is modeled as
\begin{equation}
	    (\bar\YY_t, \bar\sigma^2_t) \leftarrow \Theta (\X, \WW), \quad \forall t \in \T_y,
\end{equation}
where $\Theta$ is our neural network model, $\WW$ is the set of parameters of the model, $\bar\YY_t \in \RR^{M \times N}$ is the predicted heightmap at $t$, and $\bar\sigma^2_t \in \RR^{M \times N}$ is the corresponding variance at $t$.
Also, we denote by $\bar\Y= \left\{\bar\YY_t \in \RR^{M \times N} \mid \forall t \in \T_y\right\}$ the prediction set that contains all of the predicted heightmaps for the prediction horizon $\T_y$, and denote by $\bar\Sigma= \left\{\bar\sigma^2_t \in \RR^{M \times N} \mid \forall t \in \T_y\right\}$ the corresponding variance set. The details will be specified in \secref{sec: learning and prediction}.


\begin{figure*}[!t]
	\centering
	\includegraphics[width=\textwidth]{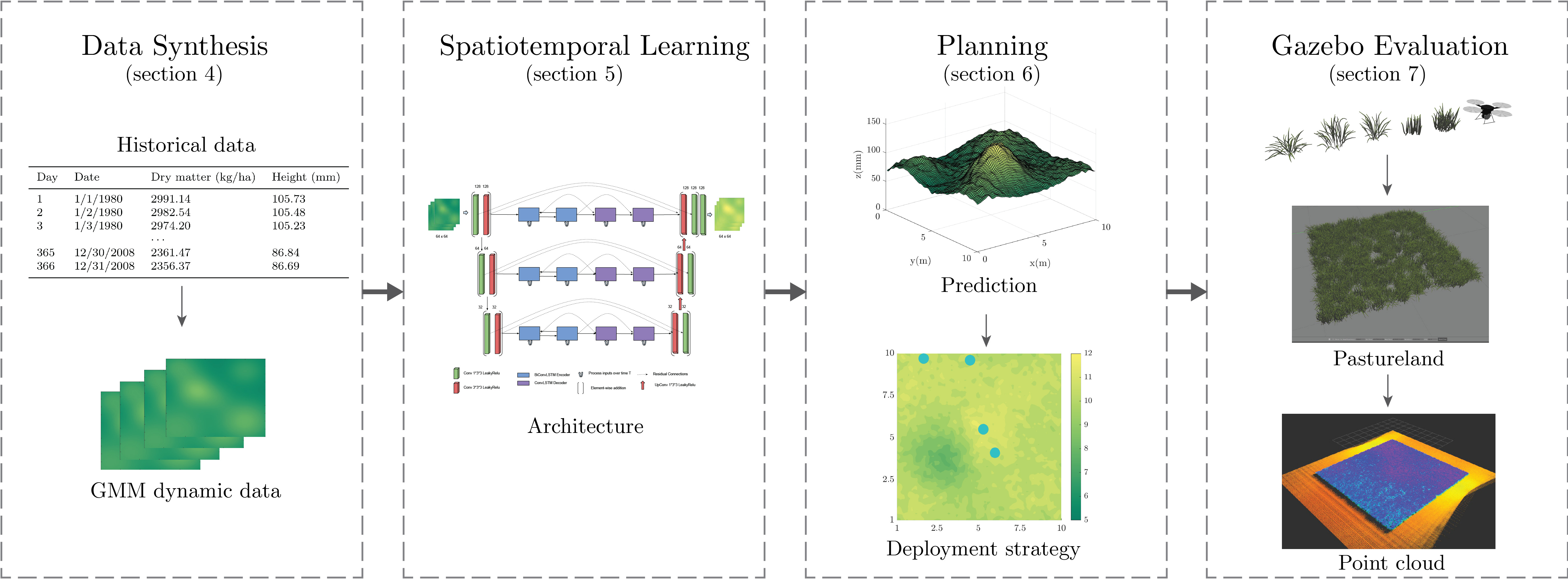}
	\caption{A diagram of our overarching solution. Data synthesis is first used to simulate historical pasture data. Then spatiotemporal learning is used for neural network training and prediction based on historical data.  Next, the planning aspect is used to make multi-robot deployment decisions for collecting new data.  Finally, on the far right we illustrate the high-fidelity simulation we have built for evaluating our pipeline.}
	\label{fig: pipeline}
\end{figure*}

\paragraph{Planning} Based on the predicted heightmaps $\bar\Y$ and variance maps $\bar\Sigma$, we seek a multi-robot deployment strategy to collect data, reinforce our predictions, and ultimately make better pastureland management decisions. To determine a deployment strategy, we first need to build the ground set $\V$, which contains all possible deployment decisions for robots over space and time. Consider a team of robots available to deploy over the time horizon $\T_y$, the prediction horizon.
The family of all available locations is denoted by $\P$. Thus, we have $(x, y) \in \P$, and $|\P| = M \cdot N$ is the cardinality of the deployable locations. We also denote by $\R$ the set of indices of all robots. Each robot $r \in \R$ may have a different sensing ability, i.e., sensing noise. Therefore, the ground set $\V$ at time $t$, containing all available deployment choices, is as follows:
\begin{equation*}
	\V_t = \left\{\left(x, y, r, t\right): \forall (x,y) \in \P, r \in \R, t \in \T_y \right\}.
\end{equation*}
We interpret $(x, y, r, t)$ as ``location $(x, y)$ is sensed by robot $r$ at time $t$''. To simplify the notation, we will use a 4-tuple $v \coloneqq (x,y,r,t)$ to denote the deployment decision factor in the following. Finally, the ground set of the deployment problem is $\V = \bigcup_{t \in \T_y} \V_t$. The cardinality of the ground set is $|\V| = |\P| \cdot |\R| \cdot |\T_y|$. The associated predicted variance with respect to $v$ at $t$ is represented by $\bar\sigma^2_t(v), \forall v \in \V$. Since $\sigma^2_t(v)$ is the predicted variance map at time $t$ for $v$, we can use $\bar\sigma^2_t(x,y) \in \RR$ to denote the predicted variance that corresponds to location $(x,y) \in \P$ at time $t \in \T_y$. We want to determine a deployment policy set $\S \in \V$ to maximize the information we can get from the environment while respecting the system budgets. To this end, we denote by $f: 2^\V \mapsto \RR$ the objective function, and denote by $\M$ a set of sets defining the system's admissible deployment policies. The details will be specified in \secref{sec: decision making}.

\paragraph{Problem Statement} With the basics of prediction and planning outlined, we now formalize the problem we solve in this paper.

\begin{problem}[Intermittent Deployment]
\label{prb: problem formulation}
Consider a historical set $\X = \left\{ \XX_t \in \RR^{M \times N} \mid \forall t \in \T_x\right\}$, containing heightmaps of a discretized $M \times N$ pasture over a time horizon $\T_x$. Let $\V$ be the ground set containing all possible multi-robot deployment factors $v \coloneqq (x,y,r,t)$ over a time horizon $\T_y$.  Assume further the existence of a predicted variance set $\bar\Sigma= \left\{\bar\sigma^2_t \in \RR^{M \times N} \mid \forall t \in \T_y \right\}$.  The intermittent deployment problem is maximize a set function $f(\cdot)$ by selecting appropriate deployment factors while respecting budget constraints. That is,
\begin{equation*}
	\begin{split}
		\underset{\S \subseteq \V}{\text{maximize}} \quad & f(\S) \\
		\text{subject to}\quad & \S \in \M.
	\end{split}
	\label{eq: problem formulation}
\end{equation*}
\end{problem}
By selecting the set $\S$ that maximizes the objective function, we can determine all deployment factors containing deployment locations and times to conduct deployments while maintaining efficiency.

\begin{remark}
	This work focuses on predictions and planning perspectives, especially in spatiotemporal deployment locations selection under limited energy conditions. We should note that a path planning problem is a subsequent problem of our framework, and any of them can be integrated into our framework. Therefore, we focus on prediction and planning to set up a cornerstone for the subsequent problems.
\end{remark}

The diagram of our proposed solution to the above problem is shown in \figref{fig: pipeline}. Our pipeline can be divided into the following aspects, which we detail in the sequel: data synthesis (\secref{sec: data synthesis}), deep learning prediction (\secref{sec: learning and prediction}), planning (\secref{sec: decision making}), and Gazebo evaluation (\secref{sec: Gazebo simulation}).


\section{Large-Scale Pasture Environment Synthesis}
\label{sec: data synthesis}

In this section, we focus on pasture environment synthesis. This process will be divided into two parts. First, we illustrate how to synthesize historical average pasture height data (\secref{ssec: historical data preparation}). Based on this data, we then introduce how to utilize this data to create a dynamic pasture environment (\secref{ssec: GMM}).


\subsection{Historical Data Preparation}
\label{ssec: historical data preparation}

`Historic' pasture data were generated using APSIM Next Generation's publicly available meteorological, soils, and pasture species modules.  We selected three sites in Iowa due to the availability of meteorological data for each in APSIM's Met module as a result of prior research \cite{Archontoulis_Miguez_Moore_2014}. Meteorological data spanned 1979 to 2013 and included solar radiation (MJ/m$^2$), rain and snowfall, minimum and maximum temperature, atmospheric pressure, and day length. Soils selected in the module were fine-loamy, mixed, superactive, mesic Hapludolls common in Iowa, also available in APSIM's modules \cite{Archontoulis_Miguez_Moore_2014}. APSIM's tall fescue AgPasture module was used for the forage species \cite{Li_Newton_Lieffering_2013}. Tall fescue was set to 1m rooting depth and initial belowground and aboveground biomass of $1000$ kg/ha and $3000$ kg/ha, respectively. The SoilOM module, which simulates soil organic matter processes, was set to 1000 kg/ha initial surface residue. Fertilizer application was simulated at 84 kg N/ha on January 1 and another 84 kg N/ha on August 15 each year in the form of nitrate (NO$_3$-N). The resulting simulated pasture yield was then used to generate average pasture height data using an equation reported by Schaefer and Lamb \cite{Schaefer_Lamb_2016} describing the relationship between LiDAR-measured pasture height and pasture green dry matter, i.e., green aboveground biomass.  In this paper, we denote by $\h \in \RR^{T_x}$ the historical data, where $T_x = |\T_x|$ is the length of the historical dataset horizon $\T_x$, which is also defined in the problem formulation in \secref{ssec: problem formulation}. We also denote by $\h_t \in \RR$ the average pasture height data at time $t$ in the historical dataset. Further information on the above APSIM modules' functions and processes may be found in the work of Li and colleagues \cite{Li_Newton_Lieffering_2013}.


\subsection{Pasture Environment Data Synthesis}
\label{ssec: GMM}

The generated average pasture height data is temporal data. When making predictions for different locations and developing spatiotemporal deployment strategies, we need both spatial and temporal historical data to know the different growth patterns in different places of the pasture field. Therefore, this section focuses on synthesizing spatiotemporal data from the historical average pasture height data. In general, this process fits a spatiotemporal process to the temporal aspects of the historical data.

In this work, we use a dynamic Gaussian mixture model (GMM), a combination of Gaussian distributions, to simulate the pasture evolution using historical data. The GMM used in this work is a discrete model as we need to make predictions for different days and make corresponding deployment decisions for different times and locations. We should note that there are many other ways to simulate a spatiotemporal process by using temporal data only, and we select GMM for the following reasons. GMM is a common tractable process representation, which allows us to adjust the model to match the actual field evolving process. From expert inputs or manual field measurements, we can choose reasonable spatial parameters for the GMM to complement the temporal parameters coming from historical data, so the spatiotemporal model represents the pasture evolving process.

In general, we first use $B$ components to build a dynamic GMM and then use the historical data $\h$ to adjust the generated model to match the historical information. The dynamic GMM is model as
\begin{equation}
	\GG_t(x,y) = \sum_{i=1}^B w_i(t) b_i(x,y) = \w(t)^\top \b(x,y),
	\label{eq: GMM}
\end{equation}
where $(x,y) \in \RR^2$ is a 2D location from the location set $\P$, basis $b_i(x,y) \in \mathbb R$ is the output of $i$th basis function in the location $(x,y)$, $B$ is the number of basis functions, and $\GG_t(x,y)$ is the output of the GMM at time $t$ and location $(x,y)$. The weight $w_i(t) \in \mathbb R$ is associated with $b_i(x,y)$, where $t \in \T_x$. The $i$th basis function $b_i(x,y)$ is a Gaussian kernel function. That is,
\begin{equation}
	b_i(x,y) = \exp[-\frac{[(x,y) - (b_{xi}, b_{yi})]^2}{2 c_i^2}],
	\label{eq: Gaussian kernel}
\end{equation}
where $(b_{xi}, b_{yi}) \in \RR^2$ is the 2D position of $i$th basis function $b_i(x,y)$, and $c_i$ is the corresponding length-scale. Also, $\w(t) \in \RR^B$ is the stacked weights at time $t$ and $\b(x,y) \in \RR^B$ is the stacked basis functions for the location $(x,y)$. One way to generate a dynamic 3D smooth surface to simulate pasture is to change the weights of different basis functions smoothly at different times. To achieve this goal, we use a 1D Gaussian process \cite{hennig2013animating} to model the change of each weight at different times for the entire time horizon $\T_x$. After this step, a dynamic GMM is fully built.

\begin{figure*}[!t]
	\centering
	\includegraphics[width=0.85\textwidth]{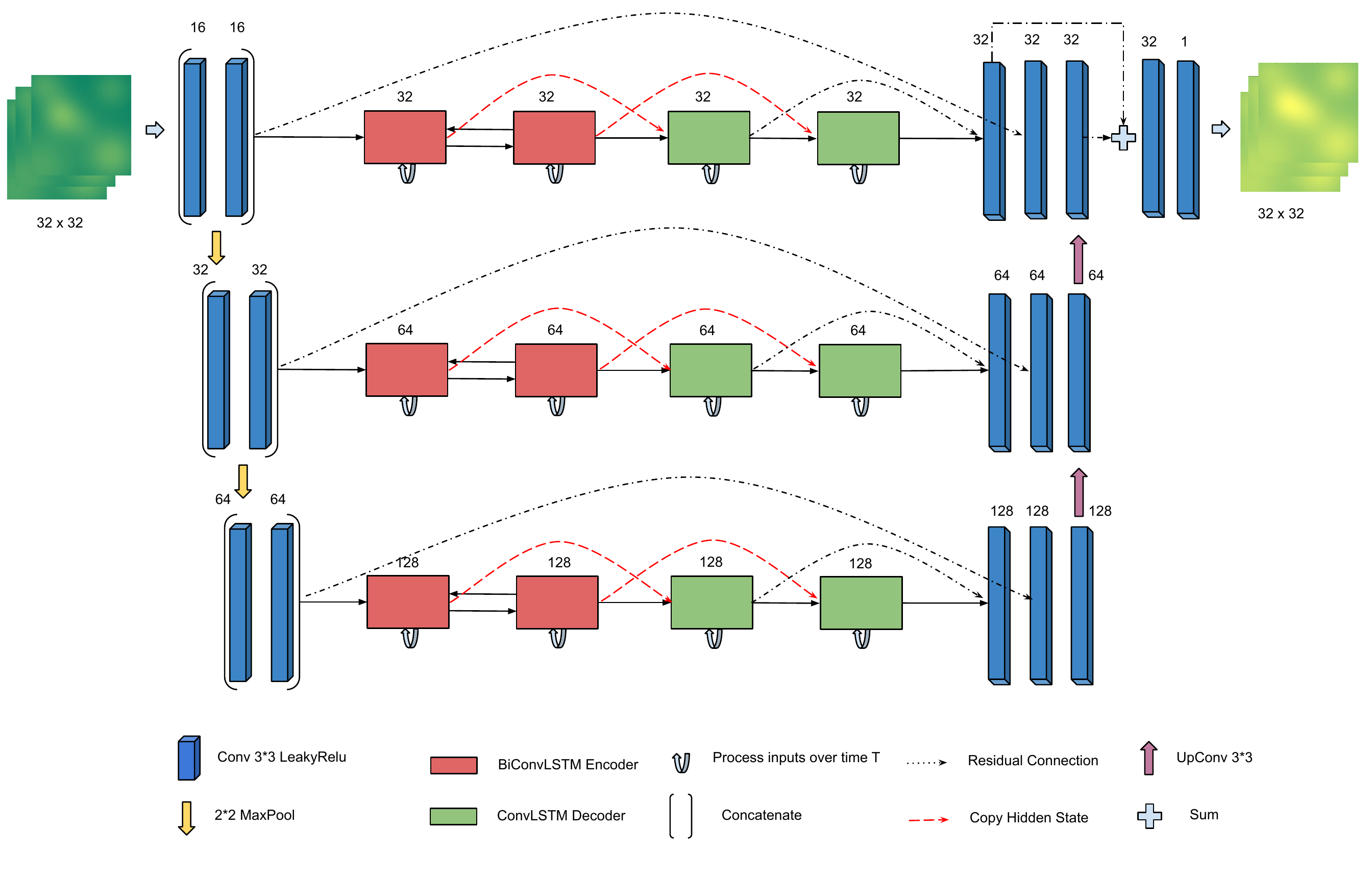}
	\caption{Prediction neural network architecture is defined as an encoder-decoder. Initial feature maps across different resolutions are generated using Conv2D encoders and then are iteratively fed to the ConvLSTM encoders to encode the input observations. The ConvLSTM decoders then recursively generate representations for each interval in the output. These recurrent hidden representations across resolutions are then merged to generate the final sequence of prediction. Residual connections are used to ease regeneration of outputs from the generated hidden representations.}
	\label{fig: DNN Architecture}
\end{figure*}

Next, we need to adjust this model to make its temporal property match the historical average height data. Specifically, we need to adjust the generated GMM surfaces at different times using the historical average height data. That is:
\begin{equation*}
	\XX_t(x,y) \leftarrow \GG_t(x,y) + \h_t - \bar\GG_t,
\end{equation*}
where $\h_t \in \RR$ is the historical average height at time $t$, and $\bar\GG_t \in \RR$ is the simulated average height of the field at that time, which is calculated by using $\bar\GG_t = |\P|^{-1} \sum_{(x,y) \in \P} \GG_t(x,y)$.
After this adjustment, the averaged height of the generated surfaces at different times meets the historical data. Meanwhile, those surfaces' heights, i.e., $\XX_t(x,y)$, are truncated if the height is negative. Finally, small noise is added to the surfaces to simulate a diversified growth pattern in the pasture. After this series of operations, the spatiotemporal dataset $\X = \{\XX_t \in \RR^{M \times N} \mid t \in \T_x\}$ is fully constructed and can be used in the later learning and predictions.

To summarize the data synthesis process, we first synthesize the historical average data $\h$. Then, we build a dynamic GMM denoted by $\GG$, and adjust this model to make the temporal property of $\GG$ match the historical average data $\h$. Finally, the adjusted dataset is denoted by $\X$ that will be used in learning and predictions (below).


\section{Spatiotemporal Learning and Prediction}
\label{sec: learning and prediction}

In this section, we will be first focusing on learning the dynamics of the generated field using the generated dataset $\X = \{\XX_t \in \RR^{M \times N} \mid t \in \T_x\}$.
Note that a pre-processing will be applied before feeding $\X$ into our training networks. Then, we will use the learned dynamics to make predictions of the field for the future, which will then serve as inputs for the deployment strategy planning (\secref{sec: decision making}).


In general, our spatiotemporal learning network is modeled as
\begin{equation*}
	    (\bar\YY_t, \bar\sigma^2_t) \leftarrow \Theta (\X, \WW), \quad \forall t \in \T_y,
\end{equation*}
where $\Theta$ is our neural network model, $\WW$ is the set of parameters of the model, $\bar\YY_t \in \RR^{M \times N}$ is the predicted heightmap at $t$, and $\bar\sigma^2_t \in \RR^{M \times N}$ is the corresponding variance at $t$. Next, we give the details of our mean and variance predictions.

Since the prediction model will take as input a sequence of heightmaps to predict multi-step long-horizon future pasture heights, we need to reorganize $\XX_t \in \X, \forall t \in \T_x$ to satisfy this requirement. Meanwhile, we need pre-processing to enable our proposed model to tackle different pasture sizes. We aim to reorganize all $\XX_t \in \X, \forall t \in \T_x$ to form multiple training sequences. We define the $i$th training sequence as $\X_i = \left\{\XX_t \in \RR^{M \times N} \mid \forall t \in \T_i\right\}$ with $\T_i = \{i, i + \delta, \ldots, i + \alpha \delta\}$, where $\delta$ is the number of intervals before each input observation in the training sequence, $\alpha$ is the number of observations in $\X_i$, and $\T_i$ contains all the time index associated with every $\XX_t \in \X_i$. Note that $\T_i \subseteq \T_x$. The complementary training label (ground truth) for an input sequence $\X_i$ is then defined as $\Y_i = \left\{\YY_t \in \RR^{M \times N} \mid \forall t \in \T_y \right\}$, where $\T_y = \{i + (\alpha + 1) \delta, i + (\alpha + 2)\delta, \ldots, i + (2\alpha + 1)\delta\}$ contains the expected prediction step at $\delta$ intervals. Therefore, $\alpha = |\T_i| = |\T_y|$. The effective length (horizon) of the input and output sequences are then calculated as $L = \delta \cdot \alpha$. Varying the number of strides $\delta$ in the sequences allows our prediction model to work with both short term and long term horizon. For example, given a input sequence of $\alpha = 15$ and a stride of $\delta=4$, we have an effective observation length of $60$ days.  Finally, the neural network outputs a prediction sequence$\bar{\Y}_i = \{\bar{\YY}_t \in \RR^{M \times N} \mid \forall t \in \T_y \}$ for each input sequence $\X_i$, where $\bar\YY_t$ is the prediction with respect to the ground truth $\YY_t$ at $t$.

\begin{remark}
	For the scope of this paper, we train our networks using the same input and output observation interval $\delta$ and the same number of observation/measurements $\alpha$. That is, $\alpha = |\T_i| = |\T_y|$. However, it is to be noted that the proposed network can be trained on varying sequence lengths without changing the architectural design. This is due to the inherent flexibility of an encoder-decoder design.
\end{remark}

The high-level idea of the mean predictions is as follows. For each time $t$, we will use the averaged prediction $\bar\YY_t$ from several Monte Carlo (MC) predictions $\hat\YY_t^{(k)} \in \RR^{M \times N}$ as the final predicted heightmap at $t$, where $\hat\YY_t^{(k)}$ is the $k$th MC prediction at time $t$. That is, $\hat\YY_t^{(k)} \leftarrow \Theta (\X, \WW), \forall t \in \T_y$.

The averaged mean prediction is given by
\begin{equation}
    \bar\YY_t \coloneqq \EE_{\Pr(\bar\YY_t \mid \X)}(\bar\YY_t) \approx \frac{1}{K} \sum_{i=1}^{K} \hat\YY_t^{(k)}, \quad \forall t \in \T_y.
    \label{eq: prediction mean}
\end{equation}
Each MC prediction is a realization of the proposed network by using a different setting, which will be specified later, and the number of sampled predictions is $K$ for each heightmap prediction $\bar\YY_t, \forall t \in \T_y$.

Our network architecture is shown in \figref{fig: DNN Architecture}. We adapt the ubiquitous U-Net Convolution Neural Network framework \cite{ronneberger2015u} with ConvLSTM cells as an encoder-decoder framework for making sequence predictions. We introduce a novel architecture, where the model learns spatiotemporal dependencies for long-horizon predictions using a mixture of Convolutional Neural Networks (CNN) and ConvLSTM encoder-decoder layers.

We can now define the process that allows the architecture to capture spatiotemporal dependencies. As a first step, we define the ConvLSTM architecture \cite{xingjian2015convolutional} that forms a building block for our model. ConvLSTMs can be seen as a special case of LSTMs, by replacing the Hadamard product in LSTMs with convolution operators. The ConvLSTM block enables the network to recursively process a sequence of representations and update its hidden states $\HHH_t$ that encode the complete spatiotemporal representations for all $\X_i$'s. The key idea behind ConvLSTMs is its capability to \textit{retain} information that is relevant to its prediction task, and \textit{forget} information over time that might be repetitive or not necessary over the sequence. The operations within the ConvLSTM block is shown as follows:

\begin{equation*}
	\begin{split}
		\II_t & = \varphi(\WW_{xi}\ast \XXX_t^c + \WW_{hi}\ast \HHH_{t-1} + \WW_{ci}\circ \CCC_{t-1} + \WW_i),          \\
		\FF_t & = \varphi(\WW_{xf}\ast \XXX_t^c + \WW_{hf}\ast \HHH_{t-1} + \WW_{cf}\circ \CCC_{t-1}+\WW_f),            \\
		\DDD_t & = \II_t \circ \tanh(\WW_{xc} \ast \XXX_t^c + \WW_{hc} \ast \HHH_{t-1}+\WW_c), \\
		\CCC_t & = \FF_t \circ \CCC_{t-1} + \DDD_t,\\
		\OO_t & = \varphi(\WW_{xo}\ast \XXX_t^c + \WW_{ho}\ast \HHH_{t-1} + \WW_{co}\circ \CCC_{t}  +\WW_o),            \\
		\HHH_{t} & = \OO_t \circ \tanh(\CCC_t),
	\end{split}
	\label{eq:convlstm}
\end{equation*}

\noindent where $\ast$ denotes the convolution operation, $\circ$ is the Hadamard product, $\XXX_t^c \in \RR^{D \times W \times H}$ is the input of the ConvLSTM block at time $t$, $D$ is the number of stacked heightmaps in the blocks, $\HHH_t\in \RR^{D \times W \times H}$ is the hidden state and also the output of the block at $t$, $\FF_t \in \RR^{W \times H}$ is a gate that controls what information needs to be forgotten or retained for the next step, $\OO_t \in \RR^{W \times H}$ is an output gate that controls what information is passed on to the hidden state, $\DDD_t, \CCC_t \in \RR^{D \times W \times H}$ are the temporary cell state and the cell state at time $t$ that work to accumulate the information over the history of the sequence, $\varphi: \RR^{W \times H} \mapsto \RR^{W \times H}$ is a sigmoid function. Finally, $\WW_\bullet \in \RR^{W \times H}$ are the learnable weights and biases of the network.

Formally, we define the model uncertainty through dropouts in the neural network by sampling $K$ different sets of parameters $\WW_\bullet$.
That is,
\begin{equation*}
	 \bar\sigma^2_t = \text{Var} (\hat\YY_t^{(k)}), \quad \forall t \in \T_y.
	\label{eq: expectation}
\end{equation*}
This operation is equivalent to performing $K$ stochastic forward passes with the dropout of weights enabled during inference and then averaging the results.  In this work, we simulate the MC dropout sampling by using the $\Pr=0.4$ probability of dropping each weight set for stochastic inference during prediction. Through the use of dropouts between each layer in our network, both during training and testing time, we enable our encoder-decoder model to estimate the prediction set $\bar\Y= \left\{\bar\YY_t \in \RR^{M \times N} \mid \forall t \in \T_y\right\}$ and the corresponding variance set $\bar\Sigma= \left\{\bar\sigma^2_t \in \RR^{M \times N} \mid \forall t \in \T_y\right\}$. The MC dropout method allows our model to generate the requisite prediction estimates and the uncertainty on its prediction in a computationally efficient process. Therefore, a more detailed network model showing our intermediate outputs, i.e., $\hat\YY_t$ and $\hat\sigma^2_t$, can be summarized as
\begin{equation*}
	(\bar\YY_t, \bar\sigma^2_t) \leftarrow (\hat\YY_t, \hat\sigma^2_t) \leftarrow \Theta (\X, \WW), \quad \forall t \in \T_y.
\end{equation*}
We also refer the reader to \cite{rangwala2021deeppastl} for more details about our mean and variance predictions.  After the spatiotemporal learning of the pasture, we are ready to make predictions for the horizon $\T_y$.


\section{Multi-Robot Intermittent Deployment}
\label{sec: decision making}

The goal of the deployment to maximize the information we can get from the environment while respecting the budgets. To this end, we model the objective function $f: 2^\V \mapsto \RR$ as follows:
\begin{equation}
	f(\S) = \sum_{s \in \S} \Big(\bar\sigma^2_t(s) \sum_{s' \in \S \setminus s} \frac{d(s,s')}{|\S \setminus s|} \Big) - w_1 (t - t_1),
	\label{eq: objective function}
\end{equation}
where $\bar\sigma^2_t(s)$ is the prediction variance of the decision factor $s$ at time $t$. Note that $t_1$ is the starting time index of the prediction horizon $\T_y$. The distance function $d(s,s')$ is the weighted Euclidean distance and time difference between $s$ and $s'$. That is,
\begin{equation}
	d(s,s') = w_2 \log(||(x, y) - (x',y')||) + w_3 ||t - t'||,
\end{equation}
where $s = (x,y,r,t)$ and $s' = (x',y',r',t')$ are two different decision factors. And $w_1$ is the waiting penalty weight, $w_2$ is the weight for the physical distance, and $w_3$ is the weight for the time difference between $s$ and $s' \in \S \setminus s$.

The objective function is a weighted sum of prediction variances. The nominator $\bar\sigma^2_t(s)$ is the prediction variance of the decision factor $s$ at time $t$. The denominator is the sum of the weighted distances between $s$ and $s' \in \S \setminus s$ for all $s \in \S$. Meanwhile, $w_1$ is the weight to penalize the waiting time. We can deploy robots using a deployment set $\S$ that maximizes the objective function $f(\cdot)$ to reduce the prediction uncertainty.

Since the deployment should be energy efficient, we have two constraints to model the deployment budgets. The first constraint is,
\begin{equation}
	\quad |\S \cap \V_t| \le \ell_t, \forall t \in \T_y.
	\label{eq: M1}
\end{equation}
This constraint (per-day budget) indicates that the number of deployments cannot be larger than $\ell_t$ at time $t$, where $\ell_t \in \RR$. The second constraint is,
\begin{equation}
	\quad \sum_{t \in \T_y} \one \left(\S \cap \V_t \right) \le \ell,
	\label{eq: M2}
\end{equation}
where $\one(\cdot)$ is an indicator function as
\begin{equation*}
	\one \left(|\S \cap \V_t| \right) =
	\begin{cases}
		1, & \text{if $ |\S \cap \V_t| \ge 1 $,} \\
		0, & \text{if $ |\S \cap \V_t| = 0 $.}   \\
	\end{cases}
\end{equation*}
This constraint (total budget) suggests that the total number of deployable days cannot be larger than $\ell$, where $\ell \in \RR$. Therefore, the details of our problem formulation is
\begin{equation*}
	\begin{split}
		\underset{\S \subseteq \V}{\text{maximize}} \quad & f(\S, \bar\Sigma) \\
		\text{subject to}\quad & |\S \cap \V_t| \le \ell_t, \forall t \in \T_y\\
		& \sum_{t \in \T_y} \one \left(\S \cap \V_t \right) \le \ell.\\
	\end{split}
\end{equation*}

\begin{algorithm}[!t]
	\caption{The algorithm for the long-term pasture prediction and sensing problem.}
	\label{alg1}
	\textbf{Input:} The inputs are as follows:
	\begin{itemize}
		\item The historical  dataset $\X = \left\{\XX_t \in \RR^{M \times N} \mid \forall t \in \T_x\right\}$;
		\item The neural network $\Theta(\cdot)$;
		\item The deployment ground set $\V$;
		\item The objective function $f: 2^\V \mapsto \RR$;
		\item The matroid intersection constraint $\M = (\V, \I)$.
	\end{itemize}

	\textbf{Output:} The deployment strategy set $\S$.

	\begin{algorithmic}[1]
		\Statex
		\For{$t \in \T_y = \{\tau, \ldots, T\}$} \label{lin: predict 1}
		\State $\hat{\YY}_t^{(k)} \leftarrow \Theta(\X, \WW), \forall k=1, \ldots, K$;
		\State $\bar{\YY}_t \leftarrow \frac{1}{K} \sum_{k=1}^{K} \hat{\YY}_t^{(k)}$; \Comment{Mean}
		\State $\bar\sigma_t^2 \leftarrow \text{Var}(\hat{\YY}_t^{(k)})$; \Comment{Variance}
		\State $\bar\Sigma \leftarrow \bar\Sigma \cup \{\bar\sigma^2_t$\};
		\EndFor \label{lin: predict 2}
		\State
		\State $\S \leftarrow \emptyset, \Z \leftarrow \V$;
		\While{$\Z \neq \emptyset$}
		\State $s \in \argmax_{v \in \V \setminus \Z} f(\{v\} \mid \S)$; \label{lin: argmax}
		\If{$\S \cup \{s\} \in \I$} \label{lin: first}
		\State $\S \leftarrow \S \cup \{s\}$;
		\EndIf \label{lin: last}
		\State $\Z \leftarrow \Z \cup \{s\}$;
		\EndWhile
		\State $\S \leftarrow $ deployment strategy;
	\end{algorithmic}
\end{algorithm}

It has been shown that both constraints are matroidal \cite{liu2019submodular}, and we will use $\M_1 = (\V, \I_1)$ and $\M_2 = (\V, \I_2)$ to denote those two, where $\M_1, \M_2$ are matroids and $\I_1, \I_2$ are independent sets. To simplify the notation, we use $\M = (\V, \I)$, where $\I = \I_1 \cap \I_2$, to denote the intersection of two constraints. Thus, $\M$ is a matroid intersection constraint and the cardinality is $|\M| = 2$.

\begin{figure*}[!t]
	\centering
	\includegraphics[width=\textwidth]{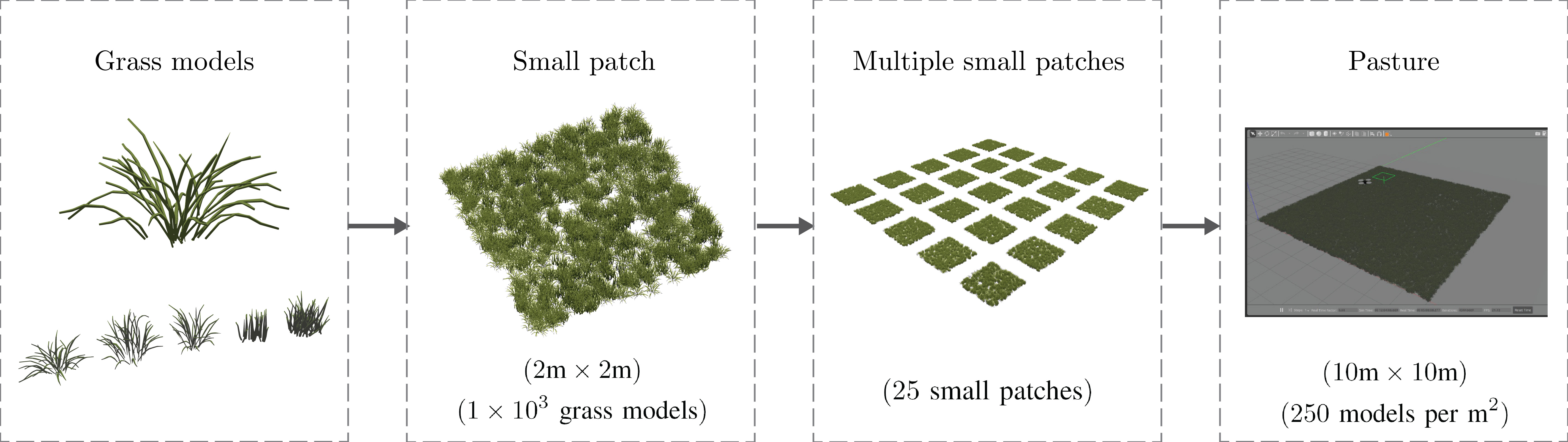}
	\caption{A diagram of our pasture construction process.}
	\label{fig: Gazebo flowchart}
\end{figure*}

Using the proposed architecture with ConvLSTM and residual connections, we have the predicted variance set $\bar\Sigma= \left\{\bar\sigma^2_t \in \RR^{M \times N} \mid \forall t \in \T_y \right\}$. Given the intermittent deployment problem (\prbref{prb: problem formulation}) and the deployment ground set $\V$, we propose to solve the problem using \algref{alg1}. From \linsref{lin: predict 1}{lin: predict 2}, we first use the proposed architecture to predict the variance set $\bar\Sigma$ for the prediction time set $\T_y$. In the second part, the algorithm greedily selects all the available decision factors $v \in \V \setminus \Z$ based on the marginal gain $f(\{v\} \mid \S)$, where $\Z$ is used to store all checked decision factors. The set $\S$ is current solution of the deployment strategy and will be updated iteratively. Specifically, we initialize a set $\Z$ as $\V$. Then, in \linref{lin: argmax}, we select one of the decision factors $v$ that maximizes the marginal gain of the objective function $f(\{v\} \mid \S)$, where $\S$ is the current solution of the problem and will be expanded as more decision factors are checked. In \linsref{lin: first}{lin: last}, we need to check if $v$ satisfies the matroidal deployment constraint $\M = (\V, \I)$. If so, $v$ is added to the solution set $\S$ as $\S \leftarrow \S \cup \{s\}$. Otherwise, the next round will be started. Meanwhile, $\Z$ is updated to store the checked decision factor $v$ as $\Z \leftarrow \Z \cup \{s\}$. The iteration will be finished when every decision factor in $\V$ is checked against the constraint $\M = (\V, \I)$.

If we define the optimal deployment policy of the intermittent deployment problem as $\S^\star$ with respect to the predicted variance set $\bar\Sigma$, we then have the following result.

\begin{theorem}[\cite{conforti1984submodular}]
	The optimality ratio of the greedy solution $\S$ generated by \algref{alg1} has the following performance:
	\begin{equation*}
		f(\S) \ge \frac{1}{|\M|+c_f} f(\S^\star) = \frac{1}{2 + c_f} f(\S^\star),
	\end{equation*}
	where $c_f$ is the curvature of $f(\cdot)$, $|\M|=2$ is the cardinality of the matroid intersection constraint, and $\S^\star$ is an optimal solution. 
\end{theorem}

The above result gives a lower bound of algorithm \algref{alg1} for our problem. Note that the curvature of the objective function $f(\cdot)$ can be evaluated by checking the contribution of every decision factor $v$ from the ground set $\V$ as shown in \defref{def: curvature}.
Therefore, by using the proposed \algref{alg1}, we can get an intermittent deployment policy using the proposed method while having a performance guarantee, as shown above.

\begin{remark}
	The proposed pipeline can also be implemented in a receding horizon manner. That is, based on the proposed deployment policy $\S$, the new robot measurements from a series of deployments can be integrated into the historical dataset $\X$ to refine the learned network $\Theta(\cdot)$. Therefore, we can produce better predictions $(\bar\YY_t, \bar\sigma^2_t)$ and thus improved plans $\S$ for the future.
\end{remark}


\section{Pasture Construction and Perception}
\label{sec: Gazebo simulation}

When we have a synthesized pasture, we need a representation of what aerial robots with LiDAR would measure. Thus, we simulate a realistic pasture environment and LiDAR measurements in Gazebo based on our synthesized data. This high-fidelity pastureland simulation environment will help us to understand the effectiveness of the simulated process. In this section, we will first focus on constructing pasture environments from the simulated data and then on the height estimation using LiDAR measurements.

\emph{Pasture construction:}
In this work, we simulate a $10$m$\times 10$m pasture using $2.5 \times 10^4$ grass models. We set the size of the simulated pasture and the density like this when considering the computational complexity. First, we randomly pick $2.5 \times 10^4$ locations from this pastureland environment. We then assign a 3D grass model in each sampled location. The heights of grass models correspond to the heights at the same locations in the smooth 3D surfaces. To accelerate the simulation speed and lower the computational requirement, we divide the pasture into $25$ small patches ($2$m$\times 2$m per patch). In each patch, the density is $250$ grass models per square meter with a total of $1 \times 10^3$ models per small patch. In this small patch, we use five species of plants to simulate different growth patterns as shown in \figref{fig: Gazebo flowchart}. This selection will be validated later in the experimental section as the actual measurements look similar to our simulated environment.
Each plant is spawned at the same randomly chosen coordinates throughout our simulation. We rescale a grass model in each dimension for each randomly selected location according to the desired height in the 3D surface. The flowchart of our pasture construction is shown in \figref{fig: Gazebo flowchart}.

\emph{Pasture perception}:
To estimate the height of the pasture, we use a UAV equipped with a LiDAR to collect point clouds over the pasture. Meanwhile, we need post-processing to remove noise after getting the point clouds. Those extra points are not part of the simulated field and need to be removed. To achieve this, we use crop box filters to remove the extra points and retain the points of the $10$m$\times 10$m pasture as well as the points in the perimeter around it.
The height of a point cloud includes two parts: the height of pasture and the height of the ground plane. To get the estimated height of the field, we use the mowed-down perimeter to estimate the ground plane. First, we have the following assumption of the ground of the environment.
\begin{assumption}[Ground Plane]
	We assume that the ground of the pastureland is a plane.
\end{assumption}
In the simulation, we use this assumption to facilitate the ground height estimation. However, other types of ground can also be integrated into our simulation framework, where we can use more sophisticated methods for ground surface regression.

In this work, we use the least squares method to compute the height of the ground by using perimeter points. The perimeter is all the points surrounding the target plot area. These perimeter points will be used to estimate the ground plane that is used for height estimation. An equation for a plane can be defined as shown below.
\begin{equation}
	Ax + By + Cz + D = 0.
\end{equation}
where $A, B, C$, and $D$ are the parameters defining the plane. Without loss of generality, we assume $C = 1$.
We denote by $\p_i = [x_i, y_i, z_i]^\top \in \RR^3$ the location of $i$th point in the perimeter. Since we are solving for a best-fit plane of multiple points, a least square form of our formulation can be written as
\begin{equation*}
	\begin{bmatrix}
		{x_{1}} & {y_{1}} & 1 \\
		{x_{2}} & {y_{2}} & 1 \\
		        & ...         \\
		{x_{P}} & {y_{P}} & 1
	\end{bmatrix}
	\begin{bmatrix}
		A \\
		B \\
		D
	\end{bmatrix}
	= -
	\begin{bmatrix}
		{z_{1}} \\
		{z_{2}} \\
		...     \\
		{z_{P}}
	\end{bmatrix},
\end{equation*}
where $P$ is the total number of points in the perimeter.

To perform the linear least squares test, we multiple a transpose of the left-most matrix on both sides. Note that we will use $\sum x_i x_i$ as $\sum_{i=1}^P x_i x_i$ for simplicity. This notation will also be applied to other relevant terms. Then the above equation is simplified as
\begin{equation*}
	\begin{bmatrix}
		\sum {{x}_{i}{x}_{i}} & \sum {{x}_{i}{y}_{i}} & \sum {{x}_{i}} \\
		\sum {{y}_{i}{x}_{i}} & \sum {{y}_{i}{y}_{i}} & \sum {{y}_{i}} \\
		\sum {{x}_{i}}        & \sum {{y}_{i}}        & P
	\end{bmatrix}
	\begin{bmatrix}
		A \\
		B \\
		D
	\end{bmatrix}
	=
	-
	\begin{bmatrix}
		\sum {{x}_{i}{z}_{i}} \\
		\sum {{y}_{i}{z}_{i}} \\
		\sum {{z}_{i}}
	\end{bmatrix}.
\end{equation*}

Defining all points, i.e., $\p_i = [x_i, y_i, z_i]^\top$, relative to the plane centroid sets the summations of the individual components to 0 and simplifies the above equations. The centroid of the plane is calculated by using $\o = P^{-1} \sum_{i=1}^P \p_i \in \RR^3$. Note that $P \neq |\P|$. Then, all the points in the perimeter are updated as $[x_i, y_i, z_i]^\top \leftarrow  [x_i, y_i, z_i]^\top - \o$. The above equation can then be simplified as
\begin{equation*}
	\begin{bmatrix}
		\sum {{x}_{i}{x}_{i}} & \sum {{x}_{i}{y}_{i}} & 0 \\
		\sum {{y}_{i}{x}_{i}} & \sum {{y}_{i}{y}_{i}} & 0 \\
		0                     & 0                     & P
	\end{bmatrix}
	\begin{bmatrix}
		A \\
		B \\
		D
	\end{bmatrix}
	=
	-
	\begin{bmatrix}
		\sum {{x}_{i}{z}_{i}} \\
		\sum {{y}_{i}{z}_{i}} \\
		0
	\end{bmatrix}.
\end{equation*}

From the last row, we observe that $D = 0$ since $P \neq 0$. Therefore, this equation can further be simplified as
\begin{equation*}
	\begin{bmatrix}
		\sum {{x}_{i}{x}_{i}} & \sum {{x}_{i}{y}_{i}} \\
		\sum {{y}_{i}{x}_{i}} & \sum {{y}_{i}{y}_{i}}
	\end{bmatrix}
	\begin{bmatrix}
		A \\
		B
	\end{bmatrix}
	=
	-
	\begin{bmatrix}
		\sum {{x}_{i}{z}_{i}} \\
		\sum {{y}_{i}{z}_{i}}
	\end{bmatrix}.
\end{equation*}
Using Cramer's rule, we get $A$ and $B$ as follows.
\begin{equation}
	\begin{split}
		A & = (\sum y_i z_i \times\sum x_i y_i - \sum x_i z_i \times\sum y_i y_i) / \Delta, \\
		B & = (\sum x_i y_i \times\sum x_i z_i - \sum x_i x_i \times\sum y_i z_i) / \Delta,
	\end{split}
\end{equation}
where
\begin{equation*}
	\Delta = \sum x_i x_i \times\sum y_i y_i - \sum x_i y_i\times\sum x_i y_i.
\end{equation*}
Therefore, the ground plane is fully defined, and the actual height of the pastureland can be adjusted accordingly. 

\begin{table}[!tbp]
    \centering
	\footnotesize
	\caption{The locations, length-scales, and initial weights of different basis functions.}
	\label{tab: basis function}
	\begin{tabular}{F{0.05\linewidth}F{0.2\linewidth}F{0.23\linewidth}F{0.25\linewidth}}
		\hline\noalign{\smallskip}
		$i$ & Location $(x,y)$ & Length-scale $c_i$ & Initial weight $w_i$ \\
		\noalign{\smallskip}\hline\noalign{\smallskip}
		1   & (5.0, 5.0)       & 0.13               & 4.17                 \\
		2   & (3.0, 4.0)       & 0.13               & 4.17                 \\
		3   & (2.0, 1.5)       & 0.15               & 2.50                 \\
		4   & (8.0, 8.0)       & 0.18               & 6.67                 \\
		5   & (8.0, 1.5)       & 0.13               & 3.33                 \\
		6   & (1.0, 1.0)       & 0.13               & 3.33                 \\
		7   & (1.0, 9.0)       & 0.25               & 4.17                 \\
		\noalign{\smallskip}\hline
	\end{tabular}
\end{table}

\begin{figure}[!tbp]
	\centering
	\hspace*{\fill}%
	\subfigure[Average height.]{
		\label{fig: average height}
		\includegraphics[width = 1.59in]{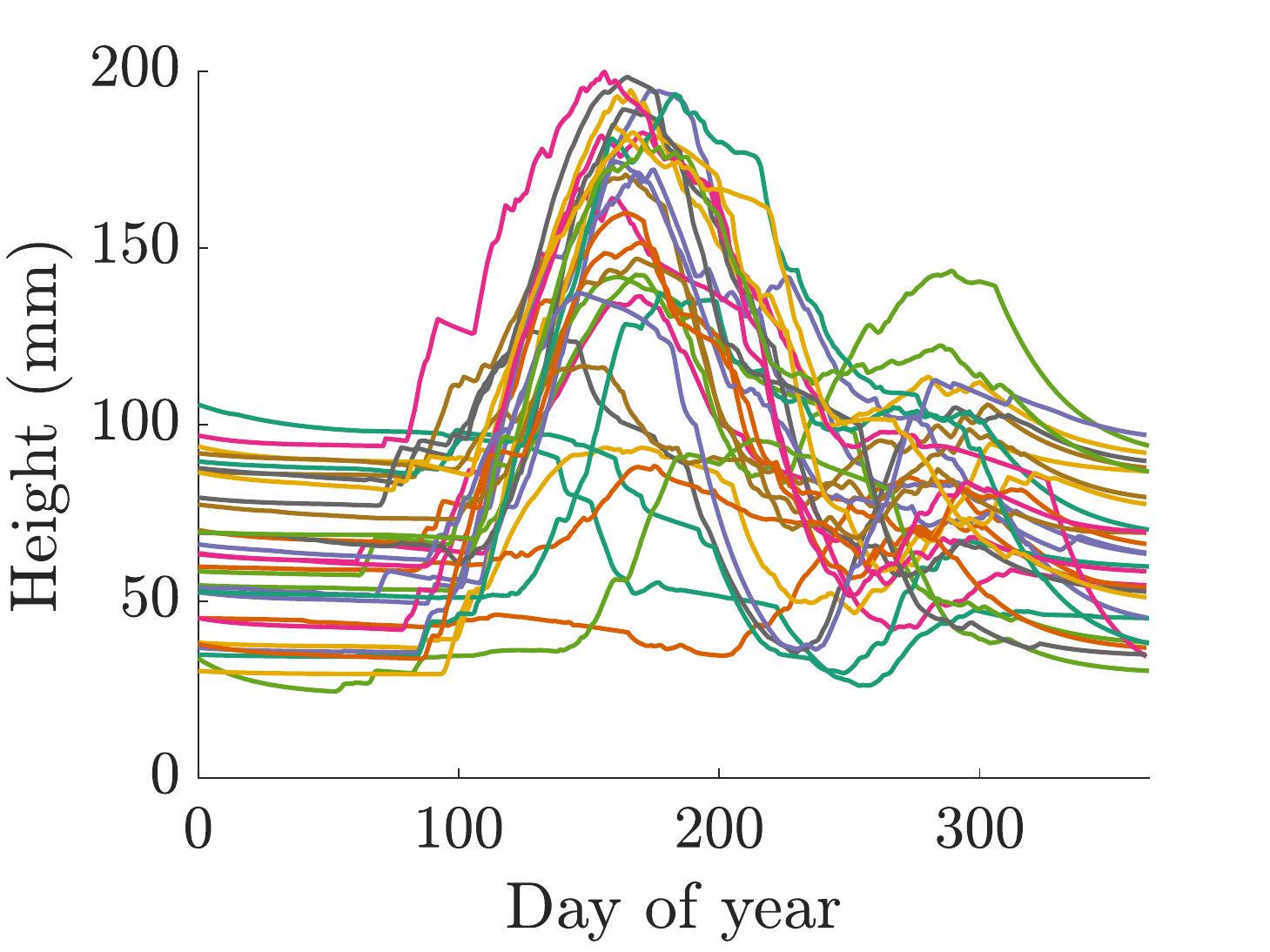}}
	\hfill%
	\subfigure[Mean \& standard deviation.]{
		\label{fig: error_bar}
		\includegraphics[width = 1.59in]{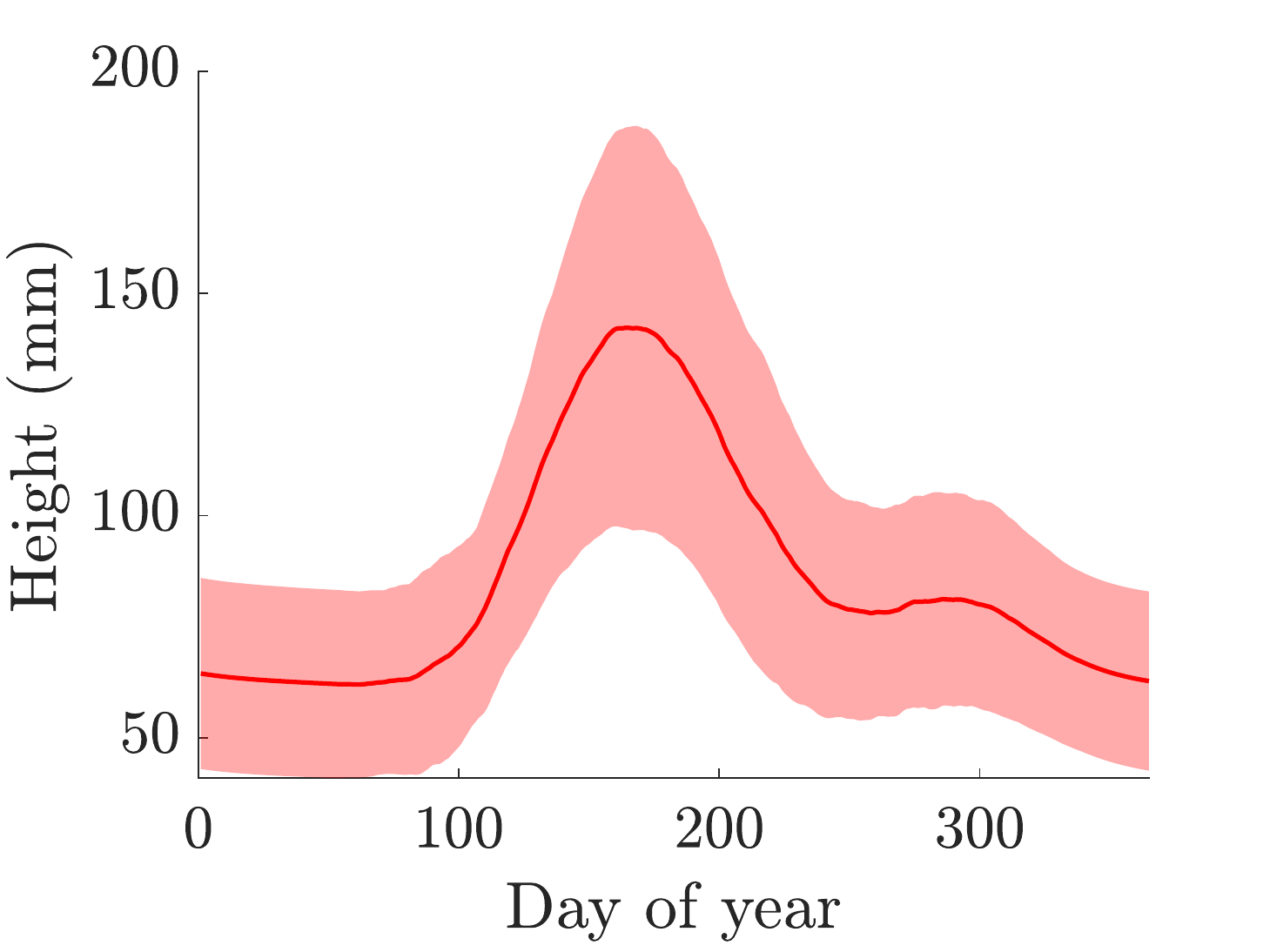}}
	\hspace*{\fill}%
	\caption{(a). The average height of the pastureland environment in 30 years (represented by 30 lines). The x-axis represents the day of a year. The y-axis represents the corresponding average height. (b). The mean and the standard deviation of the height of each day are calculated by using the historical data.}
	\label{fig: historical data}
\end{figure}

\section{Evaluation}
\label{sec: evaluation}

In this section, we demonstrate the results of each component in our pipeline. The historical data is generated using Matlab. The pasture is simulated by using Blender. The neural network training was conducted with a PyTorch backend on a 2x AMD Epyc 7742 CPUs and a multi-GPU training regime with 8x Nvidia RTX 6000 GPUs.


\subsection{Data Preparation}
\label{ssec: data preparation}

Based on historical meteorological data from a site in Iowa, we simulate $30$ years of tall fescue pasture dry matter production. 
Specifically, we will use the first $28$ years data for training and reserve the last year's data for testing purposes. In \figref{fig: historical data}, we demonstrate the statistics of the historical data. \figref{fig: average height} shows the average height change of the simulated pastureland environment, where each line denotes the height change for different days in a year. We also calculate the mean and the standard deviation for each day using the historical data as shown in \figref{fig: error_bar}. 
\begin{figure}[!tbp]
	\centering
	\includegraphics[width=\imwidth]{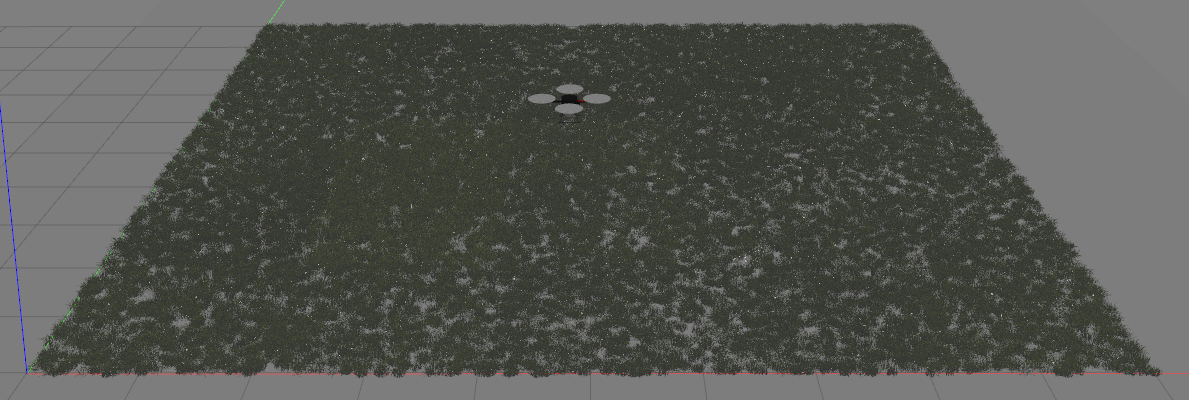}
	\caption{A simulated pastureland $10$m $\times$ $10$m environment using $2.5 \times 10^4$ grass models}
	\label{fig: average height pasture}
\end{figure}

\begin{figure}[!tbp]
	\centering
	\includegraphics[width=\imwidth]{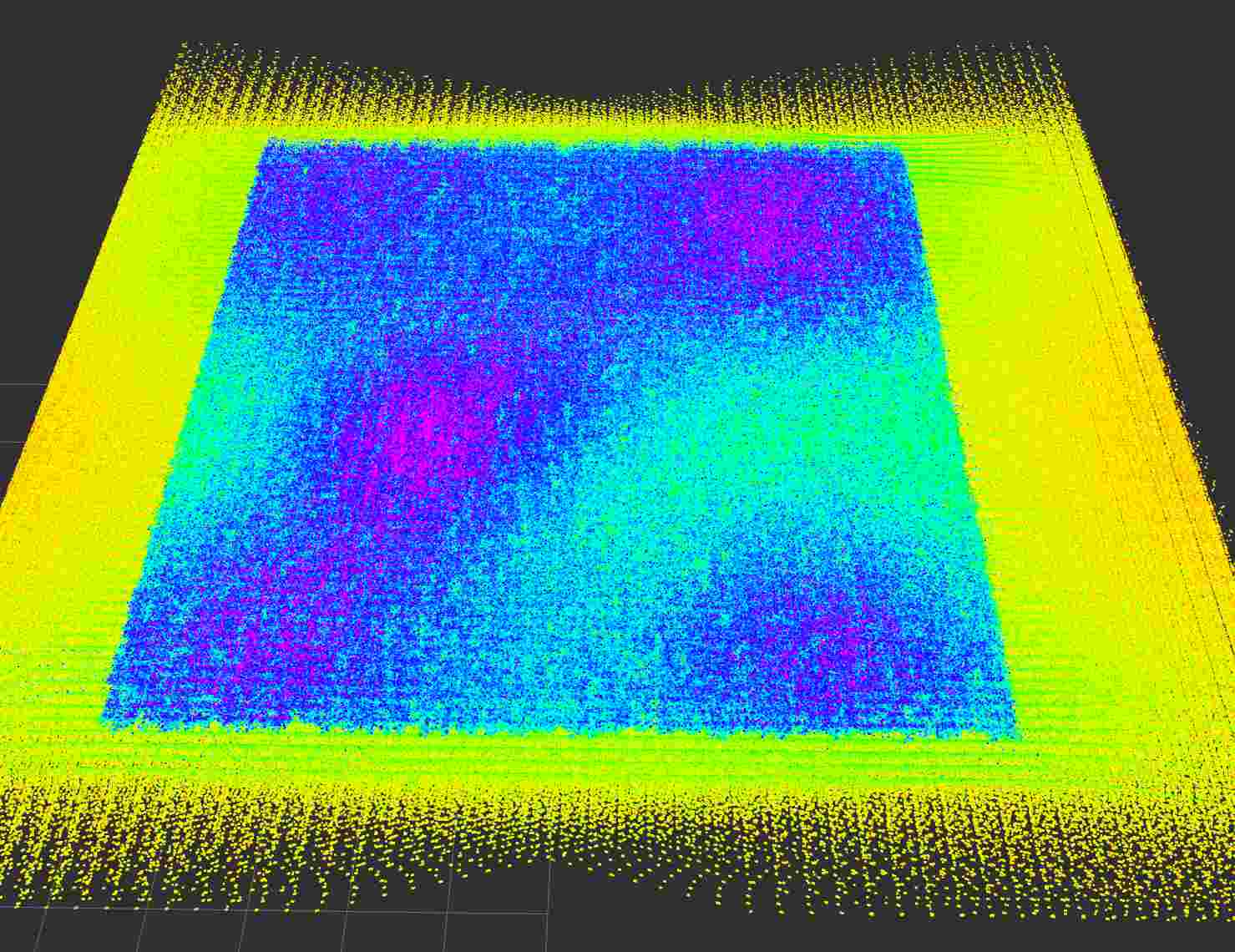}
	\caption{The corresponding point cloud of the pasture ($10$m $\times$ $10$m) shown in \figref{fig: average height pasture}.}
	\label{fig: point cloud of pasture}
\end{figure}

The simulated pasture yield was then used to generate average pasture height data based on the model in \cite{Schaefer_Lamb_2016} describing the relationship between pasture green dry matter and LiDAR-measured pasture height for each day.
Following the pastureland generation procedure described in \secref{ssec: GMM}, we generate a pastureland environment using a dynamic GMM. In the simulation, we use $B = 7$ basis functions. The random initial settings of the basis function location $(x,y)$, the length-scale $c_i$, and the initial weight $w_i$ of each basis function $b_i(x,y)$ are shown in \tabref{tab: basis function}. Later on, in our experimental section (\secref{ssec: perception results}), we can see that those parameter settings can help us to simulate pastureland environments that are very close to the real-world scenario.

\subsection{Pasture Environment Simulation Results}
\label{ssec: Gazebo simulation results}

We simulate a $10$m $\times 10$m pasture after careful consideration of the computational speed and complexity of the entire process. To further improve the simulation speed, we divide the pasture into $25$ small $2$m$\times 2$m patches. For each patch, we use $1 \times 10^3$ grass models (250 models per square meters).
Finally, $25$ small patches are attached to form one large $10$m $\times 10$m pastureland environment.

\begin{figure}[!tbp]
	\centering
	\includegraphics[width=\imwidth]{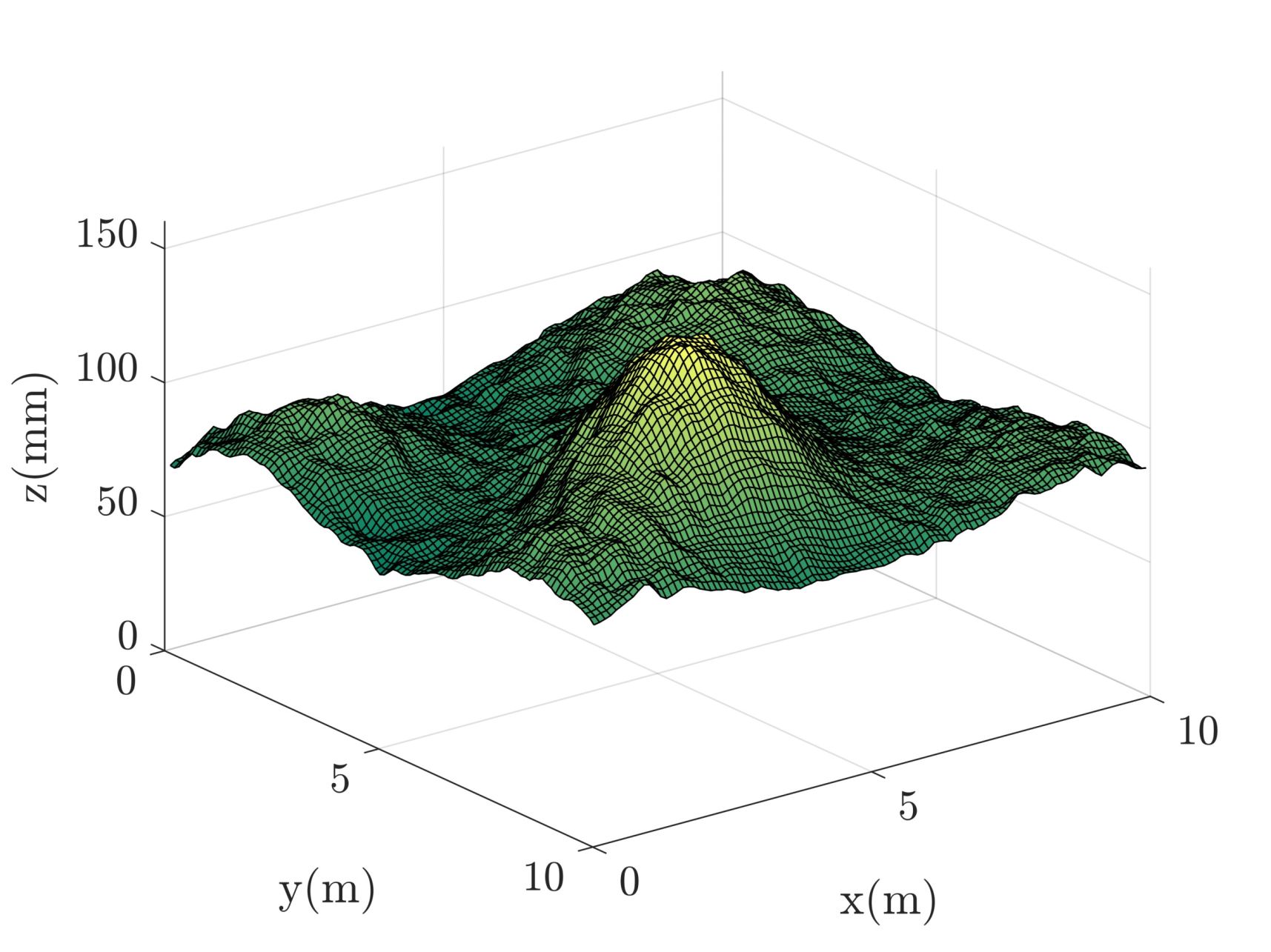}
	\caption{The downsampled surface of the point cloud shown in \figref{fig: point cloud of pasture}.
	}
	\label{fig: processed point cloud}
\end{figure}

\begin{figure}[!tbp]
	\centering
	\hspace*{\fill}%
	\subfigure[Before processing.]{
		\label{fig: contour raw}
		\includegraphics[width = 1.55in]{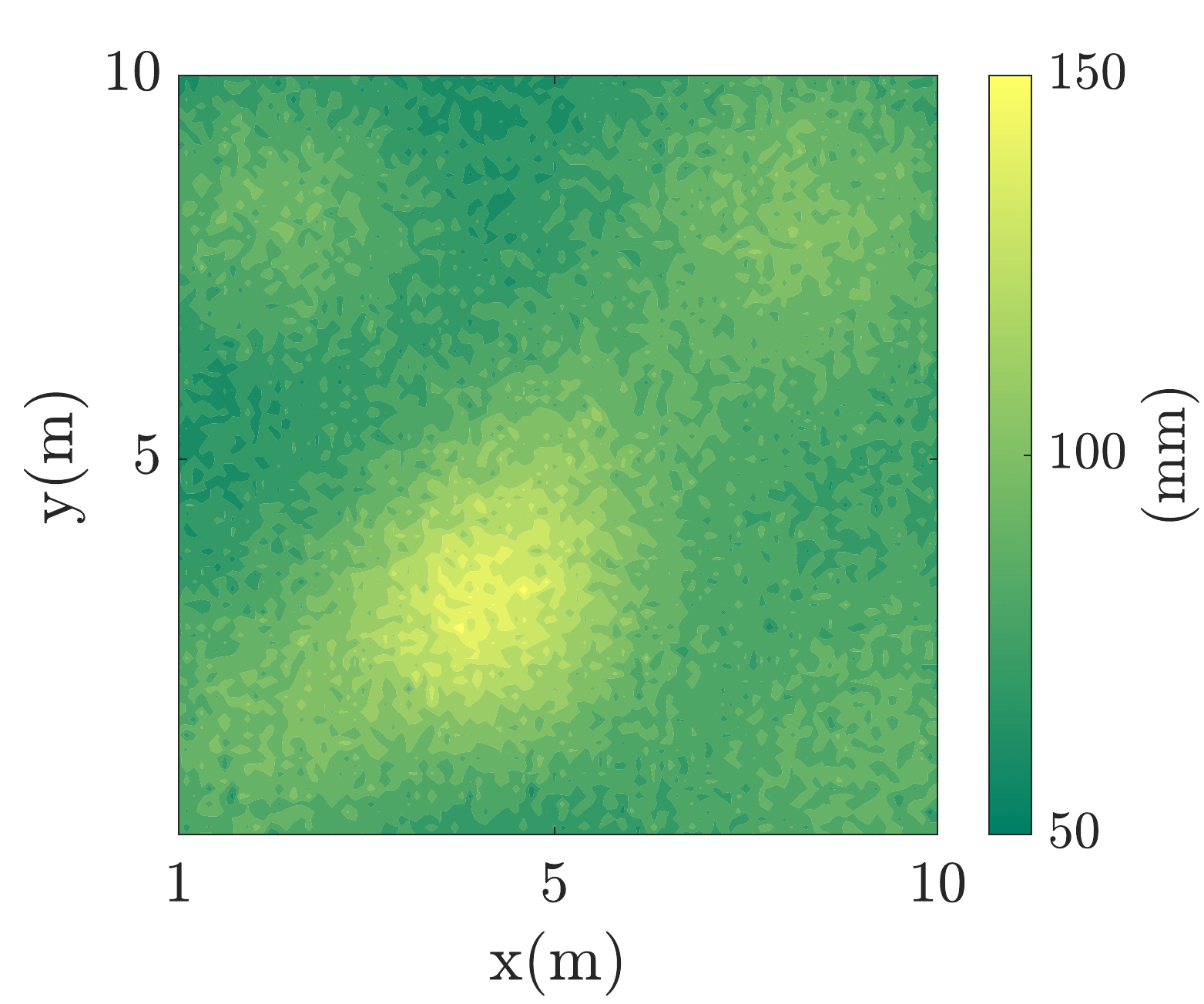}}
	\hfill%
	\subfigure[After processing.]{
		\label{fig: contour processed}
		\includegraphics[width = 1.55in]{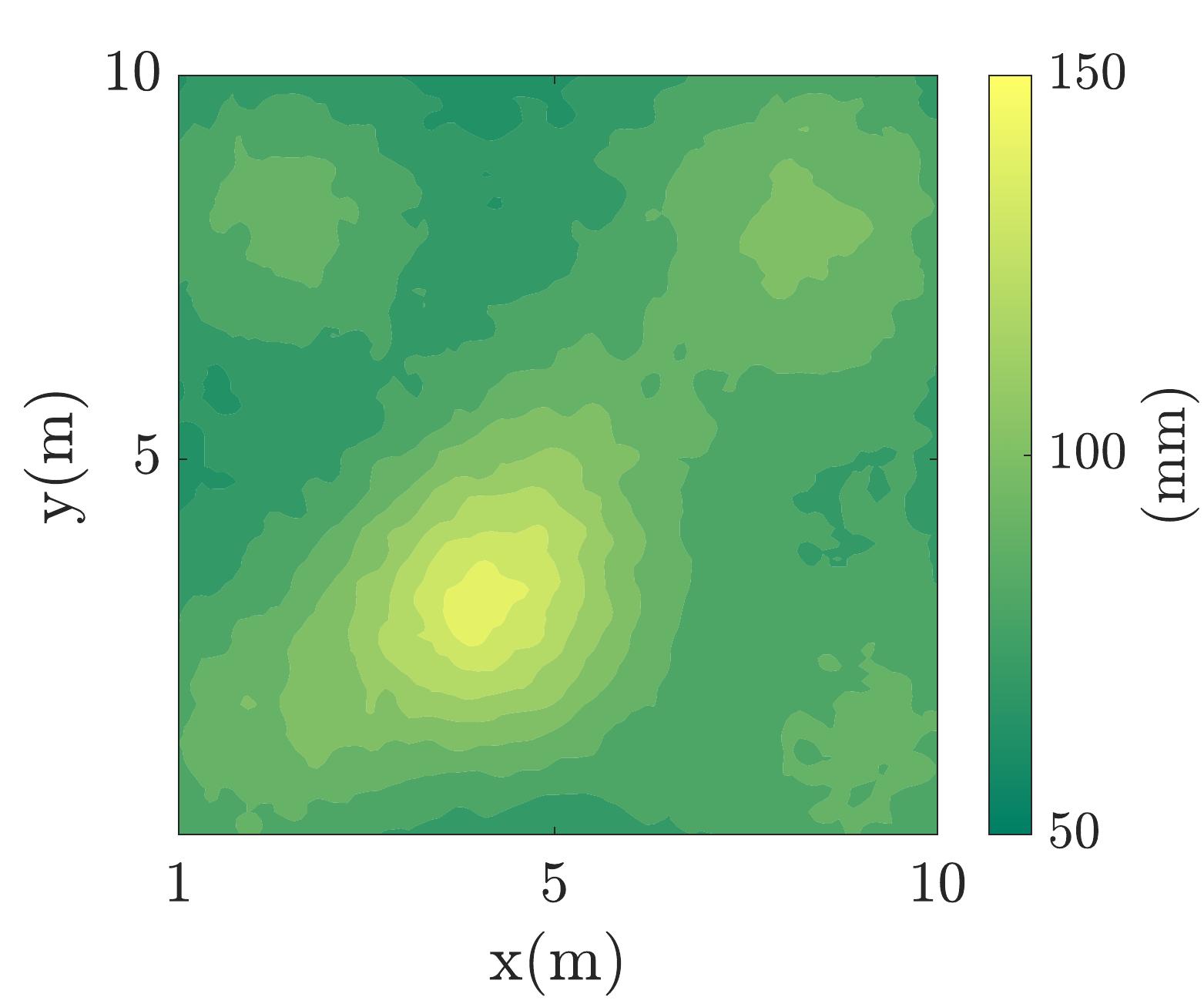}}
	\hspace*{\fill}%
	\caption{The downsampled heightmaps of the point cloud shown in \figref{fig: point cloud of pasture}.
	}
	\label{fig: contours}
\end{figure}

We randomly assign grass models and extrude the model in three dimensions to match the predefined height for the model in that location. As the swards of the grass grow with a curvature, we need to adjust the location of the base of each grass model i.e. the location of each individual grass model at the ground plane to match the location of its sward at its highest point. This offset allows us to directly calculate the vertical height of each individual plant.
Specifically, denote by $\gamma \in \RR$ the scaling factor of a grass model, which is calculated by comparing the desired height of a grass model and the original height. We denote $\m = [m_x,m_y]^\top \in \RR^2$ the original 2D location of the topmost point, and $\xi = [\xi_x, \xi_y]^\top \in \RR^2$ the offset of the topmost point, where $\xi_x, \xi_y \in \RR$ are the offset in $x$ and $y$ direction. We apply a shifting process as $\m \leftarrow \m - \gamma \cdot \xi$ to ensure that the topmost point is at the predefined location. After this adjustment, the 2D location of the topmost point of a grass model is transformed to the predefined 2D location.

\begin{figure}[!tbp]
	\centering
	\includegraphics[width=\imwidth]{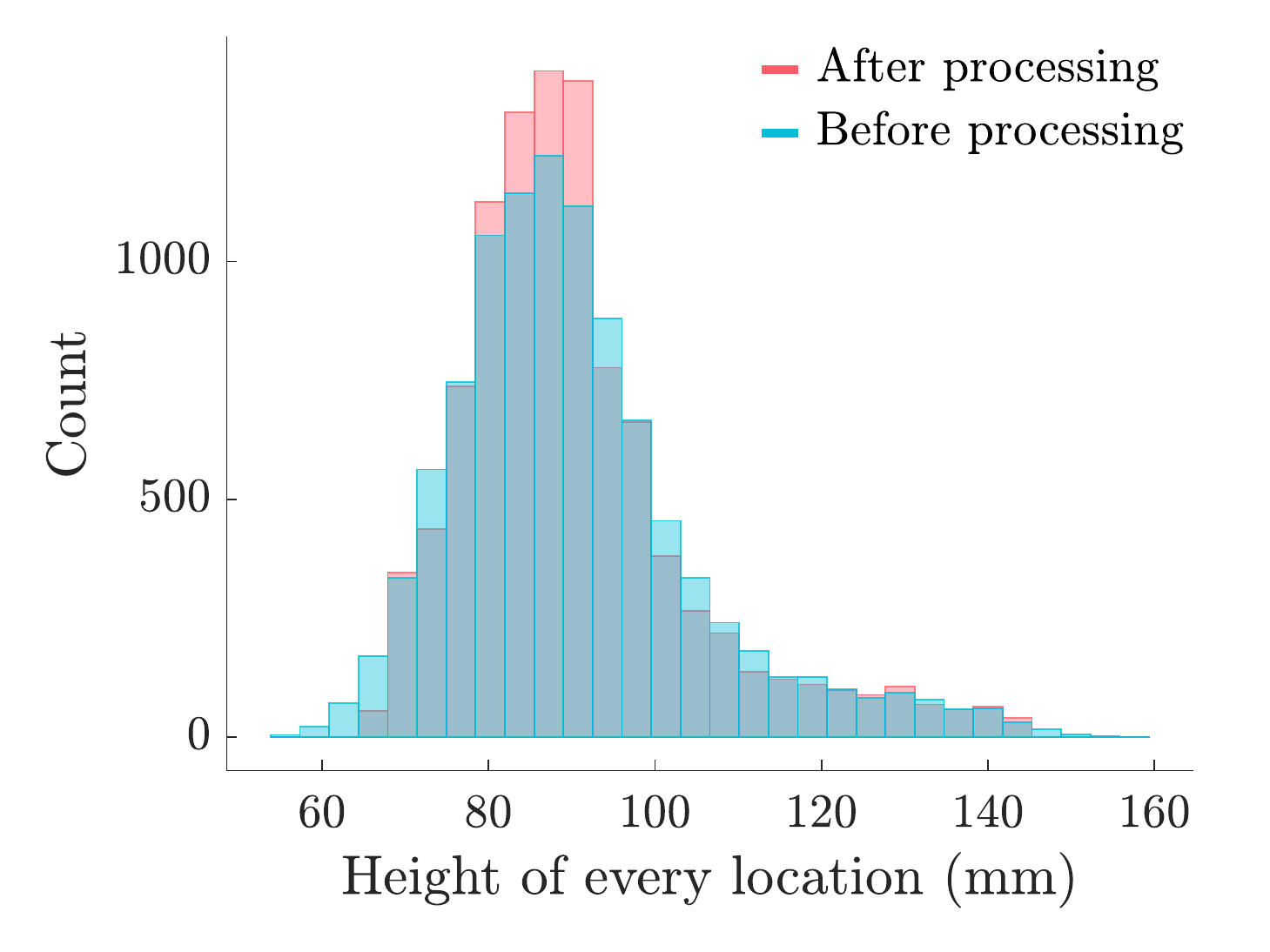}
	\caption{The histogram comparison of the two heightmaps shown in \figref{fig: contours}.}
	\label{fig: hist comparison}
\end{figure}

\begin{figure}[!tbp]
	\centering
	\includegraphics[width=\imwidth]{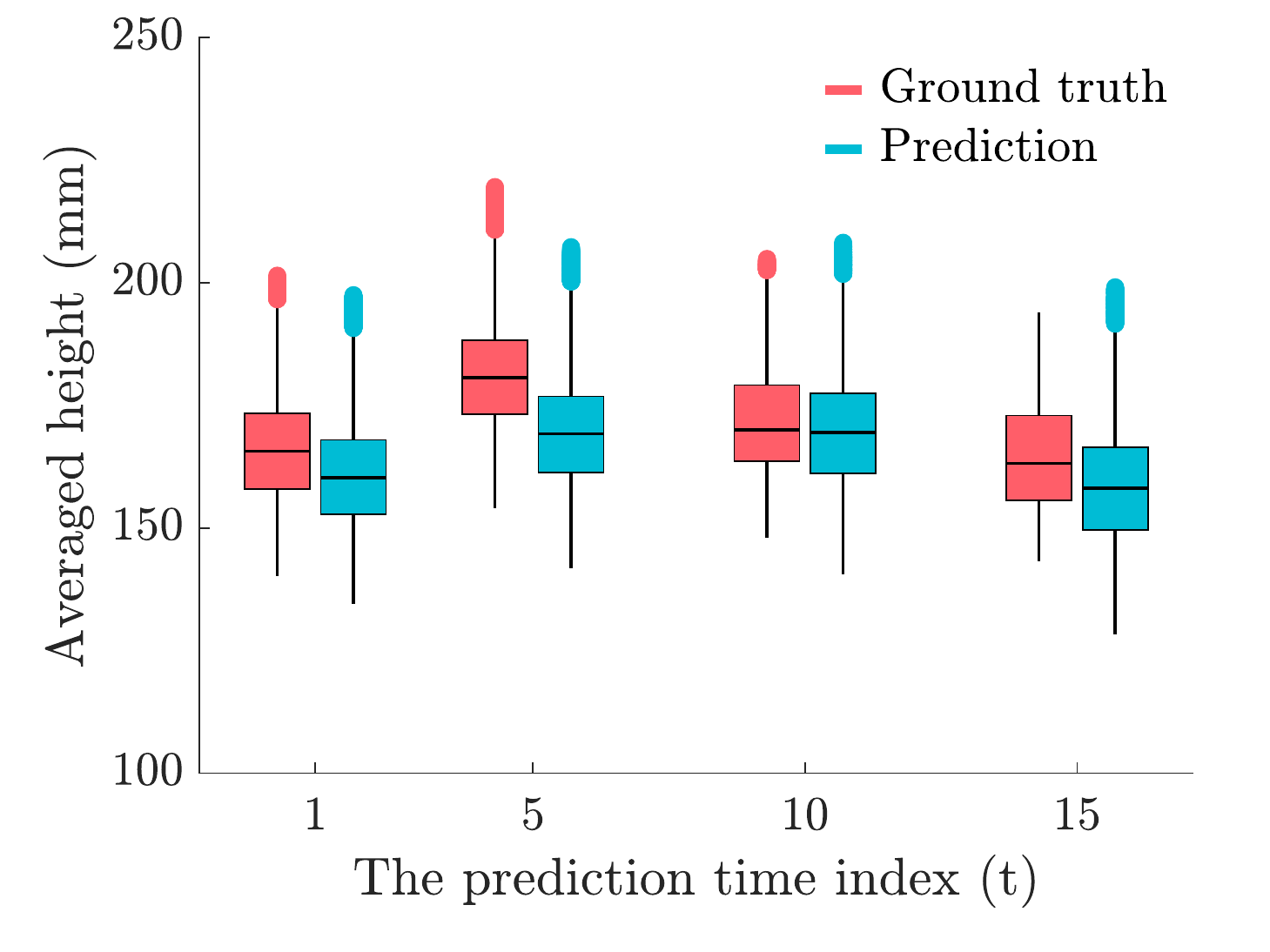}
	\caption{The averaged height comparison between the ground truth and the prediction for different prediction time index with $\alpha = 1, 5, 10$, and $15$. Since the time interval is $\delta=4$, the effective prediction length are $4, 20, 40$, and $60$ days.}
	\label{fig: boxplot}
\end{figure}

\begin{figure*}[!t]
	\centering
	\subfigure[$5$th ground truth.]{
		\label{fig: 2D_ground_truth_5}
		\includegraphics[width = 0.23\textwidth]{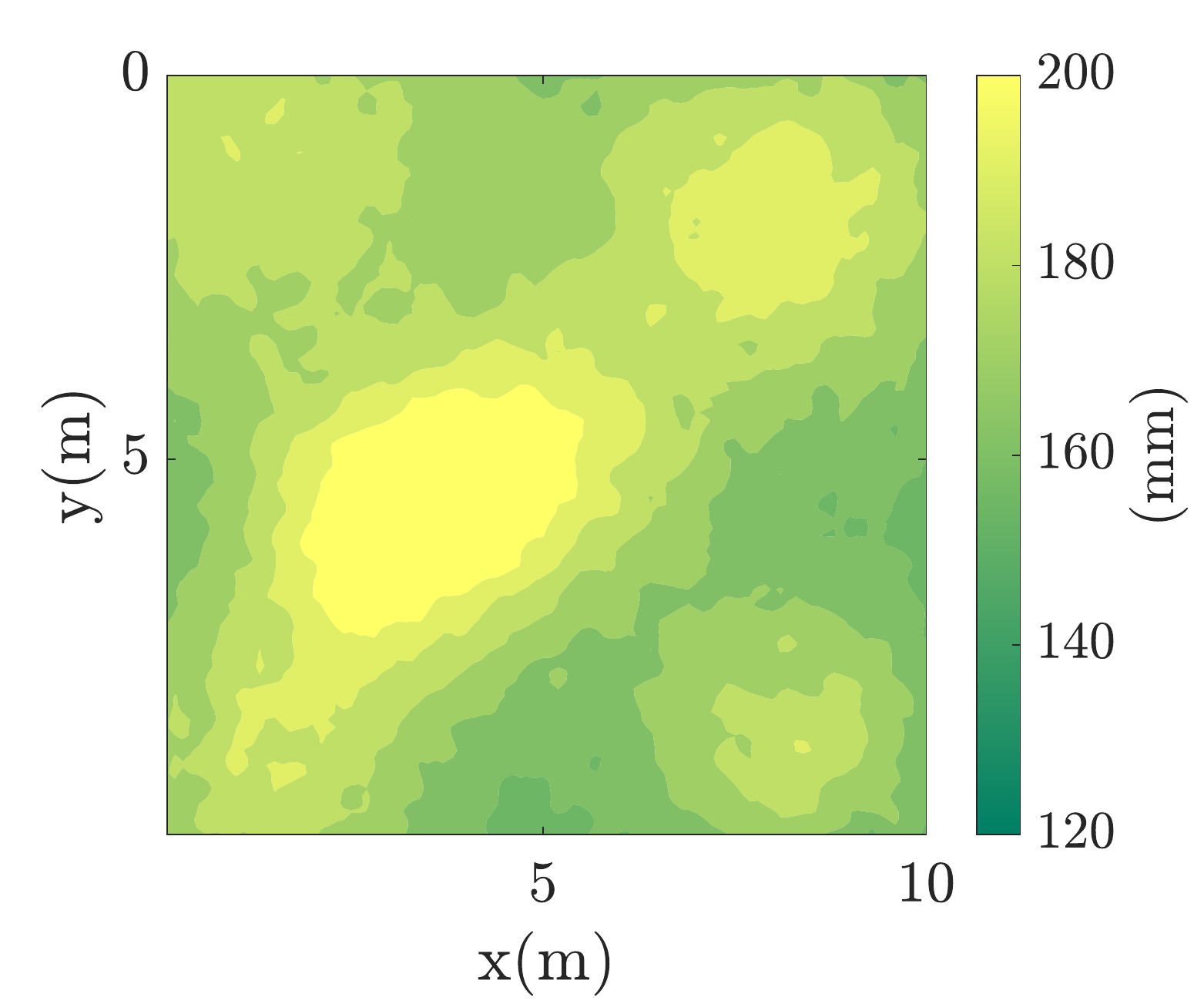}}
	%
	\subfigure[$5$th predicted mean.]{
		\label{fig: 2D_predict_mean_5}
		\includegraphics[width = 0.23\textwidth]{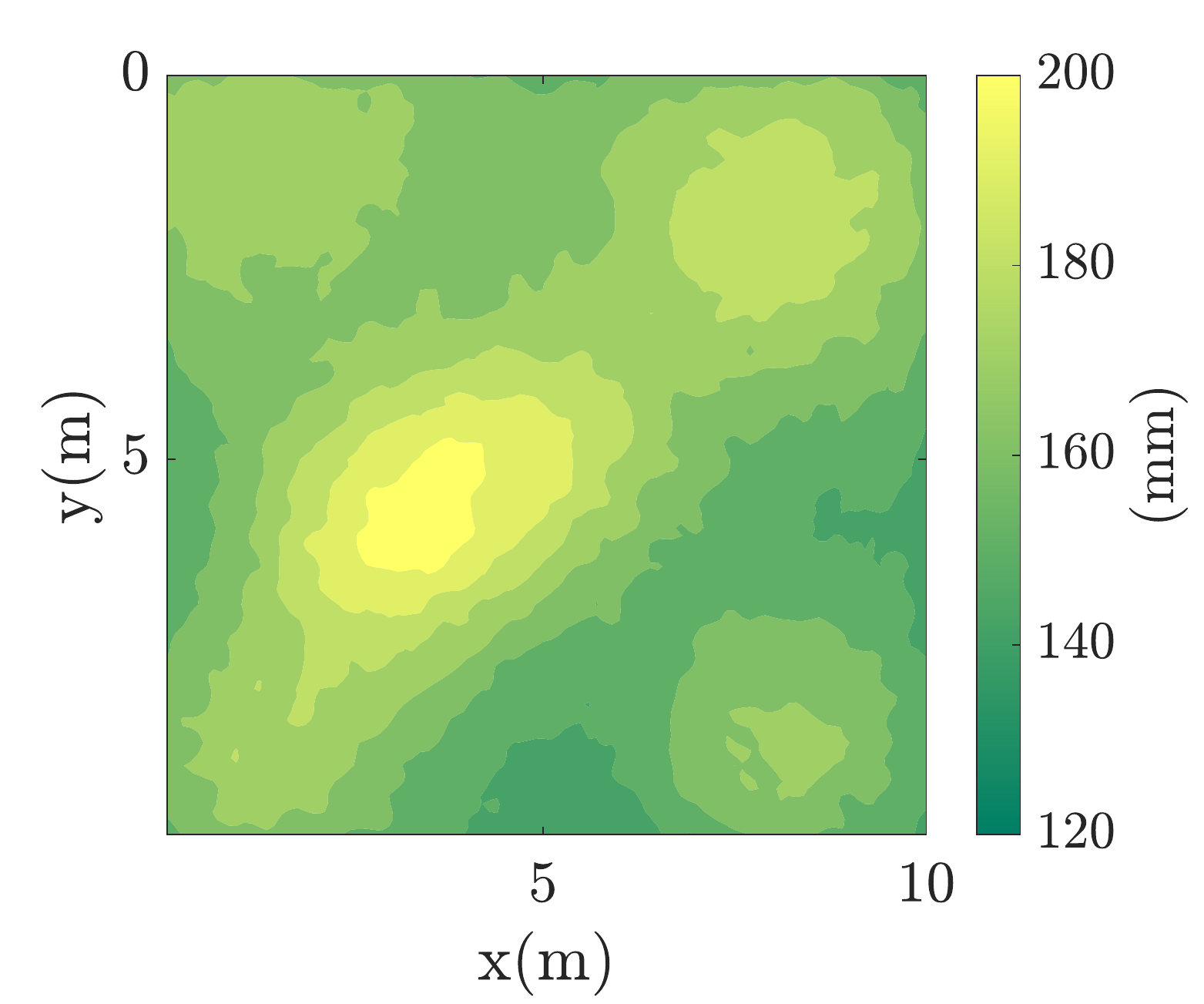}}
	%
	\subfigure[$5$th predicted std.]{
		\label{fig: 2D_predict_std_5}
		\includegraphics[width = 0.23\textwidth]{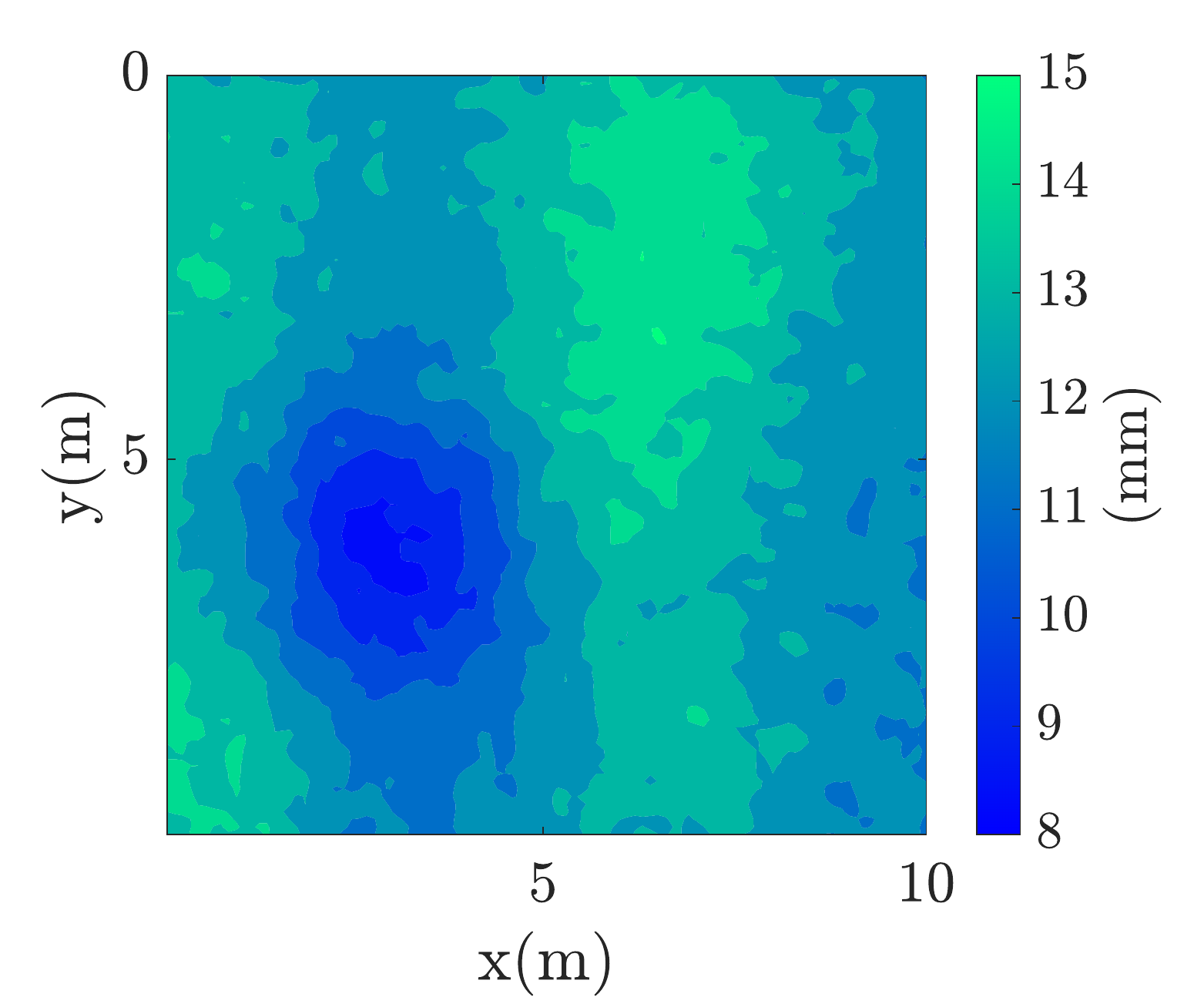}}
	%
	\subfigure[$5$th predicted error.]{
		\label{fig: 2D_predict_error_5}
		\includegraphics[width = 0.23\textwidth]{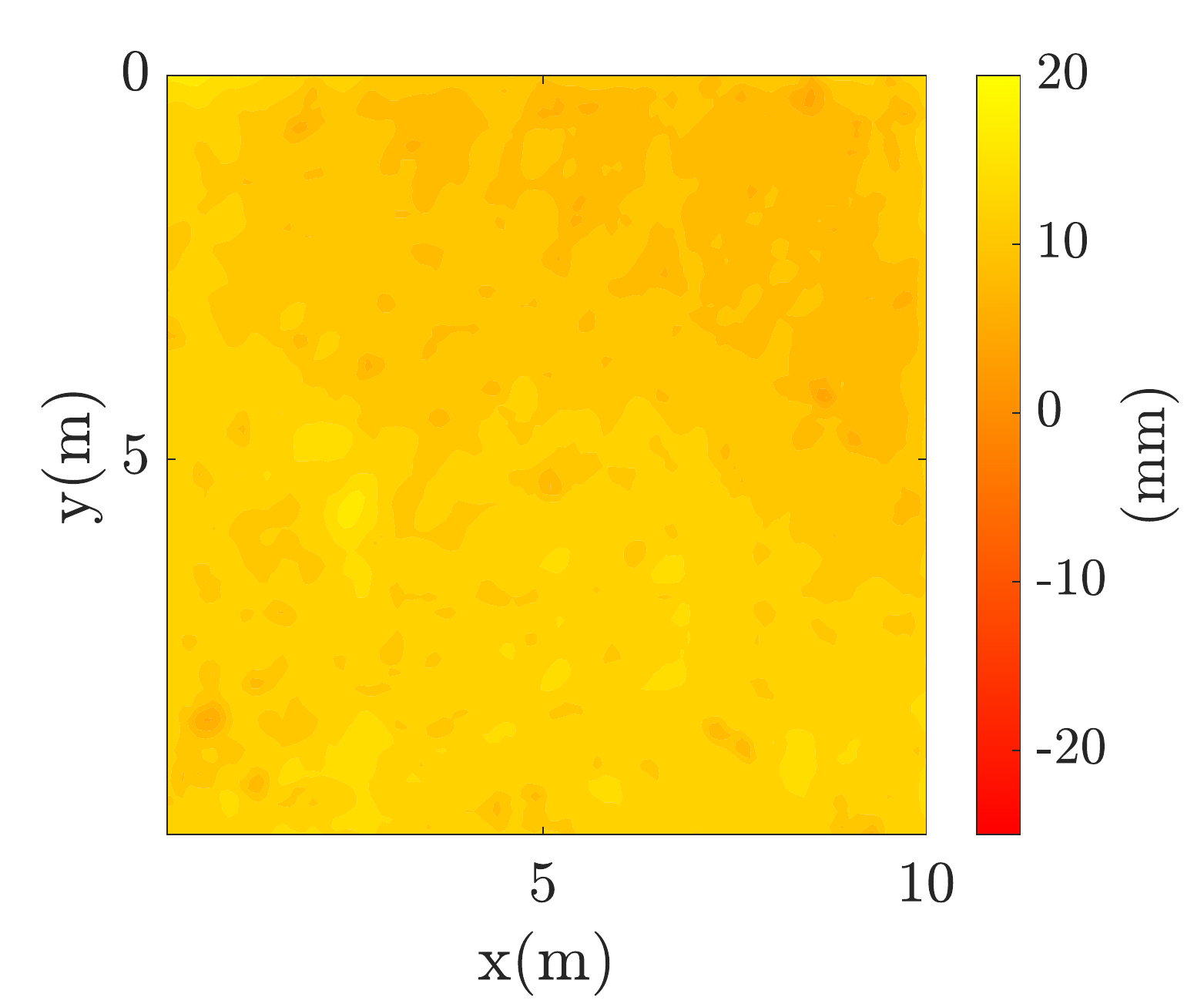}}
	%
	\subfigure[$15$th ground truth.]{
		\label{fig: 2D_ground_truth_15}
		\includegraphics[width = 0.23\textwidth]{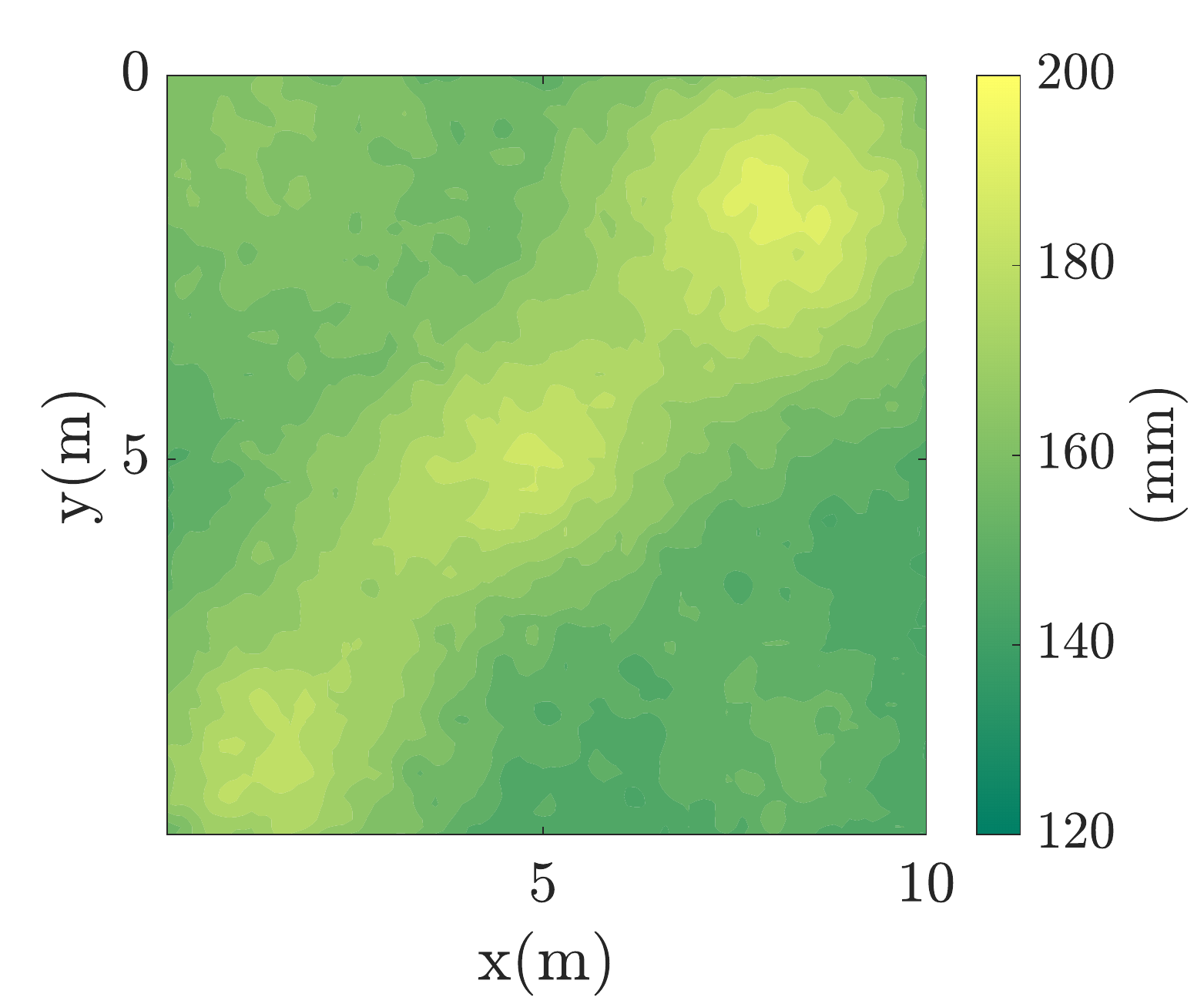}}
	%
	\subfigure[$15$th predicted mean.]{
		\label{fig: 2D_predict_mean_15}
		\includegraphics[width = 0.23\textwidth]{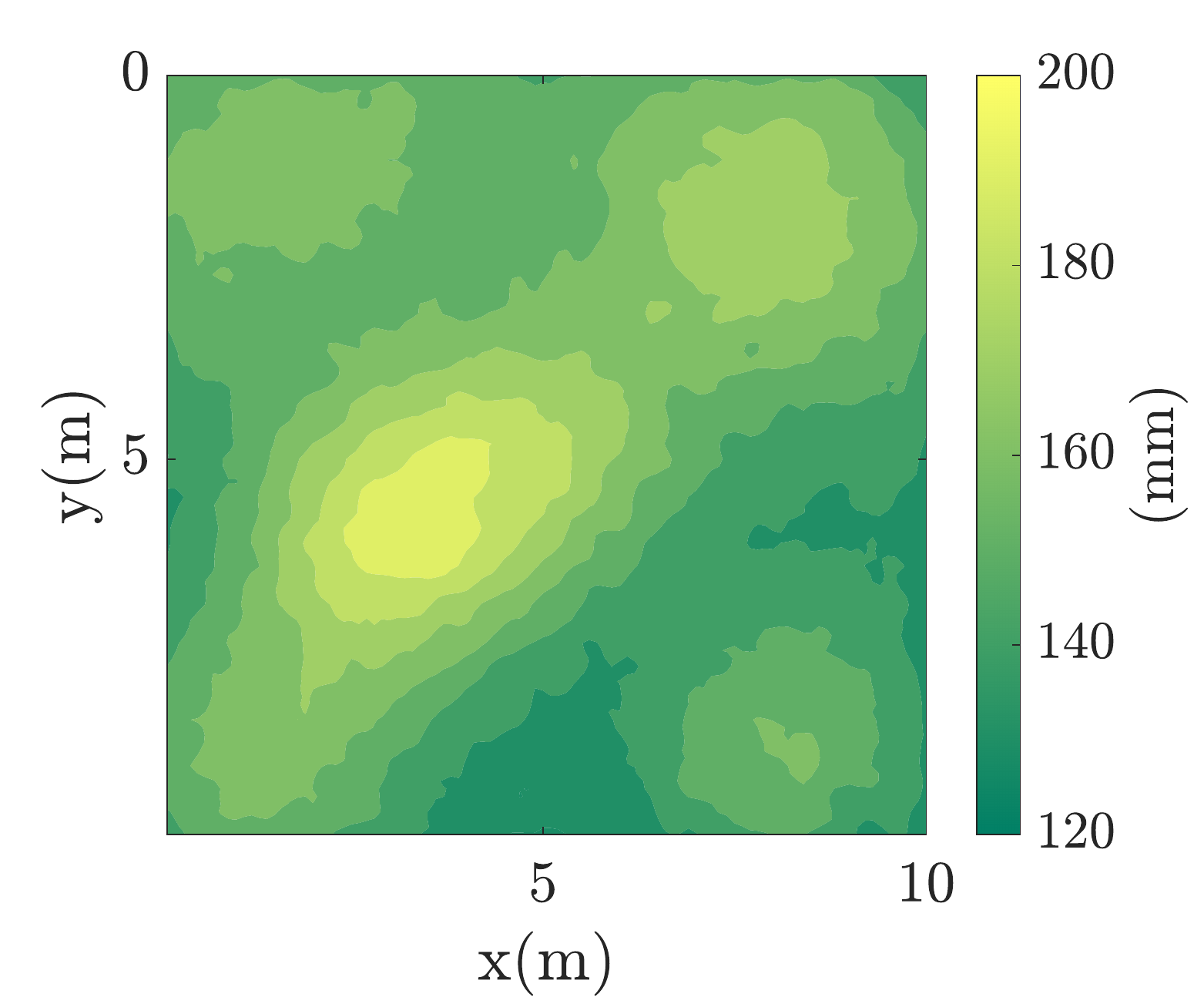}}
	%
	\subfigure[$15$th predicted std.]{
		\label{fig: 2D_predict_std_15}
		\includegraphics[width = 0.23\textwidth]{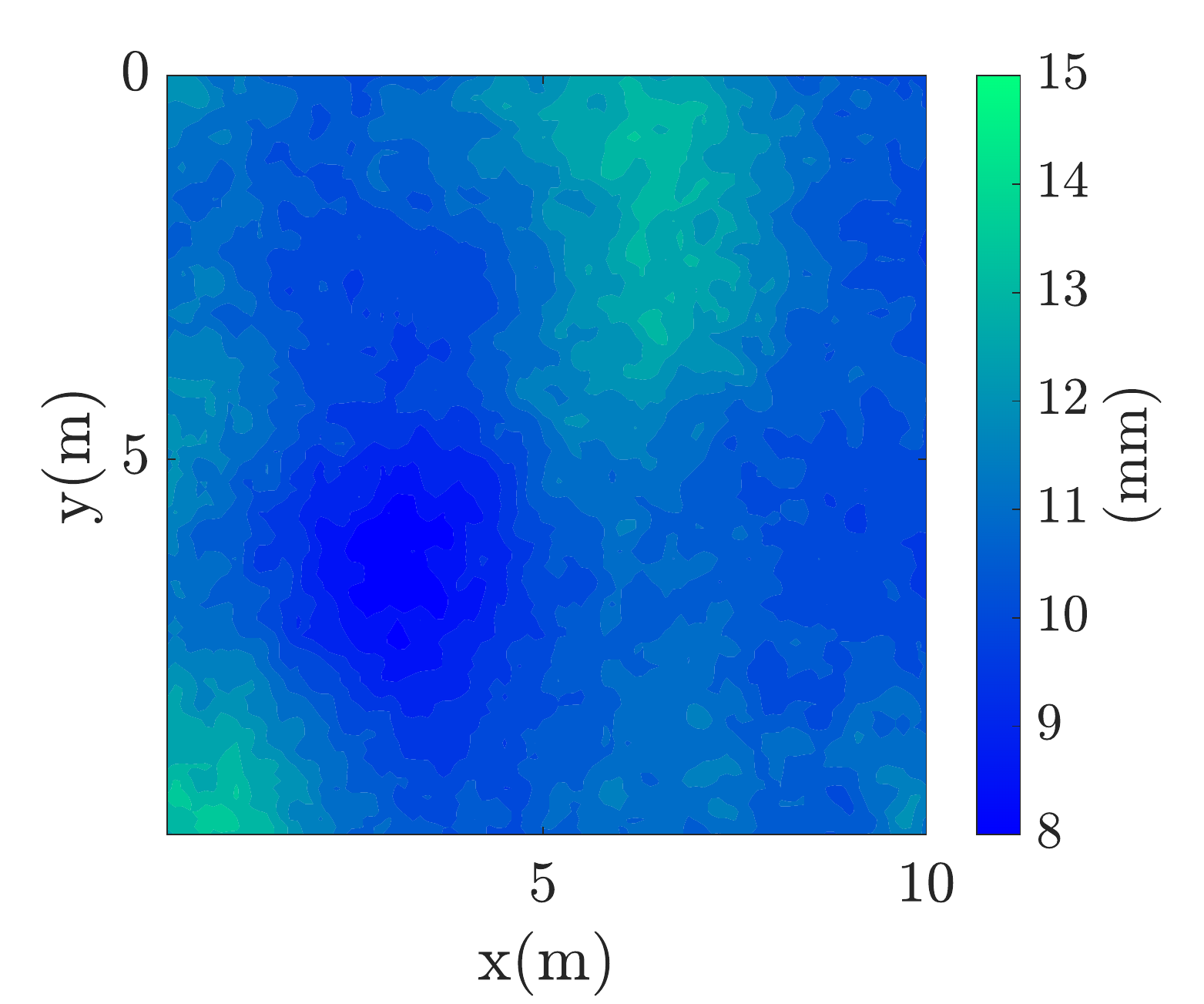}}
	%
	\subfigure[$15$th predicted error.]{
		\label{fig: 2D_predict_error_15}
		\includegraphics[width = 0.23\textwidth]{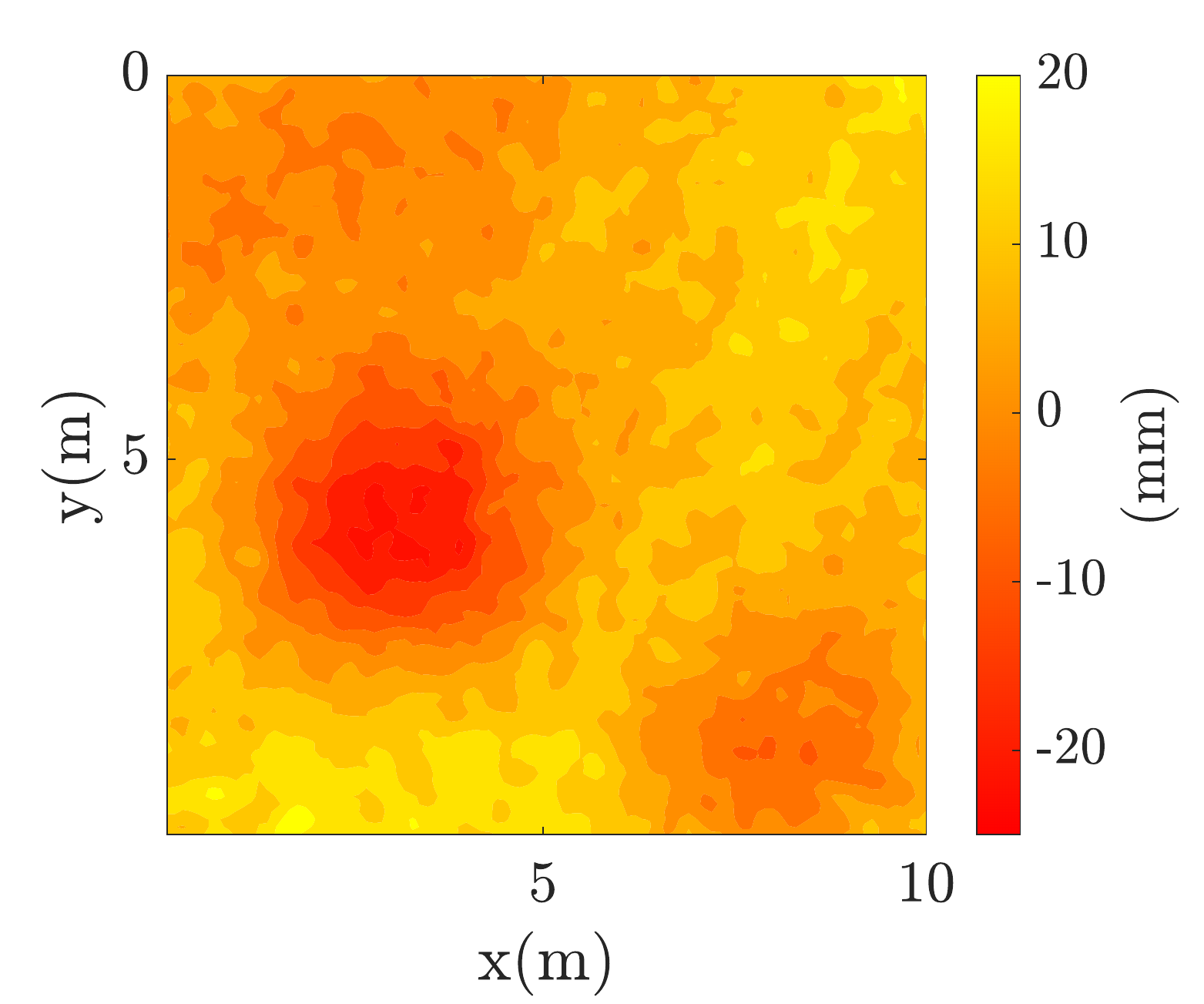}}
	%
	\caption{The comparison between deep learning predictions and the corresponding ground truth. The first row is for $\alpha = 5$ prediction (the effective horizon is $40$ days), and the second row is for $\alpha = 15$ prediction (the effective horizon is $60$ days). The comparisons include predicted mean, predicted standard deviation, and the prediction error for every location.}
	\label{fig: ground truth vs. prediction example}
\end{figure*}

Following the steps described in \secref{sec: Gazebo simulation}, we generate $15$ pastureland environments in Gazebo using the historical data with a day interval of $\delta = 4$. An example illustrating the simulated pastureland environment ($10$m $\times$ $10$m) is shown in \figref{fig: average height pasture}. In this simulated pastureland environment, there are $2.5 \times 10^4$ grass models.

Next, we use a robot to collect a point cloud for each of the simulated environments. The simulated LiDAR has an inherent standard deviation of $4$mm in its readings. We also assume the ground of the field is flat. Nevertheless, it is easy to add different terrains to simulate different types of ground. During the simulations, the plant locations, pose, and plant species for a particular location in the pasture are fixed throughout the years to accelerate those processes. Finally, all the collected point clouds will be used as testing inputs.


\subsection{Neural Network Prediction Results}
\label{ssec: neural networks prediction results}

\emph{Training:}
In the experiments, the training sequences $\X_i$'s are generated by setting the number of measurements in each sequence as $\alpha = 15$. Also, the number of intervals is $\delta = 4$. Note that the training data is generated through the GMM process defined in earlier sections. We use early stopping where training is stopped if validation loss does not improve for ten epochs, usually resulting in 30 training epochs.

\emph{Experimental Evaluation:}
We evaluate the performance of the prediction network on two separate cases: a). The first test case is on the aforementioned point cloud measurements; b). The second evaluation is on two years of the simulated GMM data. 

\begin{figure}[!tbp]
	\centering
	\includegraphics[width=\imwidth]{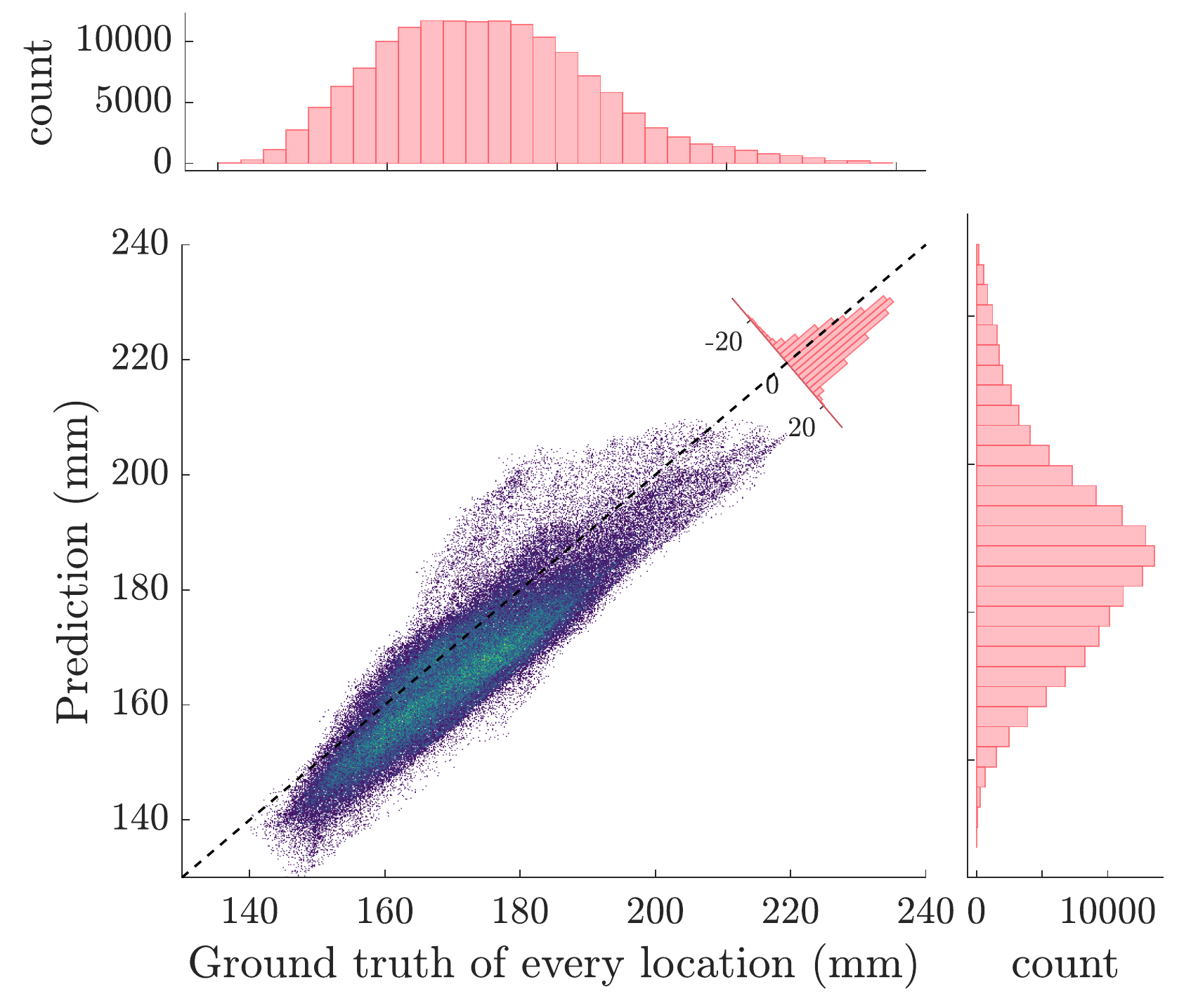}
	\caption{The relationship between the ground truth and the corresponding prediction by using all the predictions and the ground truth. The x-axis represents the ground truth of every point, and the y-axis represents the corresponding prediction result. The marginal histograms show the statistics of the ground truth and the prediction independently.}
	\label{fig: ground truth vs. prediction}
\end{figure}

\begin{figure}[!tbp]
	\centering
	\includegraphics[width=\imwidth]{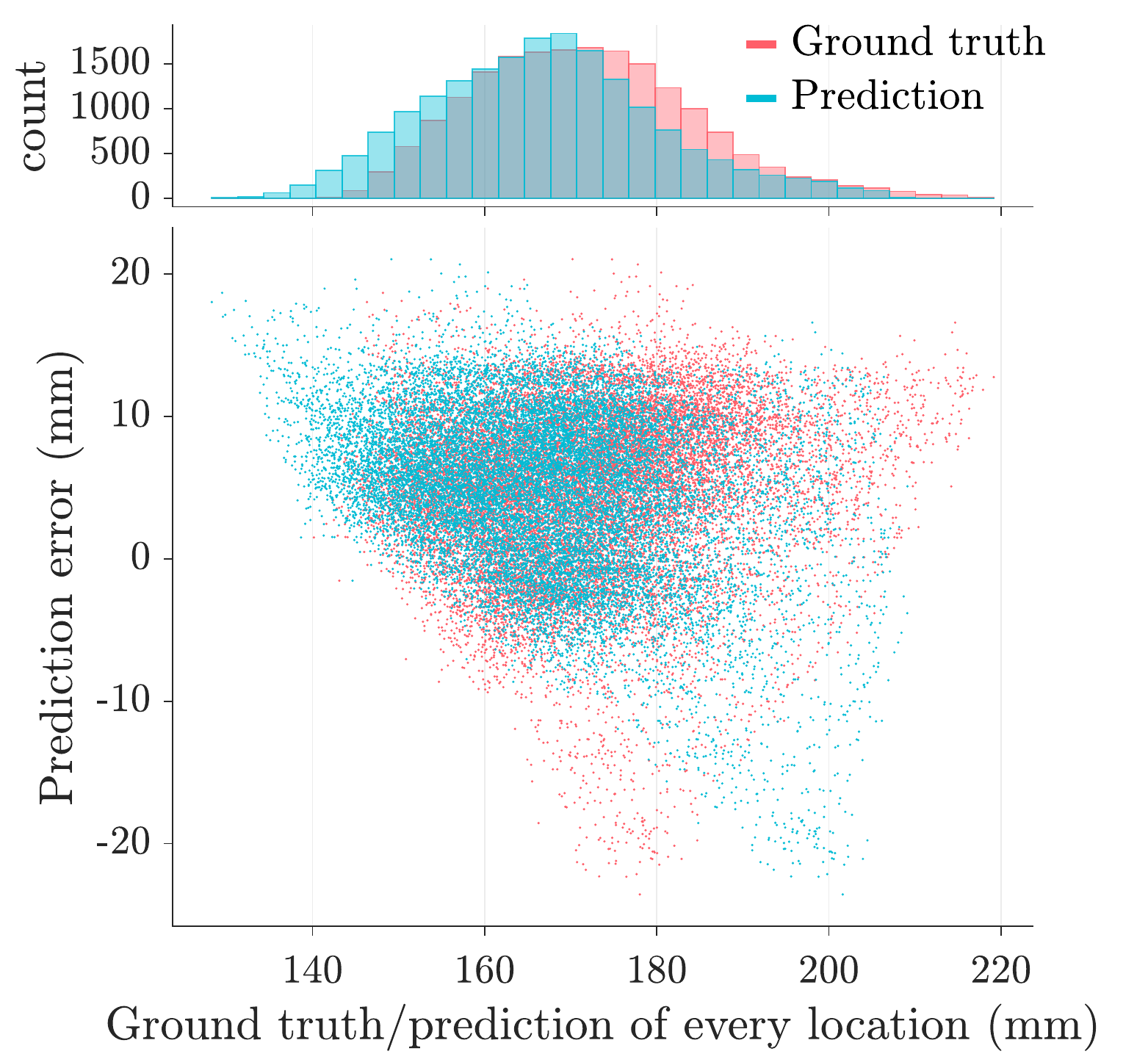}
	\caption{The comparison between ground truth/prediction and the corresponding prediction errors of every location.}
	\label{fig: ground truth and prediction vs. error}
\end{figure}

\emph{Testing by using point cloud measurements:}
We note that due to the computationally intensive process of constructing aforementioned pastures and its subsequent point cloud measurements, we only generate a single test case consisting of $30$ days, where the data is split into $15$ days as an input sequence to the prediction network, and the remaining $15$ days are used to evaluate the prediction performance.
The collected point clouds will be first converted to heightmaps of size $100 \times 100$ as the original point clouds are too large to be used as network prediction inputs. Meanwhile, we need to process those heightmaps since the measurements are noisy. In \figref{fig: processed point cloud}, we demonstrate a downsampled surface of the point cloud shown in \figref{fig: point cloud of pasture}. Specifically, after downsampling, the processing includes two steps: (a). a size of $3 \times 3$ median filtering; (b). a size of $3 \times 3$ flat convolution filtering. After processing, we see that the growth pattern of the pastureland is visible as shown in \figref{fig: contour processed} when compared with the raw map as shown in \figref{fig: contour raw}. Those processes make the neural networks' prediction of the dynamics of the pastureland more efficient. Meanwhile, this processing only has mild change on the original data, as can be seen from the histogram comparison in \figref{fig: hist comparison}.

\begin{figure}[!tbp]
	\centering
	\includegraphics[width=\imwidth]{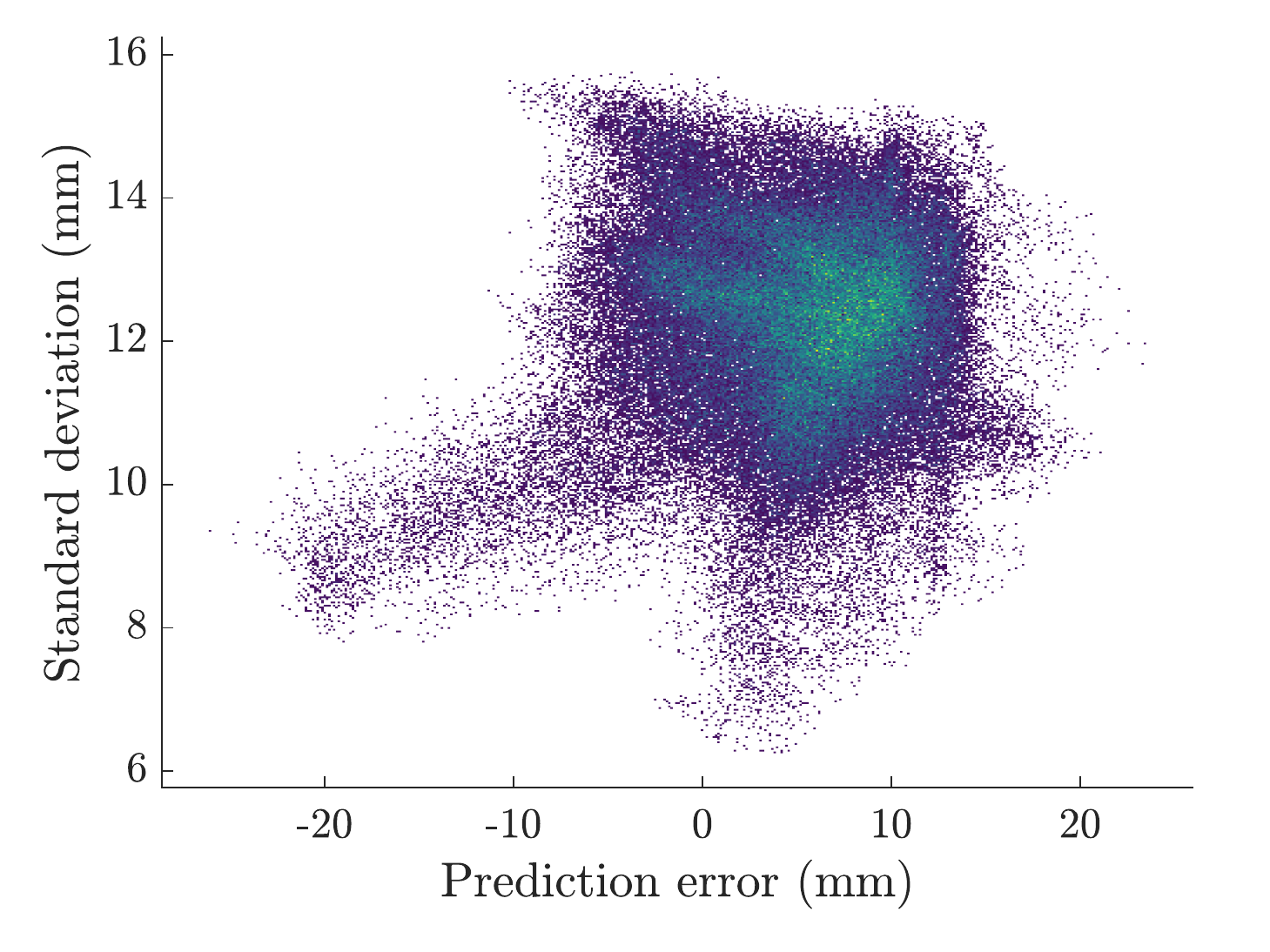}
	\caption{The relationship between the prediction error and the corresponding prediction standard deviation of every location by using all the predictions.}
	\label{fig: error vs std}
\end{figure}

\begin{table*}[!t]
	\centering
	\begin{threeparttable}
		\caption{\label{tab: performance metrics} Accuracy metrics (in mm) for different methods with and without uncertainty estimates (UE), where the stride length is $\delta=\{1, 4\}$ and the cardinality of horizons set is $\alpha=15^*$.}
		\small
		\centering
		\begin{tabular}{l c c c c c c c c c c c}
			\hline\noalign{\smallskip}
			                & \multicolumn{4}{c}{(GMM Data)} & \multicolumn{4}{c}{(Point Cloud Data)}                                                                                          \\
			Model           & RMSE                           & MAE                                    & MPAE          & ASTD          & RMSE          & MAE           & MPAE           & ASTD  \\
			\noalign{\smallskip}\hline\noalign{\smallskip}
			$\delta=4$ + UE & 19.25                          & 13.42                                  & 11.91         & 9.27          & \textbf{7.41} & \textbf{6.46} & \textbf{3.672} & 12.39 \\
			$\delta=1$ + UE & \textbf{6.91}                  & \textbf{5.12}                          & \textbf{4.65} & \textbf{8.31} & --            & --            & --             & --    \\
			$\delta=4$      & 24.65                          & 18.79                                  & 15.62         & --            & 19.31         & 18.07         & 10.58          & --    \\
			$\delta=1$      & 18.52                          & 14.38                                  & 13.13         & --            & --            & --            & --             & --    \\
			\noalign{\smallskip}\hline
		\end{tabular}
		\begin{tablenotes}
			\footnotesize
			\item $^*$(Since the tests without UE do not implement MC Dropout methods, no ASTD values are available. Additionally, due to high computation requirements for point cloud data, results for $\delta=1$ are unavailable.)
		\end{tablenotes}
	\end{threeparttable}
\end{table*}

To test the prediction performance, we predict the heightmaps for another $15$ days using a stride length of $\delta = \{1, 4\}$. The longest effective horizon for those two are $L = 15$ and $L = 60$ respectively. Then, we generate the corresponding point cloud measurement to check the testing performance. To estimate the uncertainties, we set the number of MC dropout samples as $K = 500$. In general, our network performance on average performs within a $12\%$ error rate in the worst case when $\delta=4$ and $L = 60$, and within a $5\%$ error rate when data is available more frequently when $\delta=1$ and $L = 15$. In \figref{fig: boxplot}, we plot the comparison between the ground truth and the prediction when the prediction time index $\alpha = 1, 5, 10$, and $15$ using a stride length of $\delta = 4$. The empirical results show that the proposed method performs sufficiently well in predicting and reconstructing pasture heights, even for long-horizon ($L = 60$ days) problems. To further demonstrate the details of our long-time horizon prediction results, we show the $5$th and $15$th prediction results with their ground truth in \figref{fig: ground truth vs. prediction example}.

Note that the effective time horizons for those two predictions are $L = 40$ and $60$ days. The uncertainty estimations in terms of standard deviations of the predictions run over $K = 500$ samples, and we observe that the network has higher confidence at the highest pasture heights since it is easier to learn feature mappings due to larger correlations within its neighborhood. The prediction error is lowest at the highest points in the pasture since the network learns to estimate the peak points with higher confidence.

In \figref{fig: ground truth vs. prediction}, we compare the ground truth and the corresponding predictions using all the predictions across the output sequence. From the result, we can see that most prediction-ground truth pairs lie in the $45$ degree line, where we can clearly observe that most predictions are close to the ground truth.
In \figref{fig: ground truth and prediction vs. error}, we demonstrate the relationship between the prediction errors and the ground truth/prediction for each location, which reinforces the performance of the neural network, where we observe the predictions are quite close to the ground truth.
The prediction uncertainty of the model increases as the prediction error increases, as shown in \figref{fig: error vs std}, showing a clear correlation between the confidence of the network against the prediction errors. This reiterates the suitability of the computationally efficient Bayesian approximation for uncertainty estimates, allowing a quick and efficient turnaround in prediction. This methodology reduces the need for expensive hardware for training and inference of the deep learning-based prediction model and allows even small industries or farms to tune and integrate the prediction systems into their workflow. 

\begin{remark}
Note that in the above tests, the network makes predictions generated from point cloud measurements, and the original training dataset of pasture height maps generated from GMM is not used. This indicates the ability of our network to generalize beyond the simulated training data to real-world applications by using LIDAR-based measurements.
\end{remark}

\begin{figure}[!tbp]
	\centering
	\includegraphics[width=\imwidth]{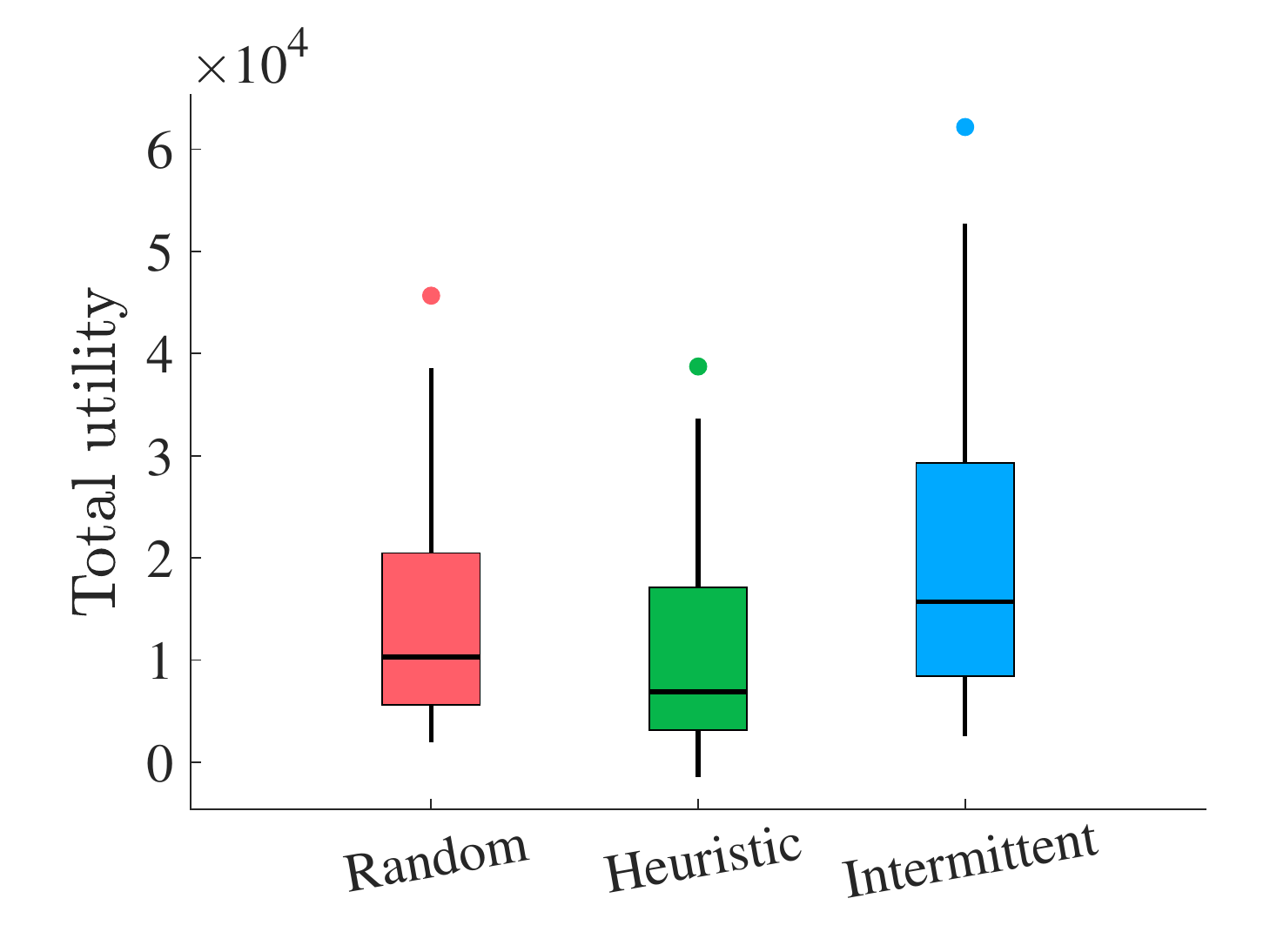}
	\caption{The utility statistics of the three methods after running $50$ trials. The proposed intermittent deployment method has the highest averaged utility.}
	\label{fig: boxplot of utility comparison}
\end{figure}

\emph{Testing by using GMM data:}
Next, we use Monte Carlo simulations to test the prediction performance using different GMM inputs. Note that the training model is unchanged in this case. We set $|\T_y| = \alpha = 15$. That is, we use a length of $15$ heightmaps to predict another group of $15$ heightmaps. Since we will test the performance using two years' data and the first $15$ heightmaps can only be used as inputs, the number of distinct output prediction sequences is $H = (365-|\T_y|) \cdot 2$. We should also note that $K$ in \eqref{eq: prediction mean} is the number of samples used for calculating one prediction $\bar\YY_t, \forall t \in \T_y$.
We evaluate the performance of the neural network predictions with the following metrics: Root Mean Square Error (RMSE), Mean Absolute Error (MAE), Mean Absolute Percentage Error (MAPE), and Averaged Standard Deviation (ASTD).
\begin{equation*}
	\begin{aligned}
		\text{RMSE} & = \sqrt{\frac{1}{H} \sum_{h=1}^{H} \sum_{t \in \T_y} (\YY_t^{(h)} - \bar{\YY}_t^{(h)})^2},                   \\
		\text{MAE}  & = \frac{1}{H} \sum_{h=1}^{H} \sum_{t \in \T_y} \left| \YY^{(h)}_t - \bar{\YY}^{(h)}_t \right|,               \\
		\text{MAPE} & = \frac{100}{H} \sum_{h=1}^{H} \sum_{t \in \T_y} \left| \frac{\YY_t^{(h)} - \bar{\YY}^{(h)}_t}{\YY^{(h)}_t} \right|, \\
		\text{ASTD} & = \sqrt{\frac{1}{H} \sum_{h=1}^{H} \sum_{t \in \T_y} \left(\bar\sigma^{(h)}_t\right)^2}, \\
	\end{aligned}
	\label{eq: prediction metric}
\end{equation*}
where $\YY^{(h)}_t$ is the $h$th prediction at time $t$, and $\bar\sigma_t^{(h)}$ is the corresponding standard deviation. Note that the each mean prediction $\bar\YY_t$ is calculated by using $K = 500$ samples, i.e., $\bar\YY_t = \frac{1}{K} \sum_{t=1}^K \hat{\YY}_t$. The results of these metrics are reported in \tabref{tab: performance metrics}, containing the results from the above two testings. These metrics quantify the performance of the prediction algorithm by aggregating the errors and uncertainty over the discretized values of the pasture.


\subsection{Pipeline Comparison Results}
\label{ssec: simulation results}

\begin{figure}[!tbp]
	\centering
	\subfigure[Collected uncertainty.]{
		\label{fig: boxplot distance}
		\includegraphics[width = 1.55in]{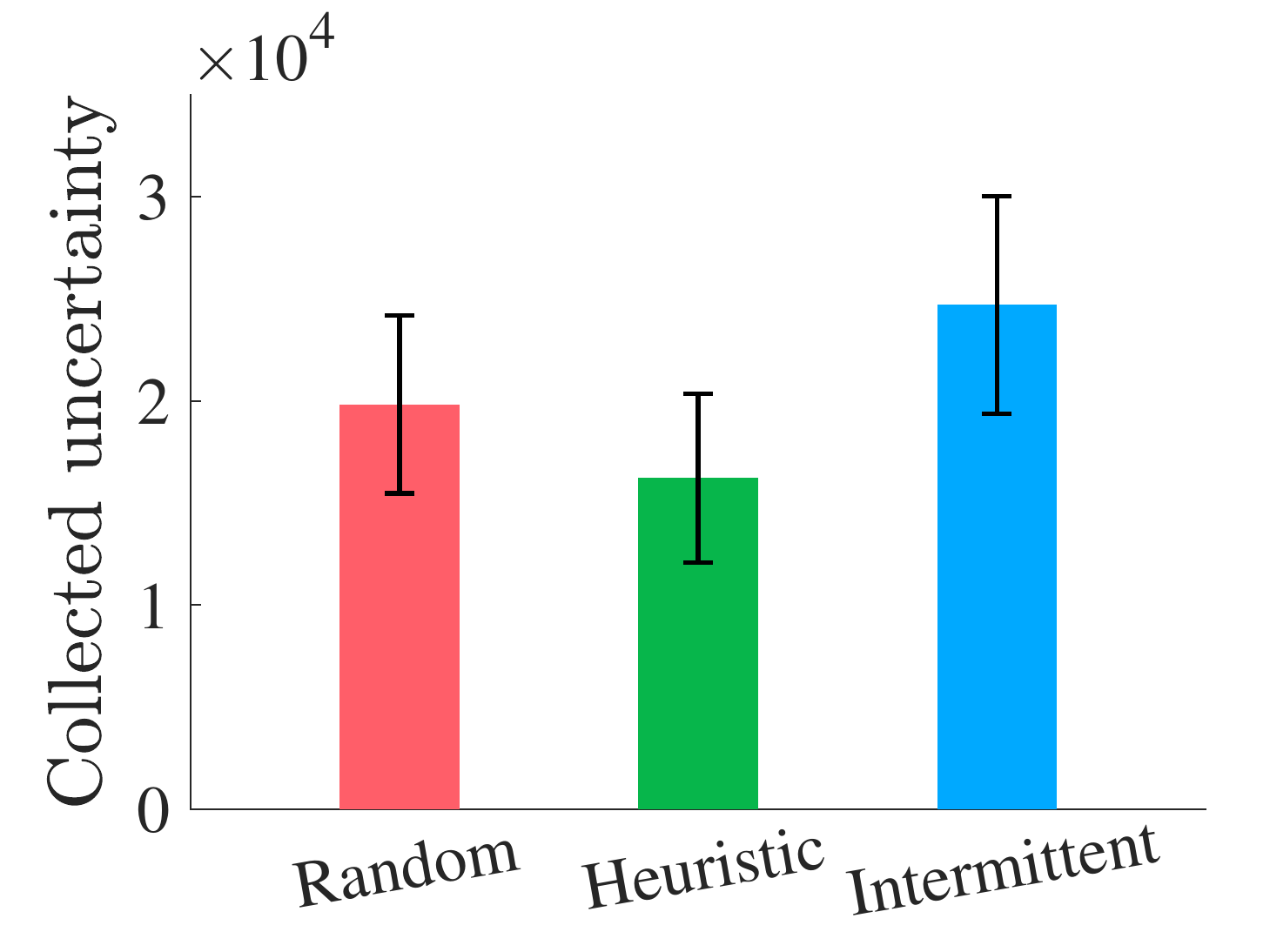}}
	\subfigure[Waiting penalty.]{
		\label{fig: boxplot waiting}
		\includegraphics[width = 1.55in]{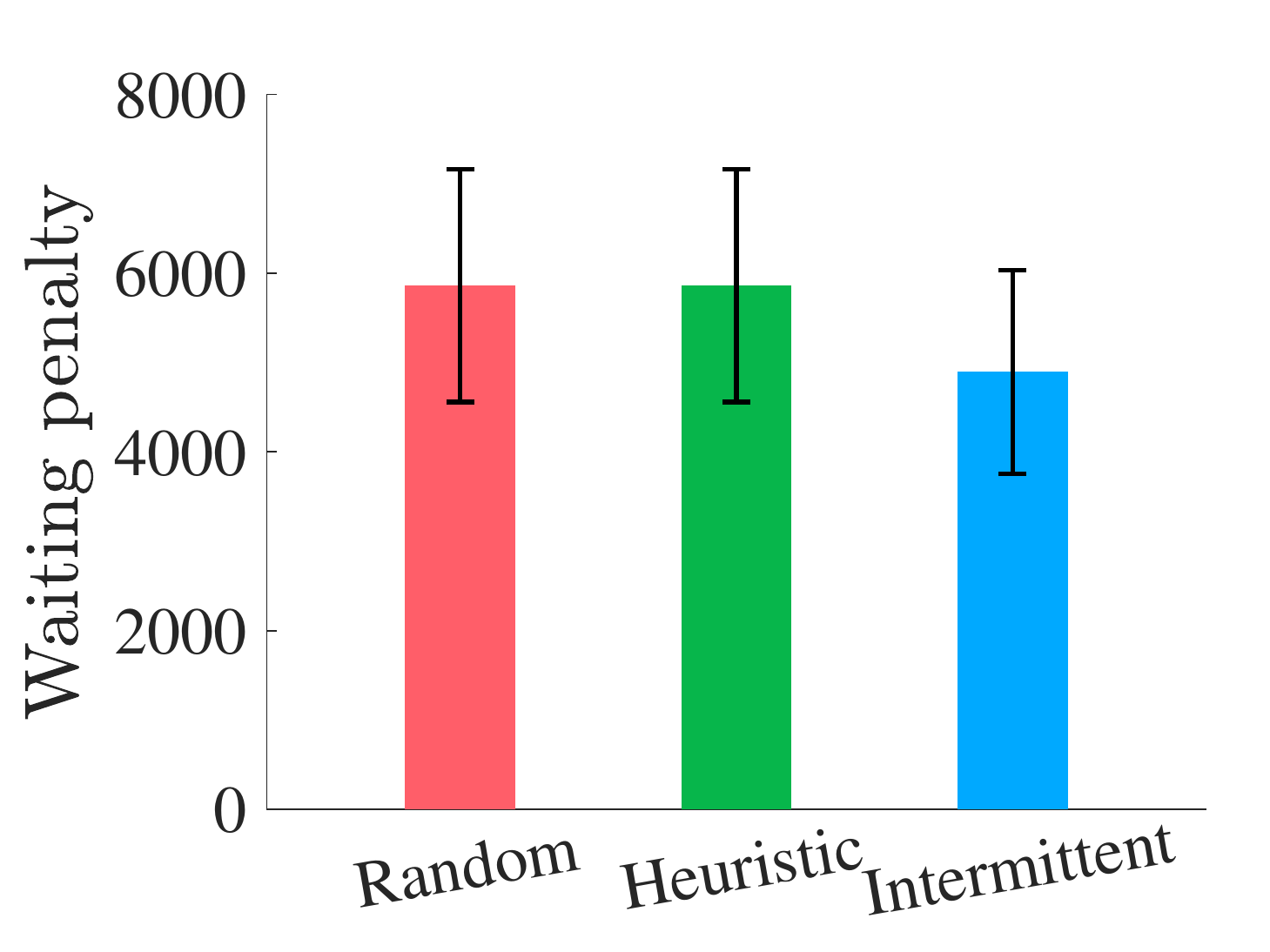}}
	\caption{(a). The comparison of the collected uncertainties of three different methods, which is the first part of the objective function. (b). The waiting penalty of three different methods, which is the second part of the objective function.}
	\label{fig: boxplot of individual utility comparisons}
\end{figure}

In this section, we use Monte Carlo simulations to test the performance of the proposed integrated pipeline with another pipeline for sensing and deployment purposes based on GMM data. The size of the heightmaps used for the plannings will be $100 \times 100$ as before. Specifically, we have two goals for the testings: (a). we want to test the effectiveness of the proposed pipeline when the deployments are different. (b). We want to test the effectiveness of the proposed pipeline when the prediction methods are different.

\emph{The first comparison:}
Based on the same deep learning prediction result, we compare the performance using the following deployment policies:
\begin{itemize}
	\item The intermittent deployment policy, where the policy is generated through the method proposed in \secref{sec: decision making}. We refer to this policy as ``Intermittent''.
	\item A fixed interval deployment policy, where the robots are deployed within a fixed time interval. Also, the deployment locations are evenly distributed across the pastureland (mimicking how humans manually monitor pasturelands). We refer to this policy as ``Heuristic''.
	\item A random deployment policy, where both deployment times and locations are randomly selected from the ground set $\V$. We refer to this policy as ``Random''.
\end{itemize}
We first evaluate the performance using the collected reward (the objective function value). Then, we collect samples based on the generated deployment policies. After that, we use the collected information from different methods to make $10$ more predictions. We finally compare the prediction performance of those methods. Thus, those steps will help verify the proposed pipeline's effectiveness.

\emph{The second comparison:}
We adopt a commonly used mutual information-based pipeline in the second comparison category and compare it with our proposed pipeline. Specifically, we will use a 2D Gaussian process (GP)
to model the environment and make predictions. We then use the proposed intermittent deployment idea to make deployment strategies based on the different prediction results. Again, we finally make another $10$ future predictions after deployments to test the performance.

\emph{The results from the first comparison:}
Based on the $\alpha = 15$ deep learning prediction results, we run $50$ Monte Carlo simulations to generate different deployments under different settings. The settings are as follows. The maximum number of deployable days (total budget) $\ell$ is randomly sampled from the set $\{5, 6, \ldots, 12\}$. The number of maximum sampling points $\ell_t$ is sampled from the set $\{2^2, 3^2, \ldots, 8^2\}$. The cardinality of the planning horizon is $|\T_y| = 15$. We also assign each robot a random weight for the same traveling cost to simulate the heterogeneity. In each run, we generate one instance and then use the above three methods to compare the performance. Also, the number of robots is set to $\ell_t \cdot \ell$ for each instance. The weights for distance and time are $w_2 = 0.1, w_3 = 1$. And the weight for time penalty is $w_1 = 5$. The definitions of those parameters can be found in our problem formulation in \secref{ssec: problem formulation}.

\begin{figure}[!tbp]
	\centering
	\includegraphics[width=\imwidth]{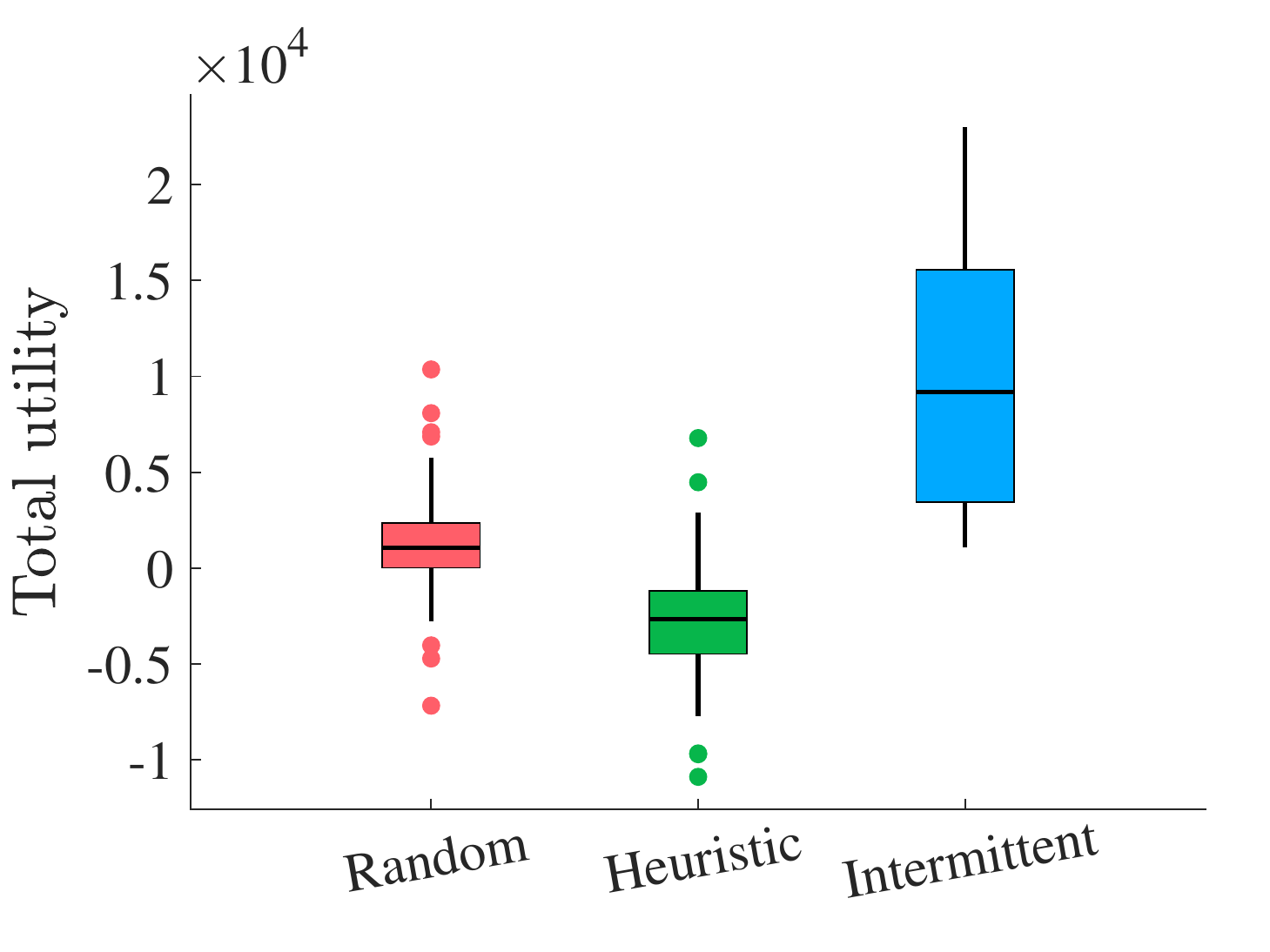}
	\caption{The utility statistics of the three methods after running $50$ trials when the waiting penalty weight $w_1$ is high.}
	\label{fig: boxplot of utility comparison high waiting penalty}
\end{figure}

\begin{figure}[!tbp]
	\centering
	\subfigure[Collected uncertainty.]{
		\label{fig: boxplot distance high waiting penalty}
		\includegraphics[width = 1.55in]{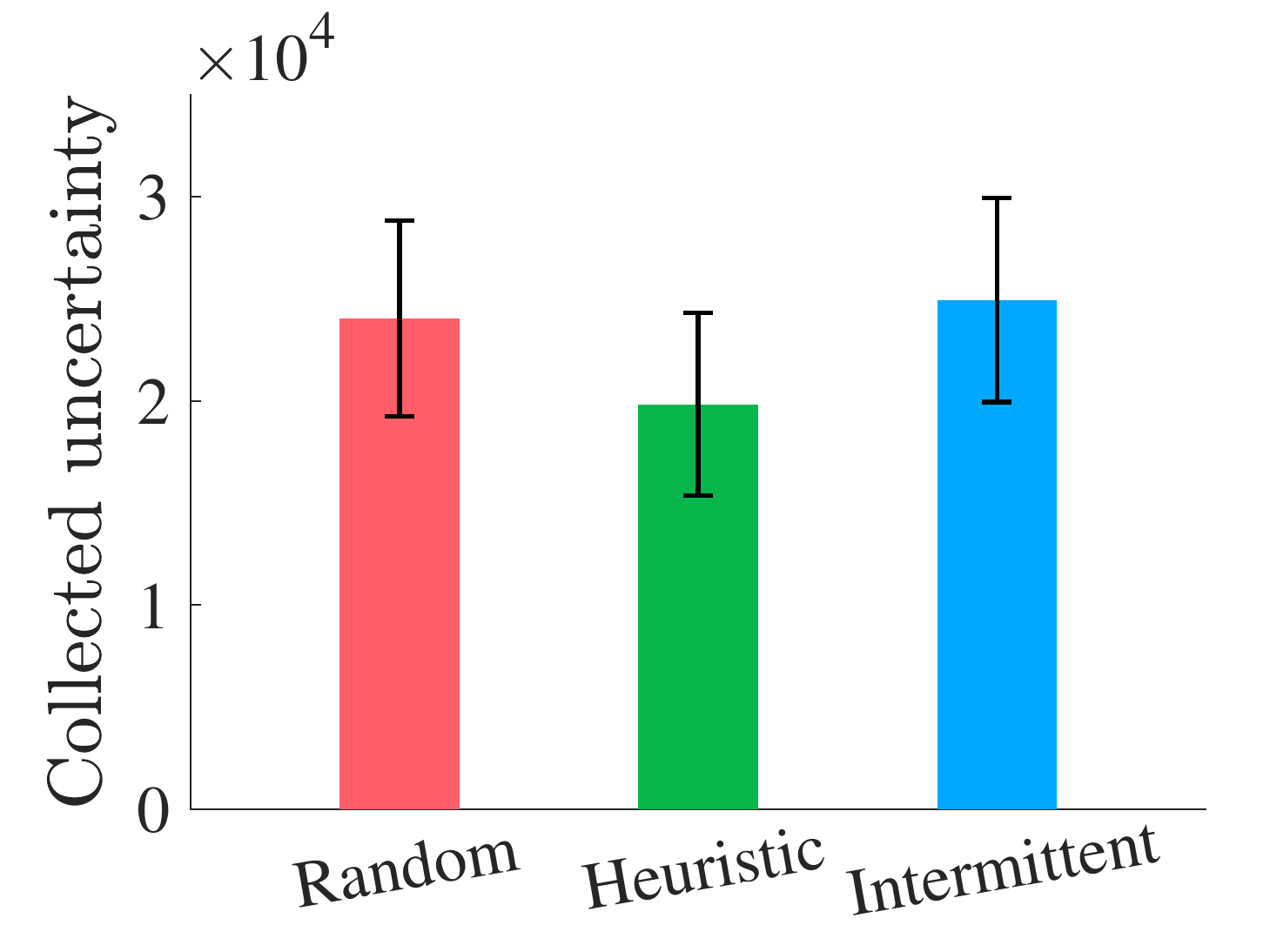}}
	\subfigure[Waiting penalty.]{
		\label{fig: boxplot waiting high waiting penalty}
		\includegraphics[width = 1.55in]{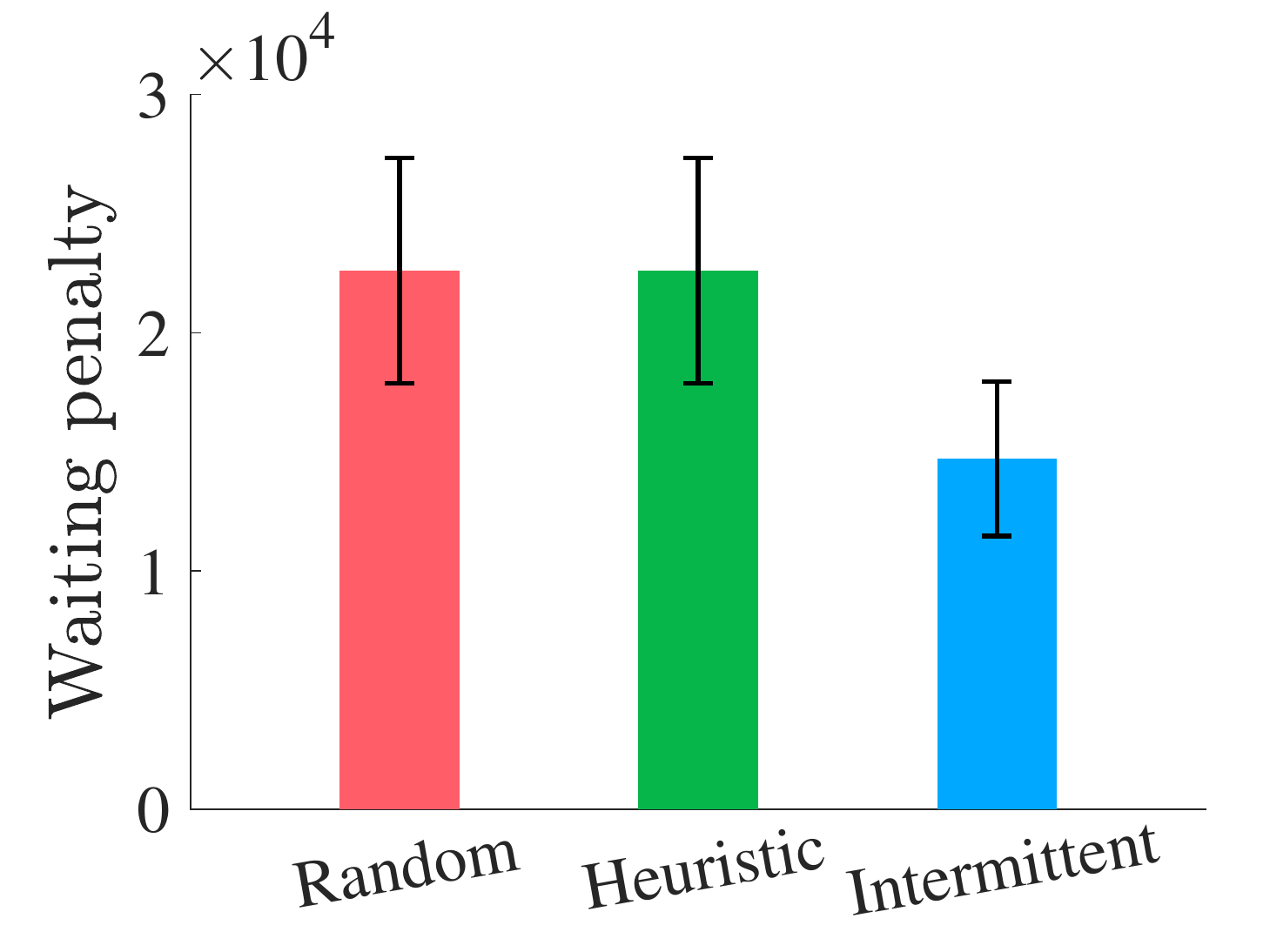}}
	\caption{The utility/cost of different parts of our objective function when the waiting penalty weight $w_1$ is high.}
	\label{fig: boxplot of individual utility comparisons high waiting penalty}
\end{figure}

\begin{figure*}[!t]
	\centering
	\subfigure[Random 1st deployment]{
		\label{fig: random 1}
		\includegraphics[width = .31\linewidth]{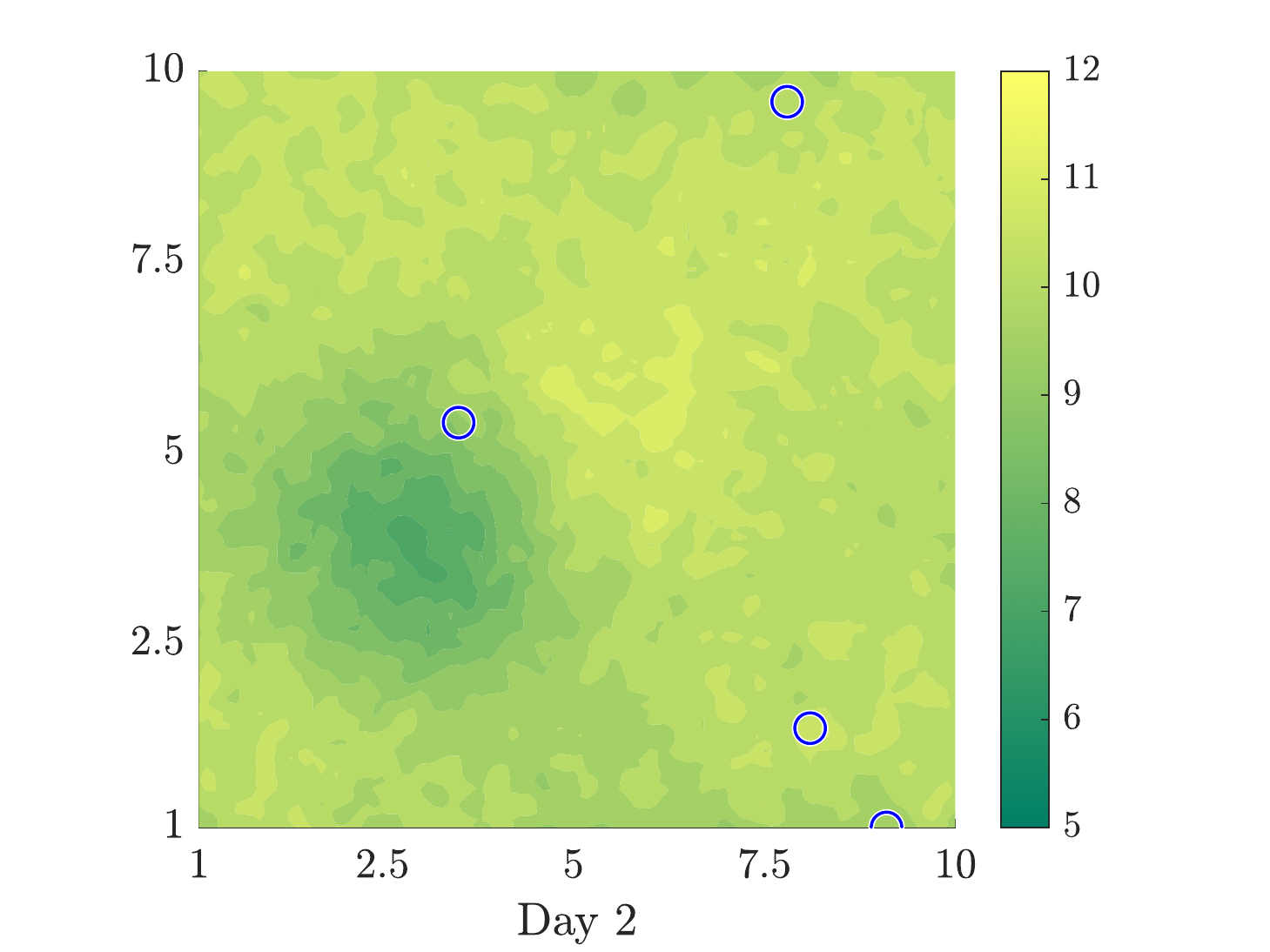}}
	\subfigure[Random 2nd deployment]{
		\label{fig: random 2}
		\includegraphics[width = .31\linewidth]{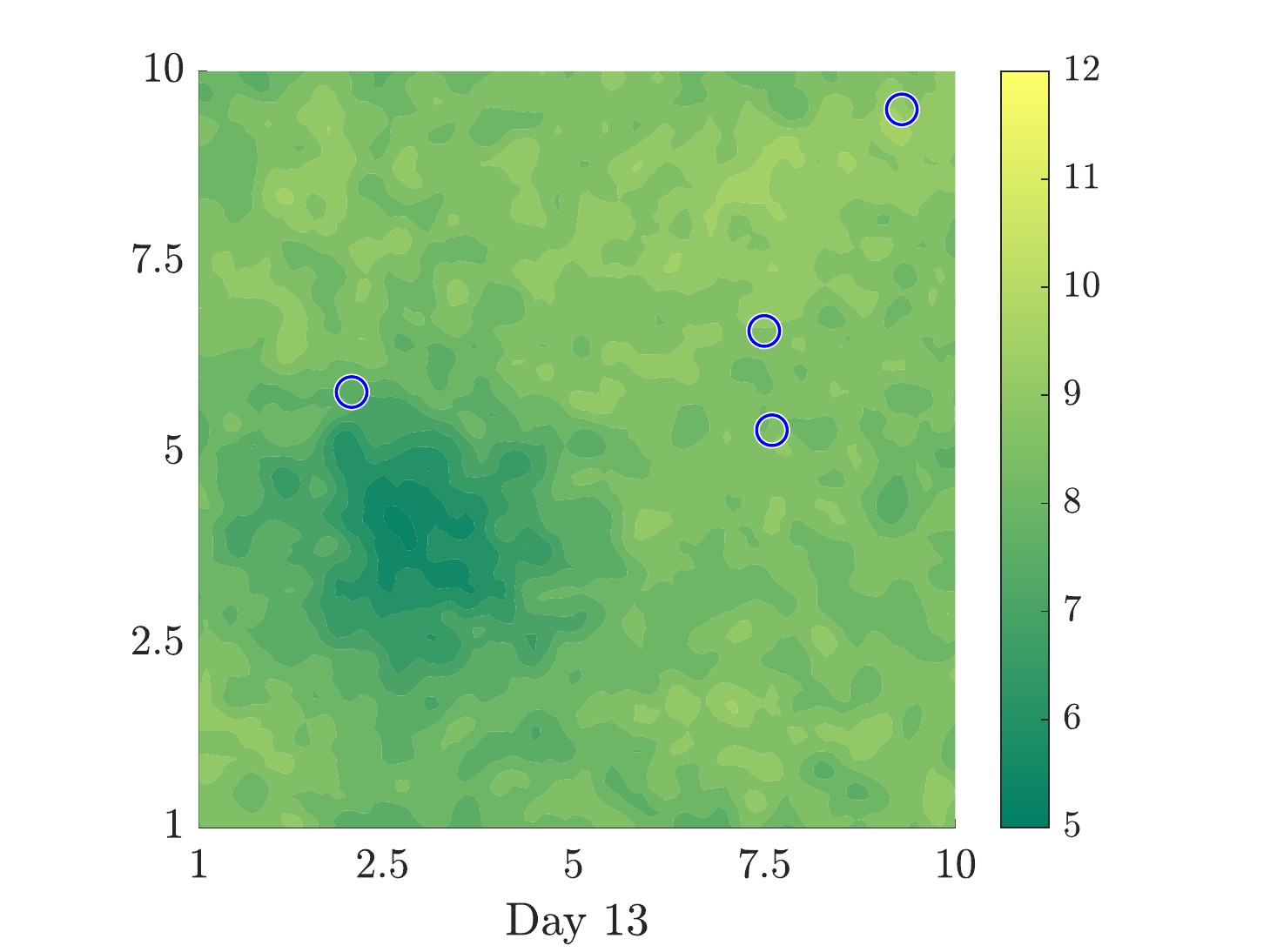}}
	\subfigure[Random 3rd deployment]{
		\label{fig: random 3}
		\includegraphics[width = .31\linewidth]{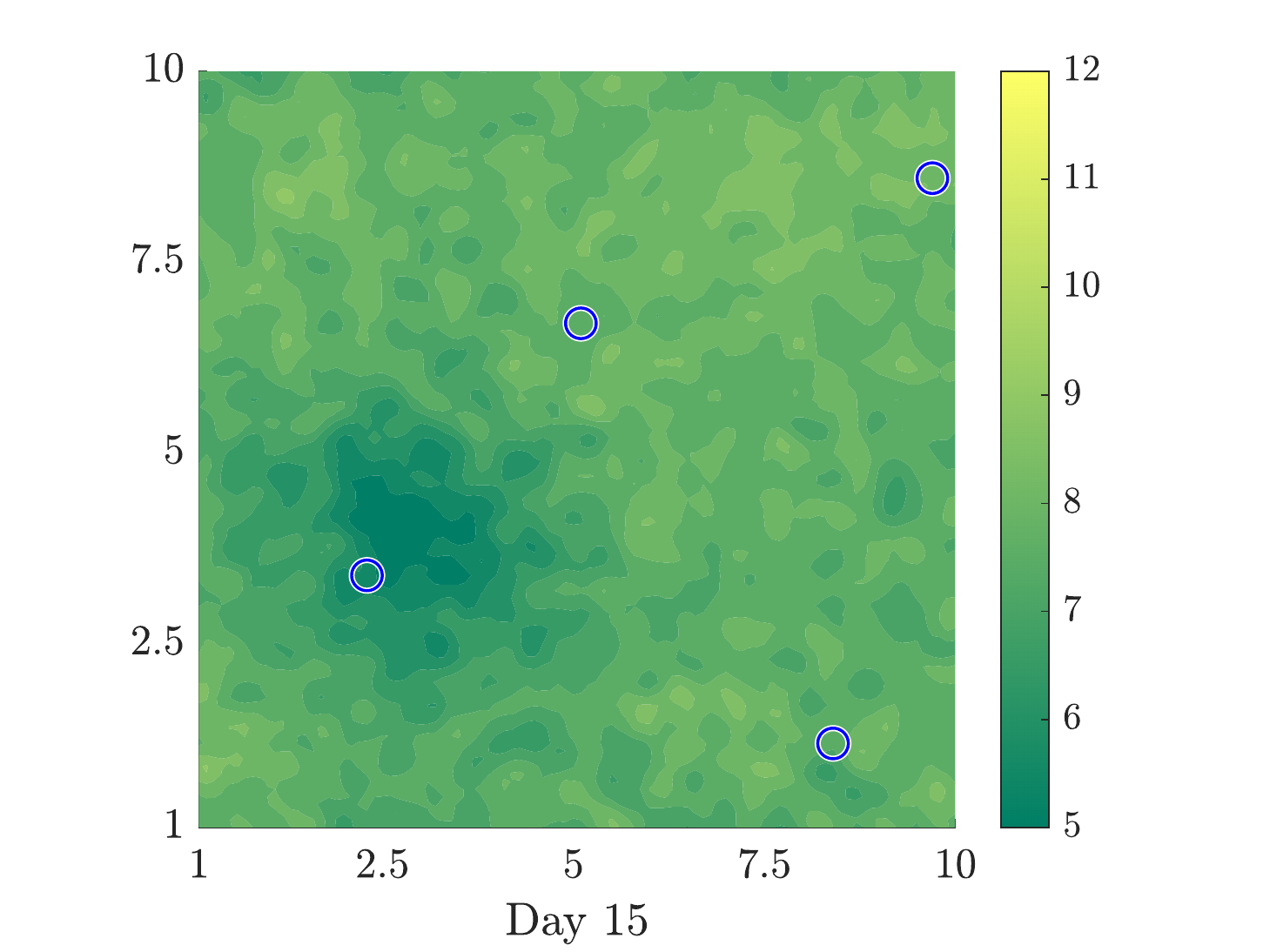}}
	\subfigure[Heuristic 1st deployment]{
		\label{fig: heuristic 1}
		\includegraphics[width = .31\linewidth]{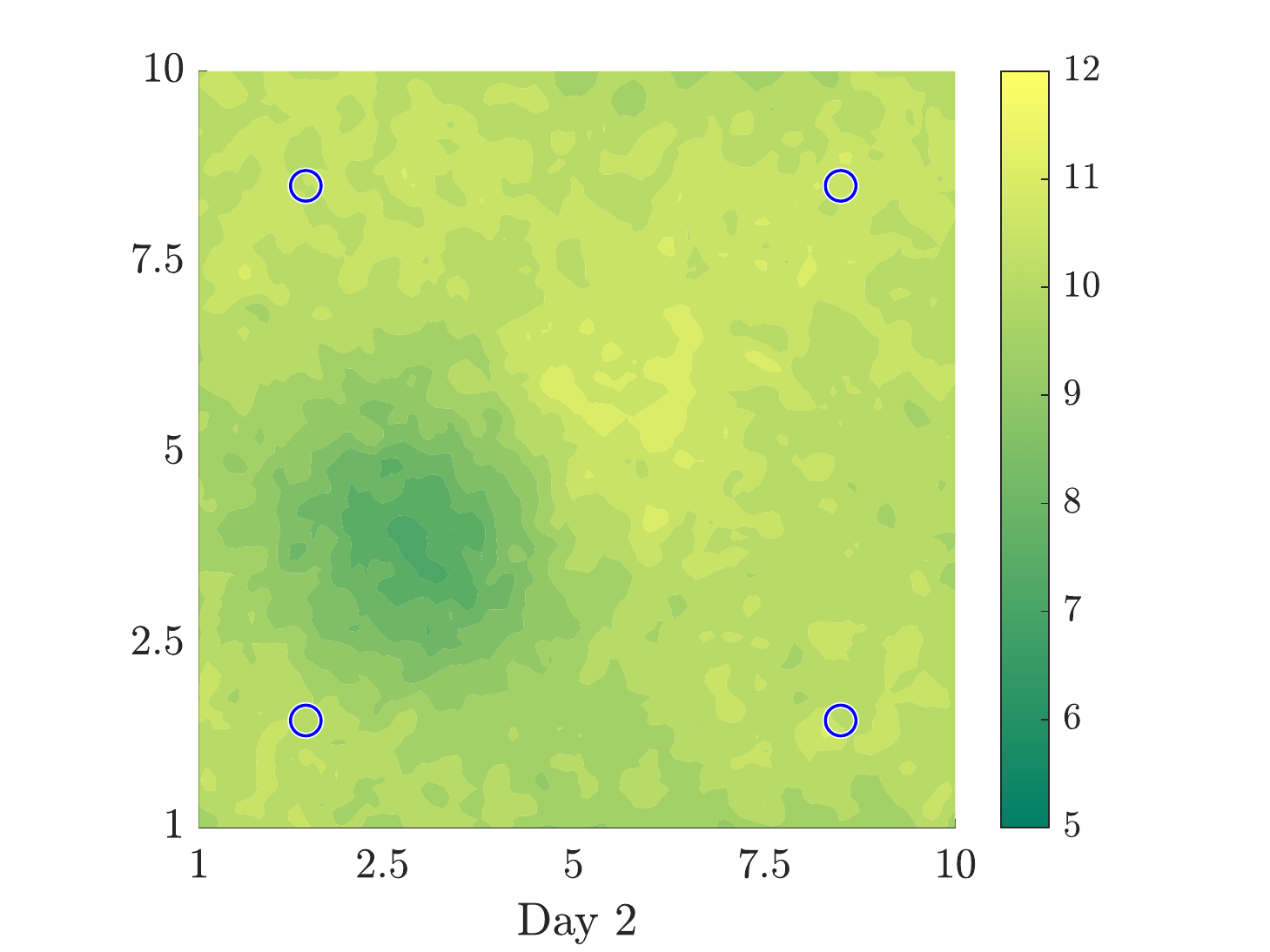}}
	\subfigure[Heuristic 2nd deployment]{
		\label{fig: heuristic 2}
		\includegraphics[width = .31\linewidth]{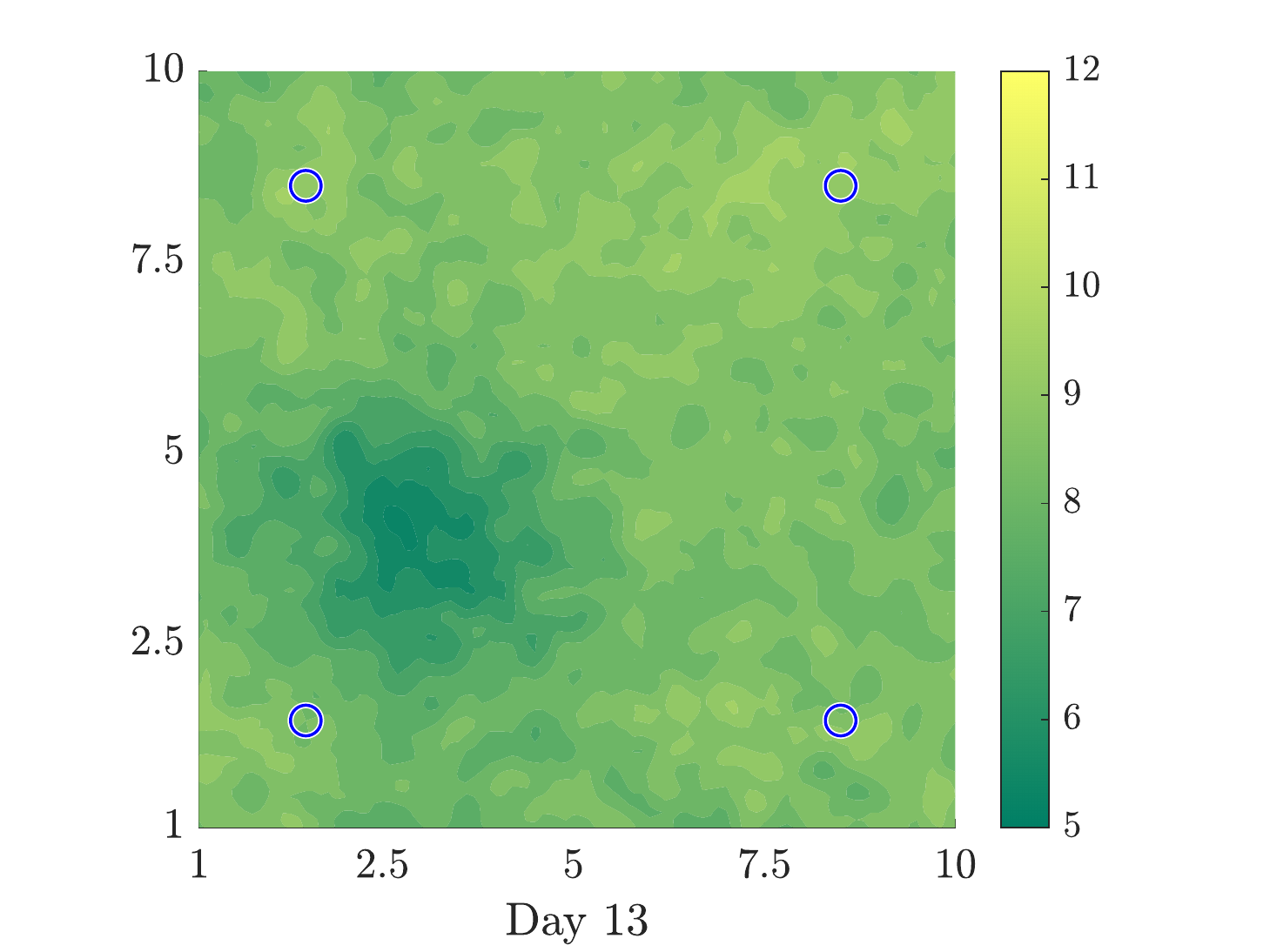}}
	\subfigure[Heuristic 3rd deployment]{
		\label{fig: heuristic 3}
		\includegraphics[width = .31\linewidth]{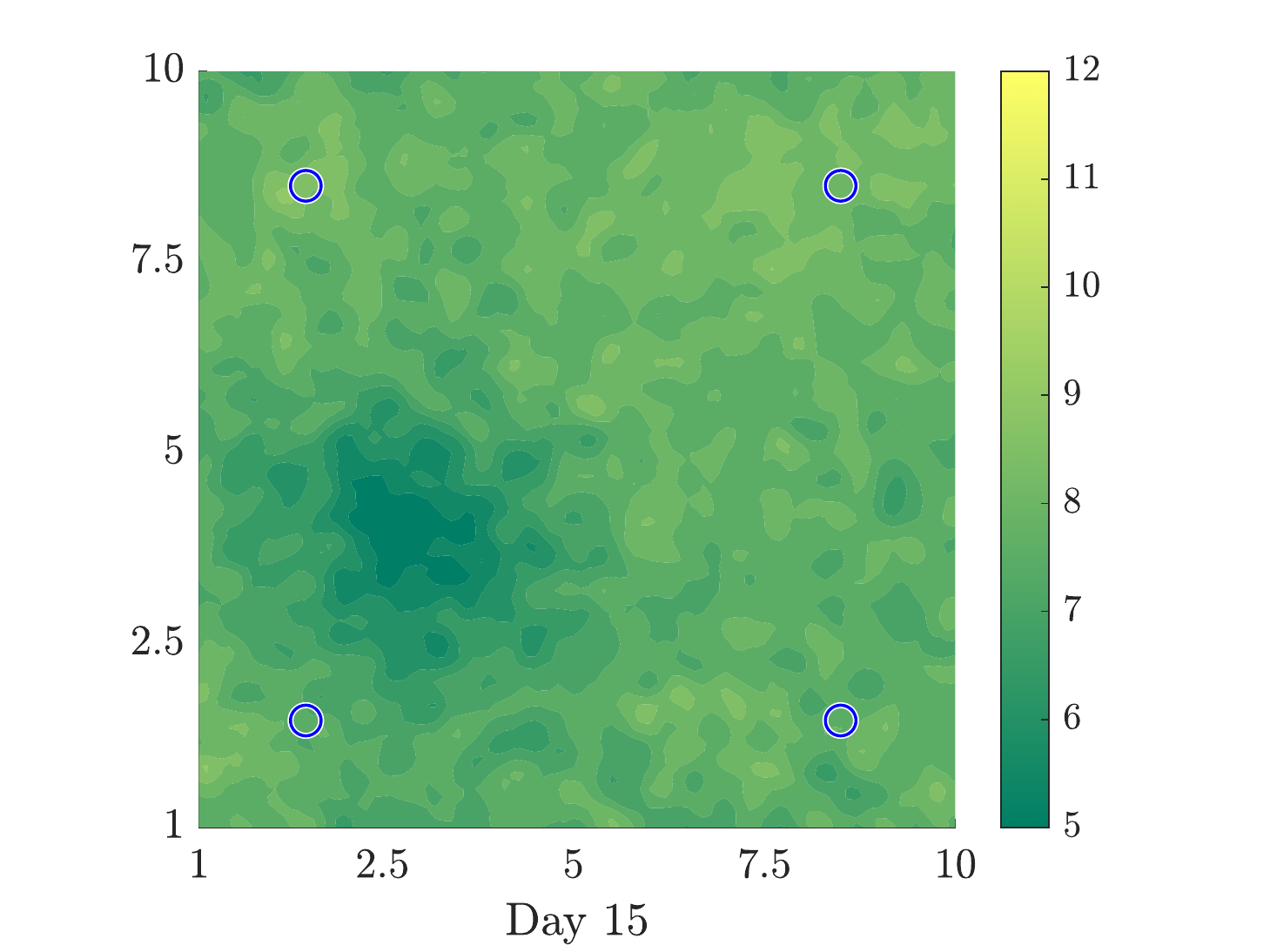}}
	\subfigure[Intermittent 1st deployment]{
		\label{fig: intermittent 1}
		\includegraphics[width = .31\linewidth]{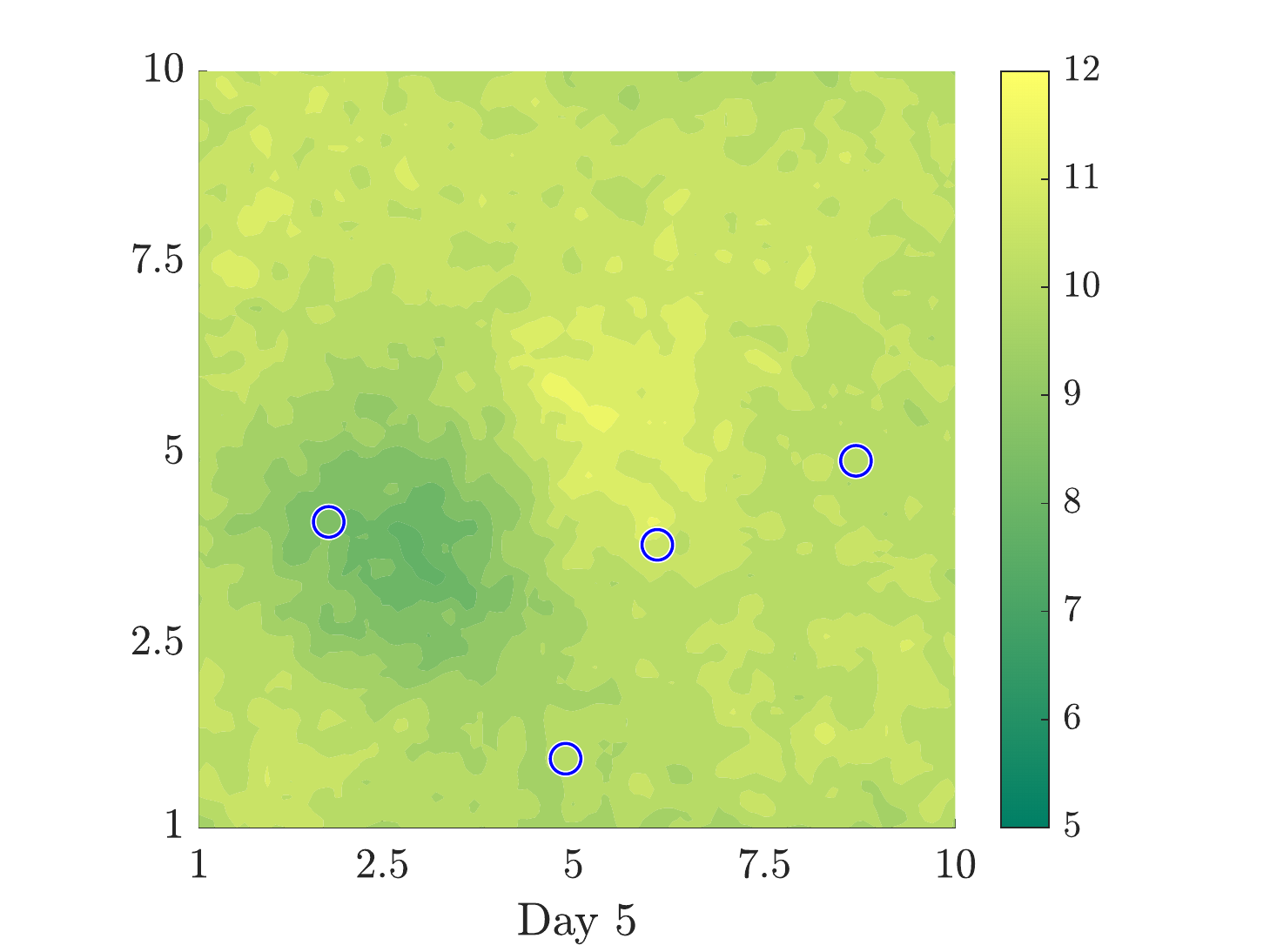}}
	\subfigure[Intermittent 2nd deployment]{
		\label{fig: intermittent 2}
		\includegraphics[width = .31\linewidth]{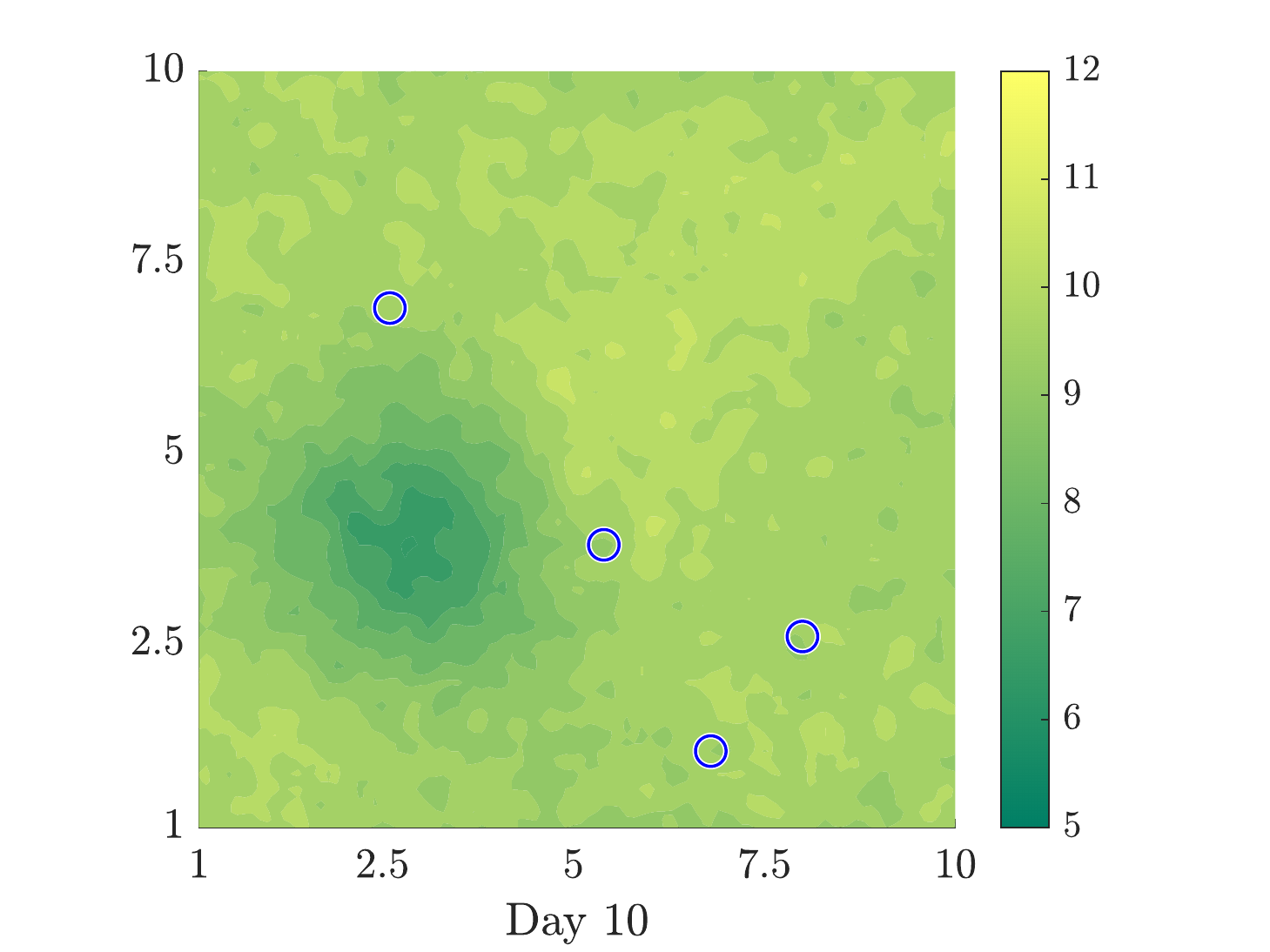}}
	\subfigure[Intermittent 3rd deployment]{
		\label{fig: intermittent 3}
		\includegraphics[width = .31\linewidth]{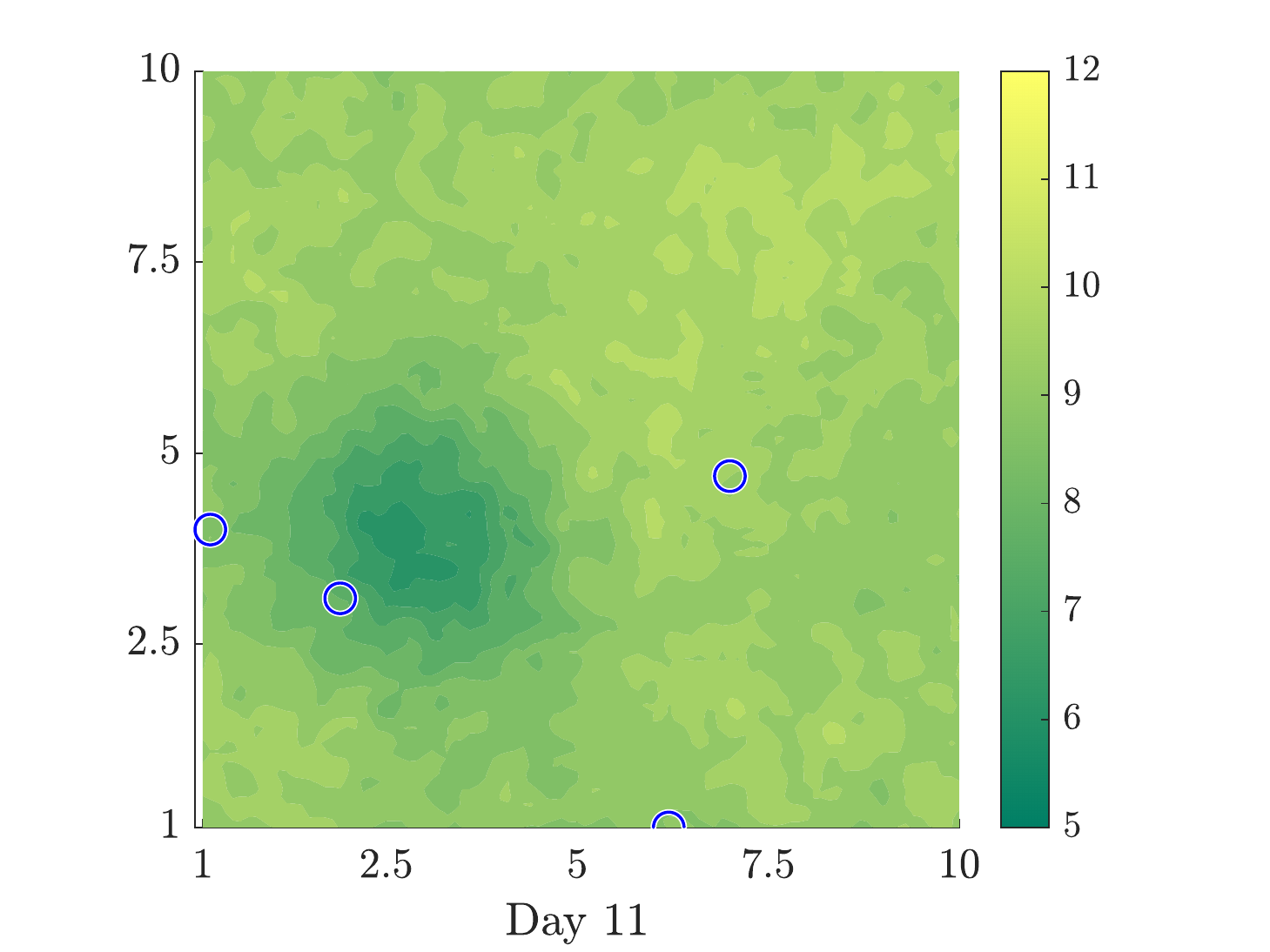}}
	\caption{An example of deployment policies generated by different methods. (a)-(c). random, (d)-(f). heuristic, (g)-(i). intermittent. Those are predicted variance maps.}
	\label{fig: deployment results}
\end{figure*}

\begin{figure}[!tbp]
	\centering
	\includegraphics[width=\imwidth]{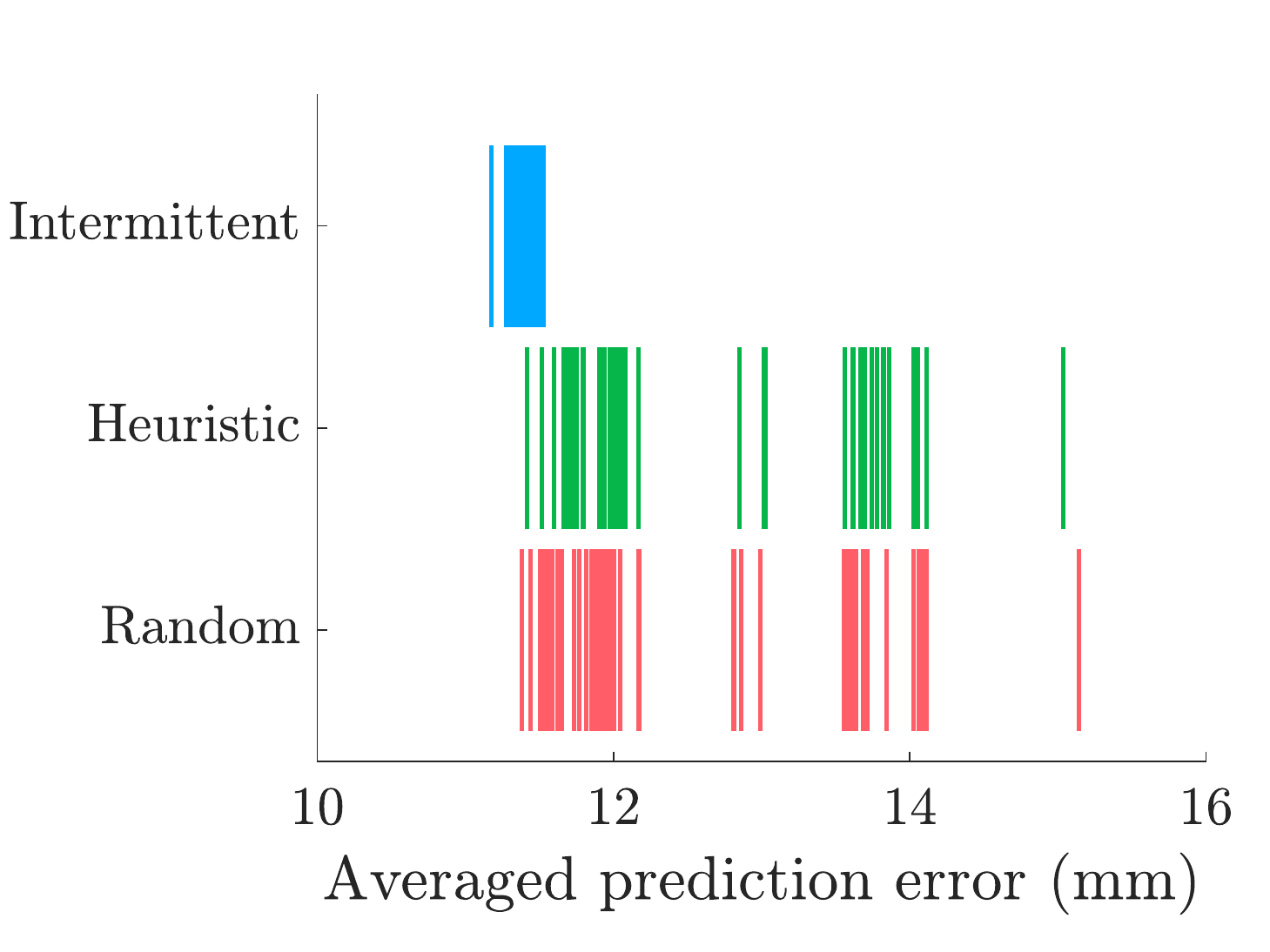}
	\caption{The comparison of the averaged prediction error by calculating the mean of the errors at each location using $50$ trials.}
	\label{fig: raster plot of error comparison}
\end{figure}

Then, we first compare the results of different deployment policies by comparing the collected reward of each method after running $50$ trials. The result is shown in \figref{fig: boxplot of utility comparison}, where the proposed intermittent policy has the highest averaged utility. Then, in \figref{fig: boxplot of individual utility comparisons}, we demonstrate the two different parts of the objective function. In \figref{fig: boxplot distance}, we show the collected uncertainty of three different deployment methods, which is the first part of the objective function. In \figref{fig: boxplot waiting}, we compare the waiting penalty part in the objective function. Then, in \figref{fig: boxplot of utility comparison high waiting penalty}, we set a high waiting penalty as $w_1 = 10$ while keeping $w_2, w_3$ the same as before. Again, we run other $50$ trials to compare the performance. In \figref{fig: boxplot of utility comparison high waiting penalty}, the values of different parts in the objective function prove that our proposed method can collect more rewards while having less waiting penalty. Similarly, in \figref{fig: boxplot of individual utility comparisons high waiting penalty}, we compare the result from each part of the objective function using three different deployment policies. This finishes up the first part of this comparison. In \figref{fig: deployment results}, we demonstrate a deployment result using one setting instance. In this result, we set the parameters as follows. The maximum number of deployments for each day is set to $\ell_t = 4$, and the maximum number of deployable days is set to $\ell = 3$.

\begin{figure*}[!tbp]
	\centering
	\subfigure[$5$th ground truth.]{
		\includegraphics[width = .3\linewidth]{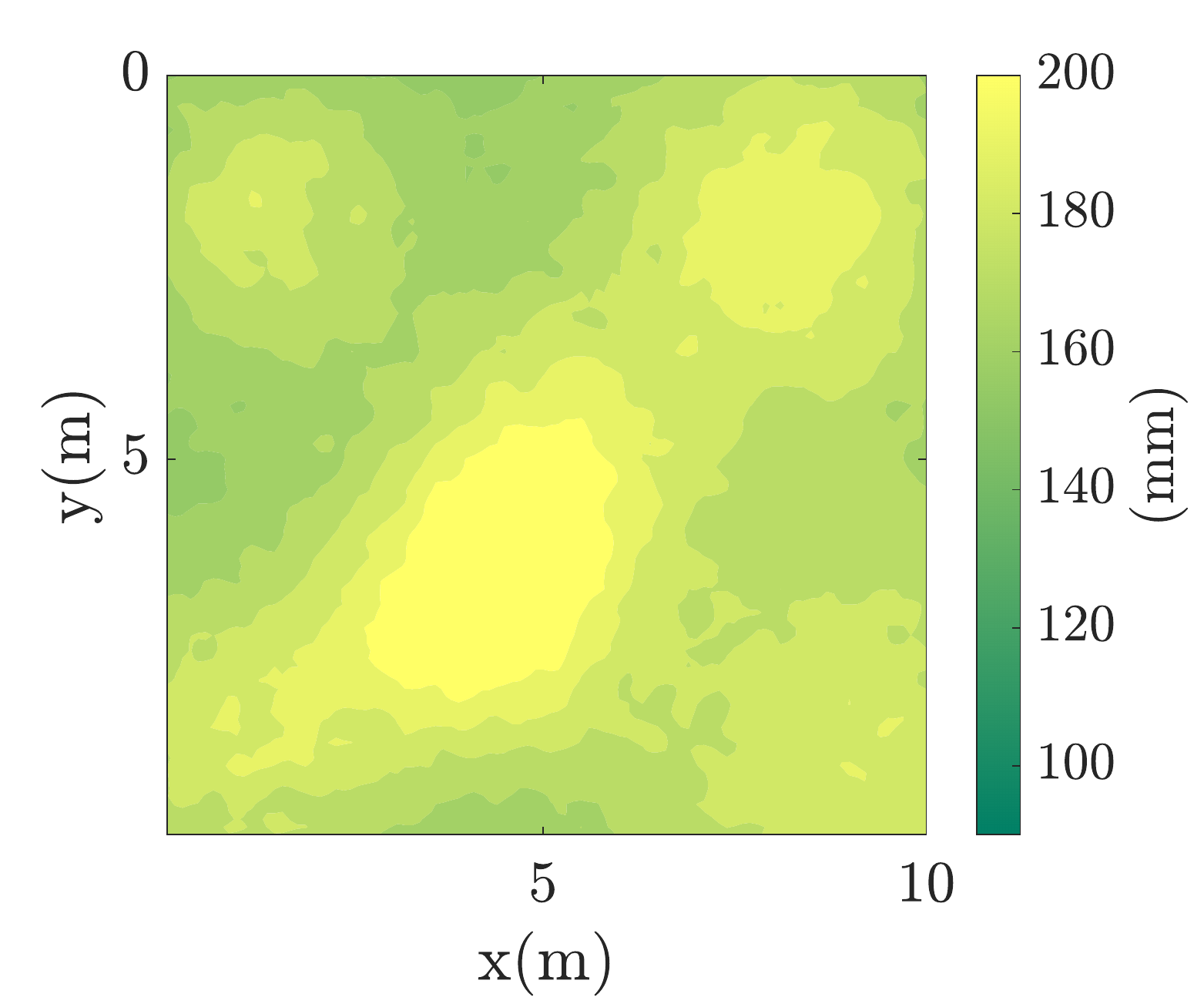}}
	\subfigure[$5$th predicted using DL.]{
		\includegraphics[width = .3\linewidth]{2D_predict_mean_5.pdf}}
	\subfigure[$5$th prediction using GP.]{
		\includegraphics[width = .3\linewidth]{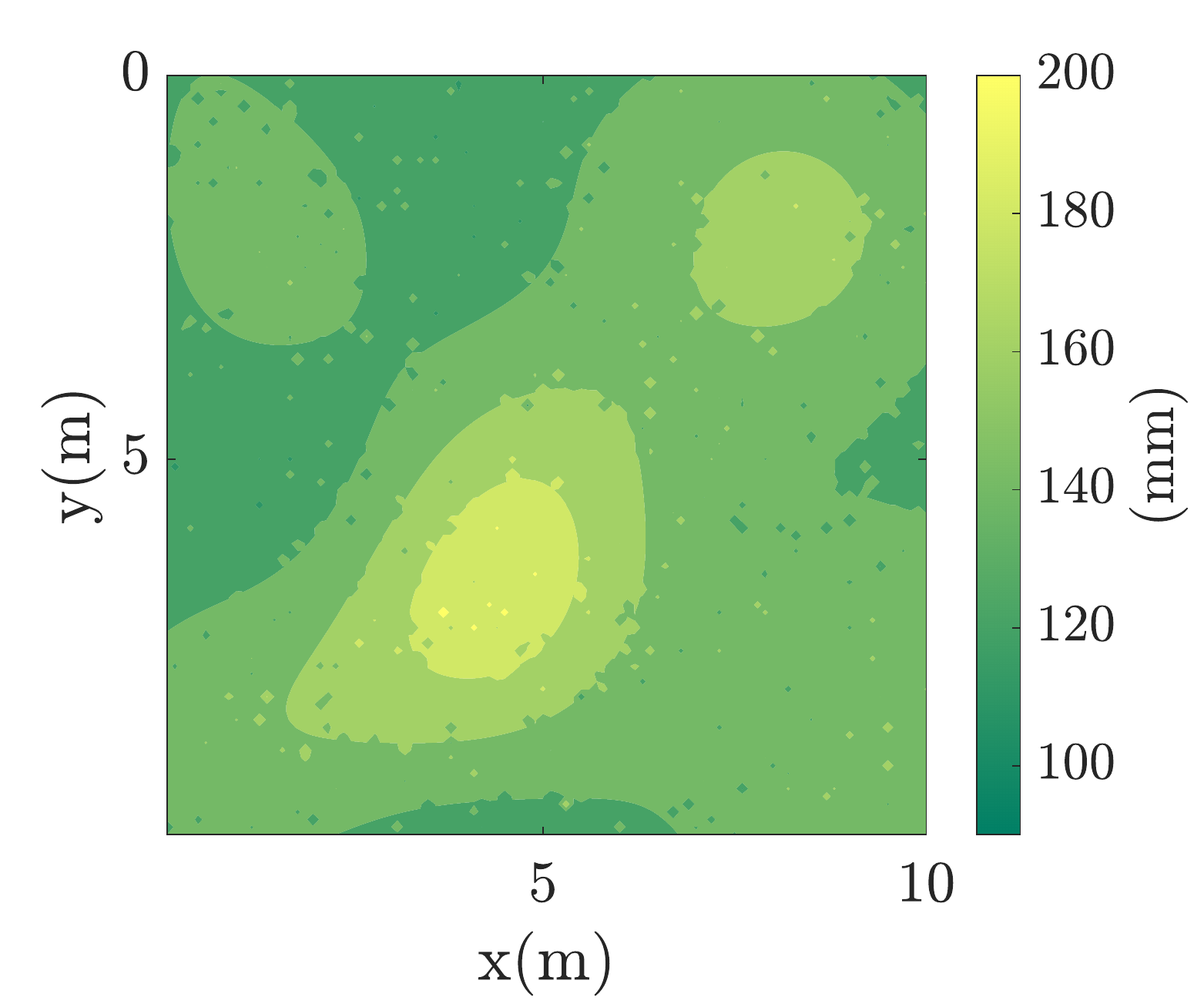}} \\
	\subfigure[$15$th ground truth.]{
		\includegraphics[width = .3\linewidth]{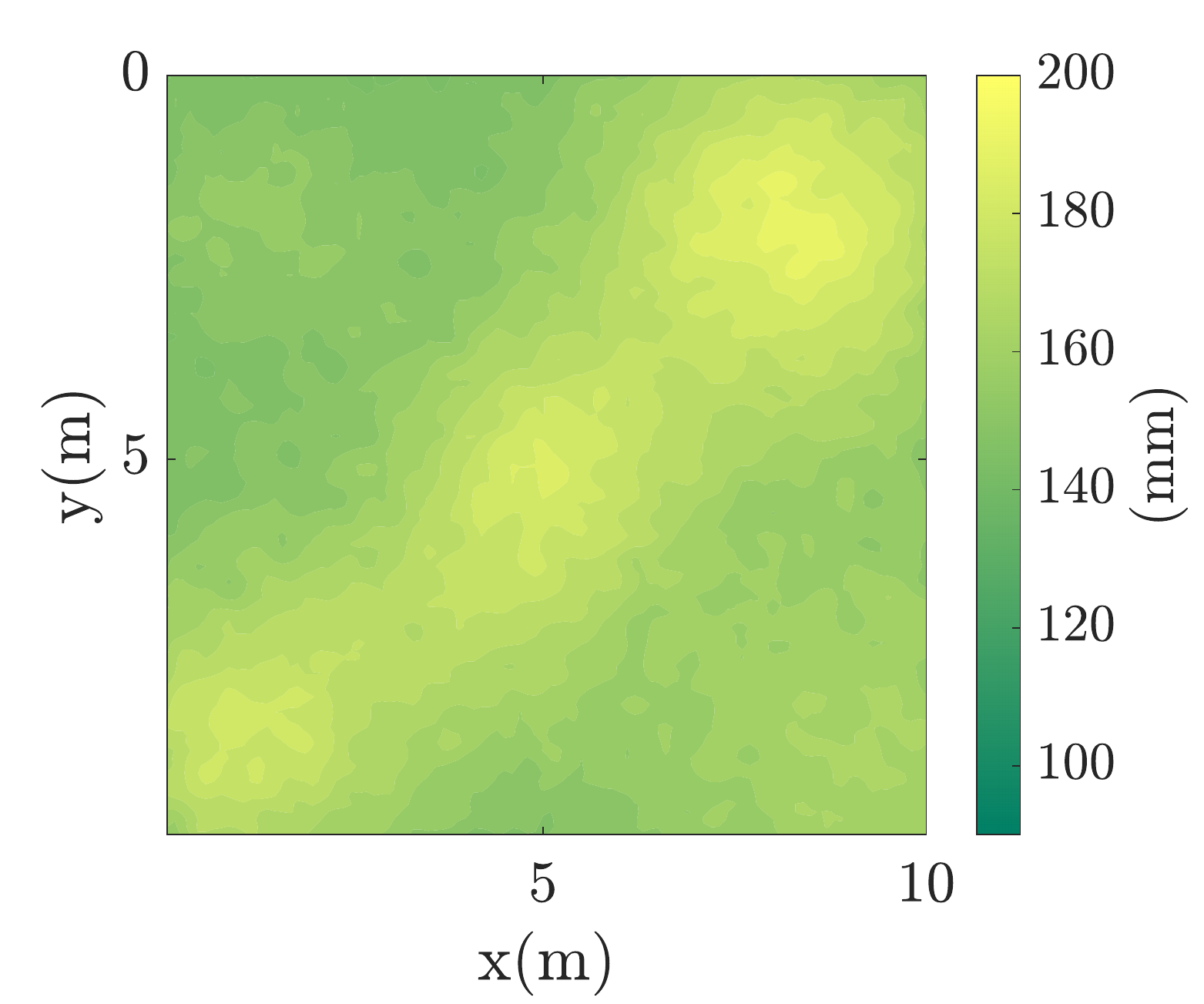}}
	\subfigure[$15$th predicted using DL.]{
		\includegraphics[width = .3\linewidth]{2D_predict_mean_15.pdf}}
	\subfigure[$15$th prediction using GP.]{
		\includegraphics[width = .3\linewidth]{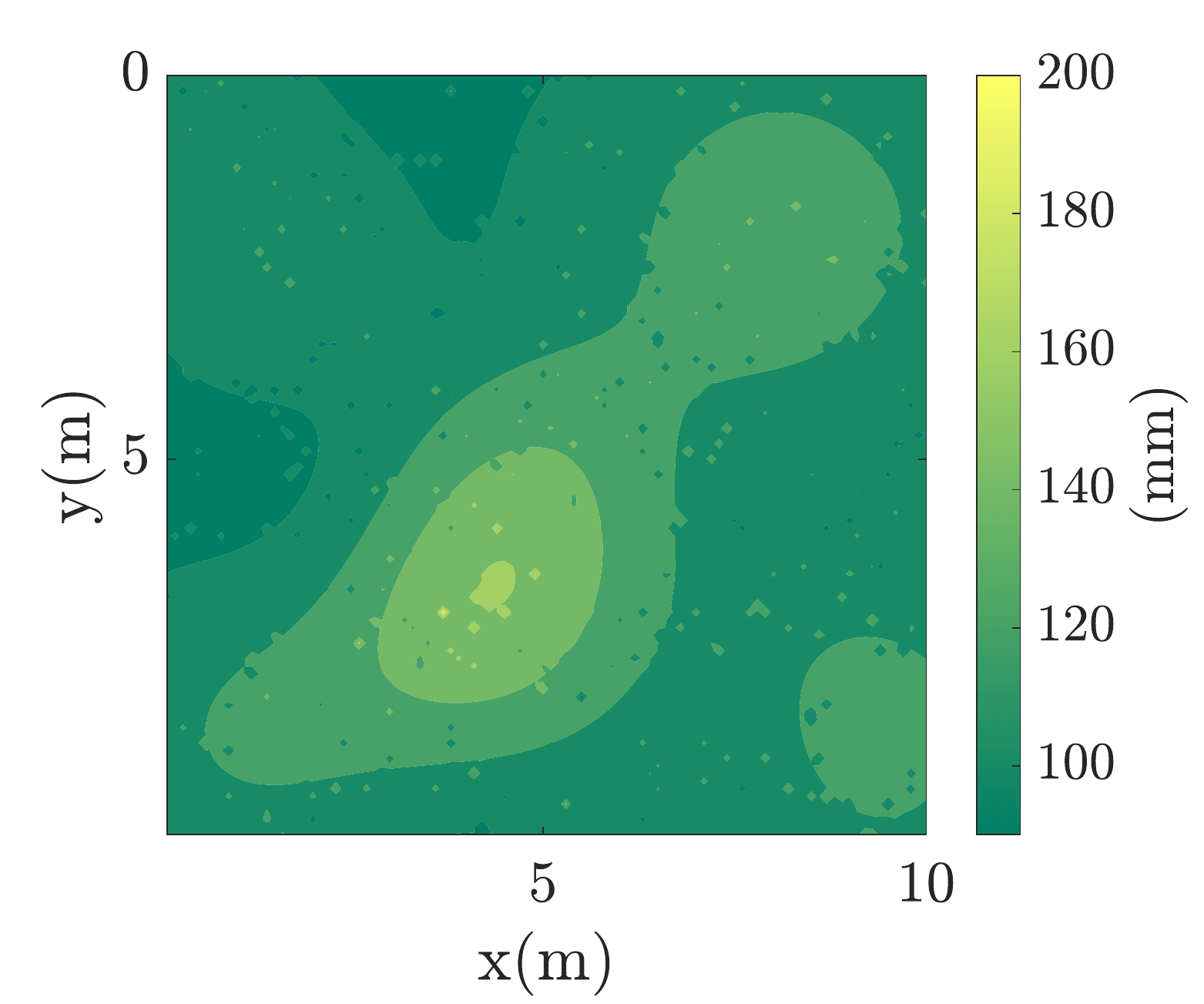}}
	\caption{The comparison between the ground truth and the predictions by using deep learning (DL) based method and Gaussian process (GP) based method.}
	\label{fig: ground truth vs. prediction GP example}
\end{figure*}

Next, we use robots to collect data based on the generated deployment policies. The collected height information will be used to update our knowledge of the environment. Finally, we compare the predictions from each method using this updated information. Note that white noise is also added to simulate measurement noise. Specifically, we use the collected information to make another $10$ more predictions, i.e., from $\alpha = 1$ to $\alpha = 10$, with the size of interval of $\delta = 2$. For each prediction, we compare it with the ground truth. Then, the final averaged prediction error is shown in \figref{fig: raster plot of error comparison}. The averaged prediction error is calculated by averaging the prediction errors from all locations. Since we run the simulation $50$ trials, there are $50$ different averaged prediction errors for each method shown in \figref{fig: raster plot of error comparison}. We see that the proposed pipeline has a lower averaged prediction error than the other two methods after using the updated information from different methods. This demonstrates the effectiveness of the proposed method.

\begin{table}[!tbp]
    \centering
	\caption{The averaged prediction comparisons (in mm) between the deep learning (DL) based method and the Gaussian process (GP) for $\alpha = 5$ and $\alpha = 15$.}
	\centering
	\resizebox{\columnwidth}{!}{\begin{tabular}{l c c c c c}
			\hline\noalign{\smallskip}
			              &        & \multicolumn{2}{c}{(DL prediction)} & \multicolumn{2}{c}{(GP prediction)}                      \\
			Time          & Truth  & Prediction                          & Error                               & Prediction & Error \\
			\noalign{\smallskip}\hline\noalign{\smallskip}
			$\alpha = 5$  & 181.53 & 169.82                              & \textbf{11.71}                        & 149.87     & 31.66 \\
			$\alpha = 15$ & 164.42 & 158.83                              & \textbf{5.60}                         & 117.08     & 47.34 \\
			\noalign{\smallskip}\hline
		\end{tabular}}
	\label{tab: prediction comparisons: DL vs. GP}
\end{table}

\emph{The results from the second comparison:}
In this comparison, we first use GP to make predictions. The settings of this GP are as follows. The training dataset is from the point clouds generated in Gazebo as described in \secref{sec: Gazebo simulation}. We also need to convert those point clouds into heightmaps. We denote the training inputs of the GP by $\x_t = [x, y, t] \in \RR^3$. The training outputs are corresponding grass height of location $(x,y)$ at time $t$. From the above $15$ heightmaps, we randomly pick $1 \times 10^3$ points across all those heightmaps to form the training set. Since GP is a non-parametric method, we select kernels as follows. The mean kernel is defined using the historical mean data as shown in \figref{fig: historical data}. The covariance kernel is a composite of a 2D Gaussian covariance kernel and a 1D linear covariance kernel with noise term. The Gaussian covariance kernel is similar to the one shown in \eqref{eq: Gaussian kernel}. Then, those two covariance kernels are summed up to form the final covariance kernel. In \figref{fig: ground truth vs. prediction GP example}, we show two prediction results when $\alpha = 5$ and $\alpha = 15$. We notice that the proposed learning-based method can maintain more details than that of the GP based predictions. We also include the comparison details of those two comparisons in \tabref{tab: prediction comparisons: DL vs. GP}. Note that the comparison is based on the averaged prediction errors.

\begin{figure}[!tbp]
	\centering
	\includegraphics[width=\imwidth]{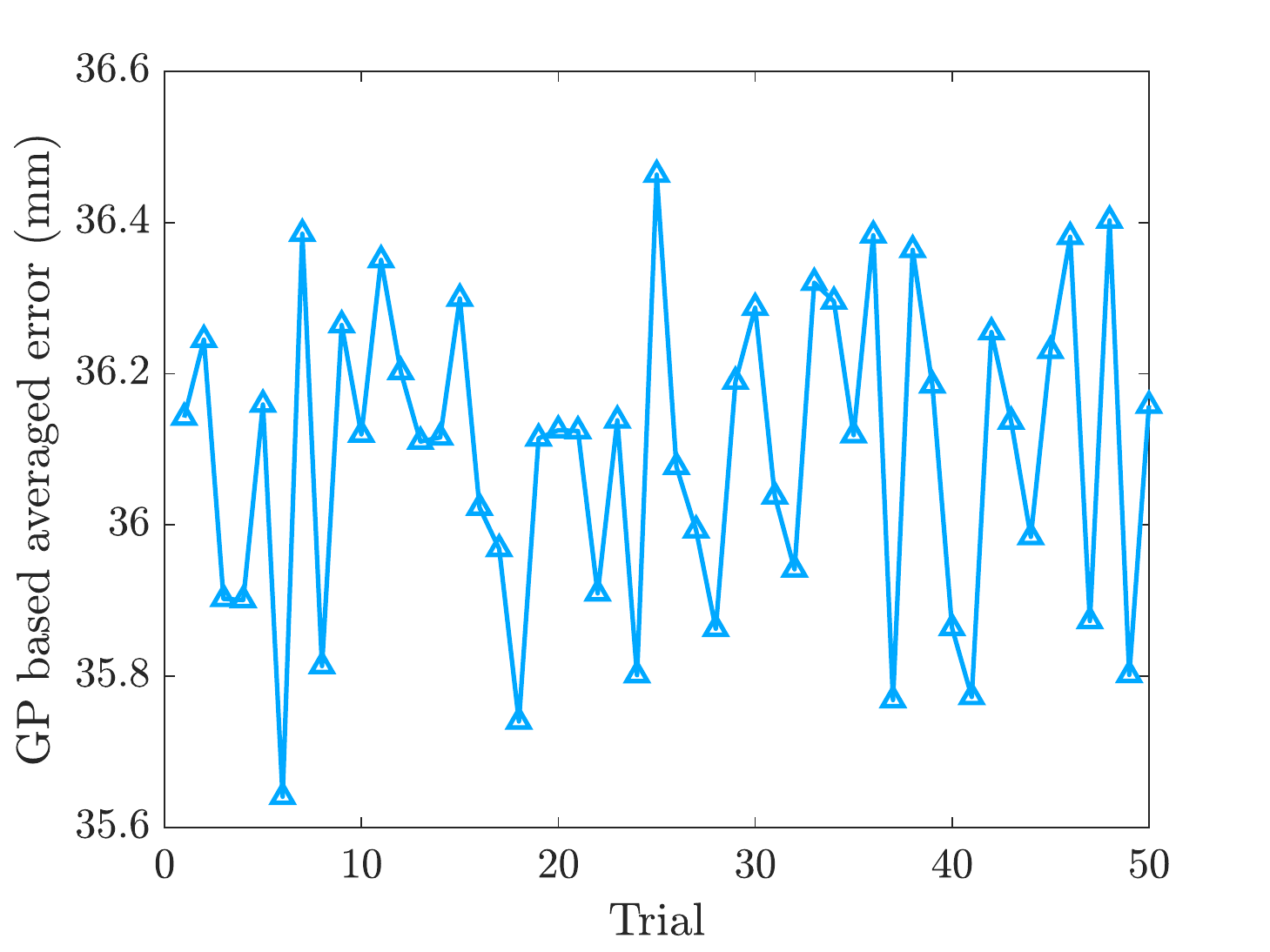}
	\caption{The averaged prediction errors using Gaussian process (GP) and mutual information based pipeline after $50$ trials.}
	\label{fig: GP result}
\end{figure}

\begin{figure}[!tbp]
	\centering
	\includegraphics[width=\imwidth]{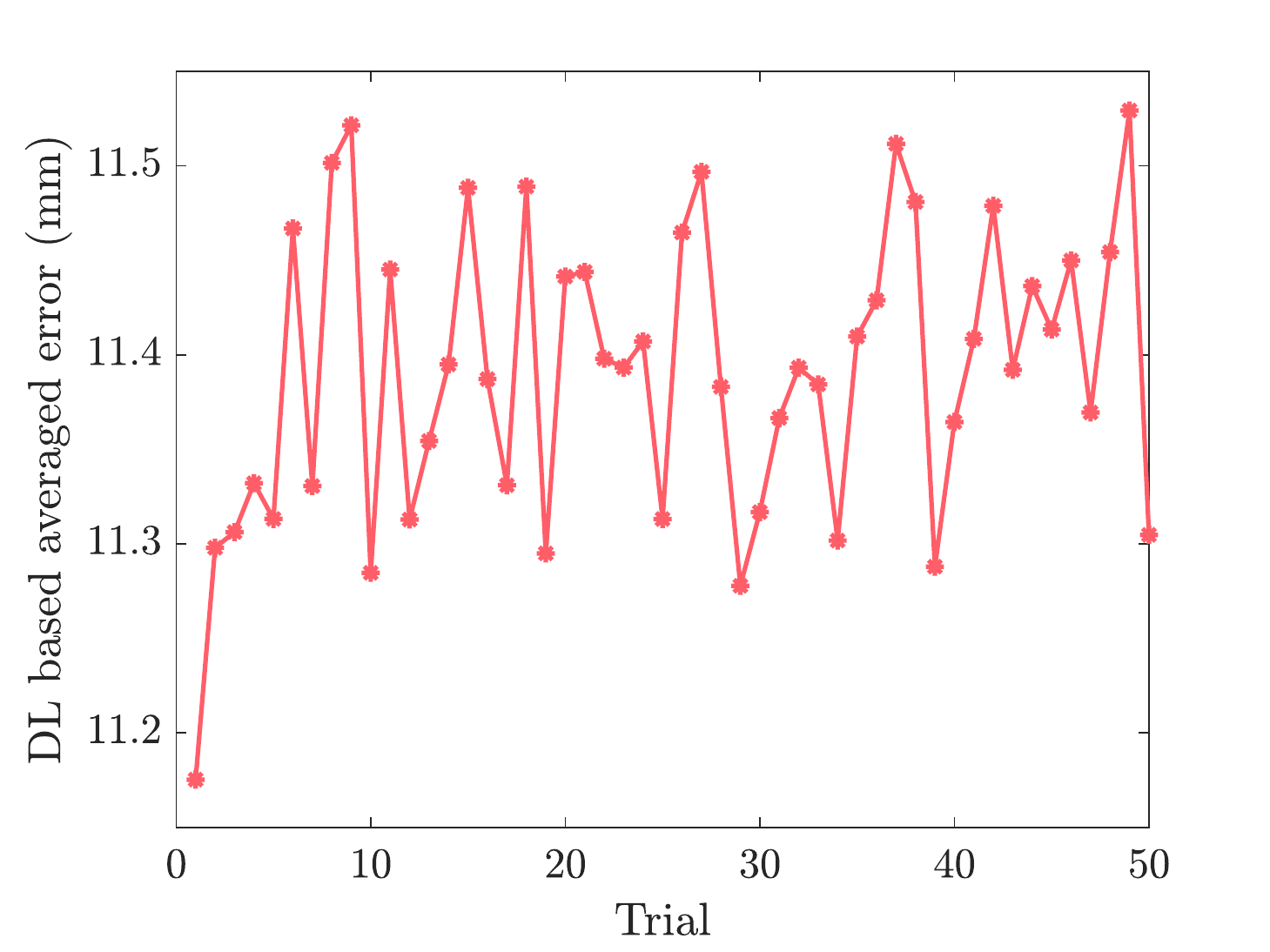}
	\caption{The averaged prediction errors using the proposed deep learning (DL) based pipeline after $50$ trials.}
	\label{fig: DL result}
\end{figure}

Next, we use the proposed intermittent deployment method to select measurement locations based on the mutual information while respecting the proposed constraints \eqref{eq: M1} and \eqref{eq: M2}. Note that the mutual information matrix is build on all the available locations at different times. The parameters and the settings are the same as described in the first comparison. We then run $50$ trials using those settings to generate different deployment policies. Since the mutual information-based method requires a full covariance matrix for the deployment ground set $\V$, which is extremely computational expensive, we choose to reduce the original $100 \times 100$ deployable locations to $10 \times 10$ deployable locations. Based on the deployment result of each trial, we finally make predictions for the next $10$ steps. Therefore, the performance comparison between this mutual information-based pipeline and the proposed pipeline is based on the final prediction results. In \figref{fig: GP result} and \figref{fig: DL result}, we demonstrate the averaged prediction error of each pipeline. Also, the statistics of this comparison are shown in \tabref{tab: pipeline DL vs. GP}.

\begin{table}[!tbp]
    \centering
	\footnotesize
	\caption{The statistics of the averaged prediction errors of the proposed deep learning (DL) based pipeline and the Gaussian process (GP) based pipeline using $50$ trials.}
	\begin{tabular}{lll}
		\hline\noalign{\smallskip}
		                  & Mean (mm)    & Std. (mm)   \\
		\noalign{\smallskip}\hline\noalign{\smallskip}
		DL based pipeline & \textbf{11.39} & \textbf{0.08} \\
		GP based pipeline & 36.09        & 0.21        \\
		\noalign{\smallskip}\hline
	\end{tabular}
	\label{tab: pipeline DL vs. GP}
\end{table}

\subsection{Perception Results}
\label{ssec: perception results}

\begin{figure}[!tbp]
	\centering
	\subfigure[]{
		\label{fig: drone1}
		\includegraphics[width = 1.1in]{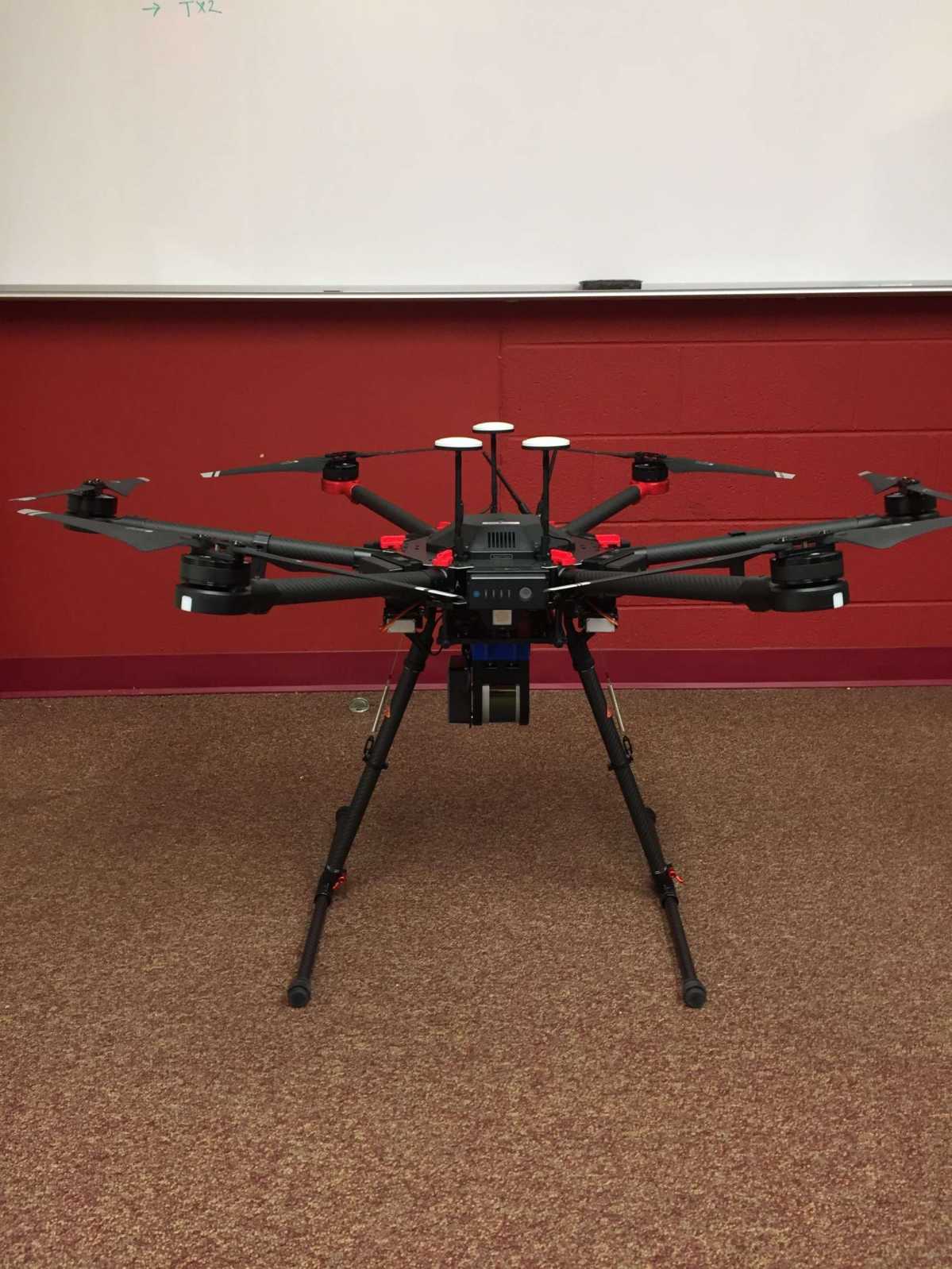}}
	\hspace{3mm}
	\subfigure[]{
		\label{fig: drone3}
		\includegraphics[width = 1.1in]{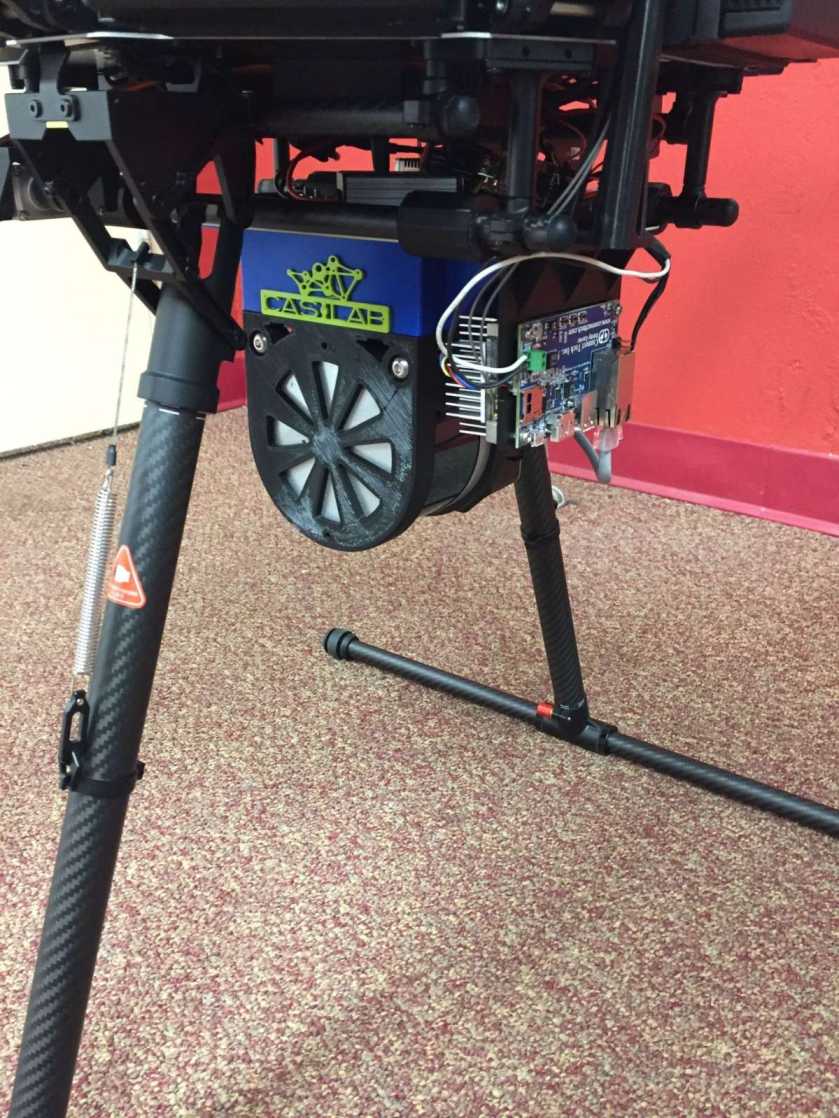}}
	\caption{A UAV (DJI M600) mounted with a LiDAR (Velodyne VLP-16) and an onboard system (Nvidia Jetson TX2).}
	\label{fig: drone}
\end{figure}

In this section, we demonstrate the result from a field experiment to illustrate the effectiveness of our simulated pastureland environment.

We experimented at Virginia Tech's Turfgrass Research Center to test out our growth analysis algorithm. A $10$m $\times 10$m region of the Turfgrass center was reserved, and the grass was grown. It was untouched over time, but the perimeter around it was consistently mowed down. A DJI M600 UAV with a bottom-facing Velodyne VLP-16 3D-LiDAR, as shown in \figref{fig: drone}, was flown over the region. The localization information and LiDAR scans during flights were collected using an onboard NVIDIA Jetson TX2. This data was used to build a point cloud map of the region using a similar technique to the one used in the point cloud generation for the simulation discussed previously. These flights were conducted weekly to get temporal point cloud maps of the region. Manual measurements of the region were also collected. Nine spots throughout the region were used as the manual measurement points. These were averaged to get the manual measurements shown in Table~\ref{tab:turfgrass_height}.

\begin{table}[!tbp]
    \centering
	\caption{Turfgrass experiment heights.}
	\resizebox{\columnwidth}{!}{\begin{tabular}{lllll}
			\hline\noalign{\smallskip}
			                    & Week 1     & Week 2     & Week 3     & Week 4     \\
			\noalign{\smallskip}\hline\noalign{\smallskip}
			Hand Measurement    & $48.00$ cm & $68.17$ cm & $74.11$ cm & $69.04$ cm \\
			$50$th Percentile   & $3.97$ cm  & $10.67$ cm & $12.17$ cm & $9.60$ cm  \\
			$75$th Percentile   & $6.31$ cm  & $16.01$ cm & $18.41$ cm & $14.60$ cm \\
			$90$th Percentile   & $8.73$ cm  & $22.08$ cm & $26.11$ cm & $19.93$ cm \\
			$95$th Percentile   & $11.03$ cm & $27.48$ cm & $31.53$ cm & $24.49$ cm \\
			$97.5$th Percentile & $13.43$ cm & $32.22$ cm & $36.64$ cm & $30.43$ cm \\
			$99$th Percentile   & $15.63$ cm & $36.43$ cm & $41.65$ cm & $36.17$ cm \\
			$99.5$th Percentile & $16.70$ cm & $39.23$ cm & $44.29$ cm & $39.42$ cm \\
			\noalign{\smallskip}\hline
		\end{tabular}}
	\label{tab:turfgrass_height}
\end{table}

\begin{table}[!tbp]
    \centering
	\footnotesize
	\caption{Turfgrass experiment growth.}
	\begin{tabular}{llll}
		\hline\noalign{\smallskip}
		                  & Week 1-2 & Week 2-3 & Week 3-4 \\
		\noalign{\smallskip}\hline\noalign{\smallskip}
		Hand Measurement  & 20.17 cm & 5.94 cm  & -5.07 cm \\
		50th Percentile   & 6.72 cm  & 1.49 cm  & -2.58 cm \\
		75th Percentile   & 9.70 cm  & 2.40 cm  & -3.81 cm \\
		90th Percentile   & 13.34 cm & 4.03 cm  & -6.18 cm \\
		95th Percentile   & 16.45 cm & 4.05 cm  & -7.05 cm \\
		97.5th Percentile & 18.79 cm & 4.40 cm  & -6.21 cm \\
		99th Percentile   & 20.80 cm & 5.22 cm  & -5.48 cm \\
		99.5th Percentile & 22.54 cm & 5.05 cm  & -4.87 cm \\
		\noalign{\smallskip}\hline
	\end{tabular}
	\label{tab:turfgrass_growth}
\end{table}

\begin{figure}[!tbp]
	\centering
	\includegraphics[width=\imwidth]{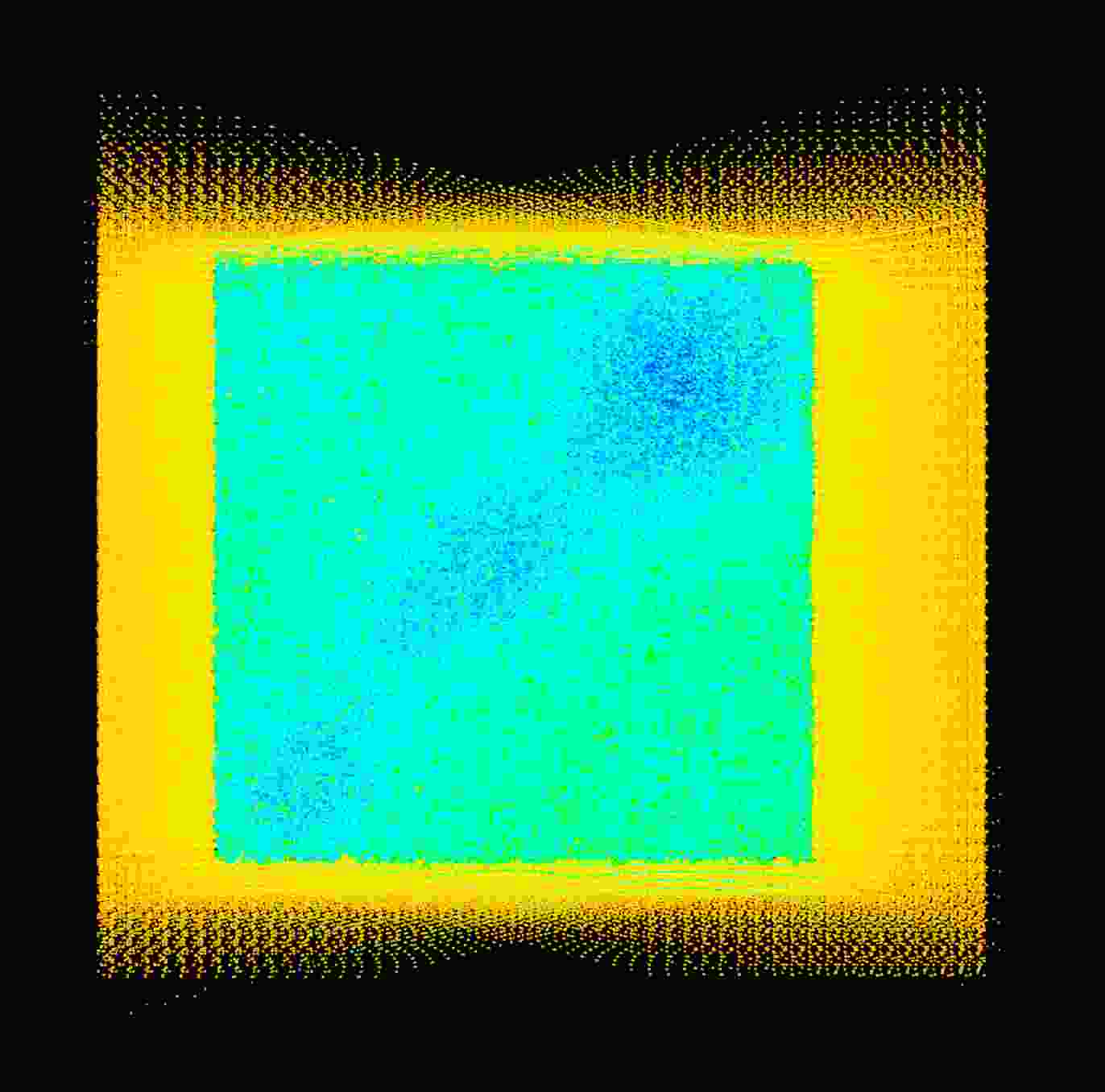}
	\caption{Unprocessed point cloud of $10 \times 10$ meter Turfgrass plot with the perimeter. The UAV flew over the generated plot in the simulation. The data collected from the 3D LiDAR is shown in this figure. The yellow points are the perimeter around the plot, and the blue/green points are the points in the plot. With the plot being taller, the points registered were closer. The bluer the point, the closer it is.}
	\label{fig: step_1_day211_raw_data_pc}
\end{figure}

The distance between every point in the region's point cloud and the ground plane was computed to estimate the height of these points. This was done using the normal vector determined using the method described \secref{sec: Gazebo simulation}. Because the vast majority of LiDAR points were not at the very top of the grass, the height estimations using this method were under-estimations regardless of which percentile we looked at. This is shown in Table~\ref{tab:turfgrass_height}, and the result is shown in \figref{fig: step_1_day211_raw_data_pc}. However, since growth is what we are looking into, we can look at the relative differences in the estimations between each flight. These differences are shown in Table~\ref{tab:turfgrass_growth} with the first row showing the manually measured growth between sampling weeks. With perfect estimations, the percent difference between the manual measurements and estimated height would be $0\%$. However, our best results were found using the 99th percentile of growth estimates which had an average percent difference of $7.9\%$ compared to manual measurements.

Those real-world experiments shown above validate our pasture simulation regime and thus our evaluation results. Meanwhile, the point cloud results from the real-world experiments are close to the results in our simulated world. All those results demonstrate the effectiveness of our proposed pipeline.

\section{Conclusions and Future Work}
\label{sec: conclusions and future work}

In this work, we proposed an integrated pipeline that can be used for long-term, large-scale forage perception applications. From the perspective of simulation, we demonstrated how to simulate large pastureland environments reasonably fast using parallel processing. From the perspective of pasture prediction, we proposed a new deep learning architecture that can be used for long-term pasture predictions. From the perspective of perception, we demonstrated how to get accurate pasture height estimation through regression. From the perspective of autonomy, we combined predictions and an intermittent deployment policy to deploy robots with high accuracy while at low cost.

This work resulted in novel approaches from the initial data-generating to the final deployment testing. The proposed pipeline offers a promising and cost-effective alternative to real-life experiments and can be used as a platform for other testings.

\bibliographystyle{IEEEtran}
\bibliography{ms}


\end{document}